\documentclass[final]{cvpr}

\usepackage{times}
\usepackage{epsfig}
\usepackage{graphicx}
\usepackage{amsmath}
\usepackage{amssymb}

\usepackage{caption}
\usepackage[misc]{ifsym}
\usepackage{array}
\usepackage{epstopdf}
\usepackage{color}
\usepackage{psfrag}
\usepackage[nocompress]{cite}
\usepackage{hhline}
\usepackage{multirow}
\usepackage{bm}
\usepackage{adjustbox}
\usepackage{overpic}
\usepackage{rotating}
\usepackage[american]{babel}
\usepackage{booktabs}
\usepackage{makecell}
\usepackage{threeparttable}
\usepackage{cuted}
\usepackage{bbding}

\usepackage[pagebackref=true,breaklinks=true,colorlinks,bookmarks=false]{hyperref}

\def\cvprPaperID{3000} 
\def\confYear{CVPR 2021}

\begin{document}

\title{Two-Stage Single Image Reflection Removal\\with Reflection-Aware Guidance}
\author{Yu Li\textsuperscript{1}, Ming Liu\textsuperscript{1}, Yaling Yi\textsuperscript{1}, Qince Li\textsuperscript{1}, Dongwei Ren\textsuperscript{2}, Wangmeng Zuo\textsuperscript{1(\Letter)}\\
	$^1$School of Computer Science and Technology, Harbin Institute of Technology, China\\
	$^2$College of Intelligence and Computing, Tianjin University, China\\
	{\tt\small \{\href{mailto:liyuhit@outlook.com}{\color{black}liyuhit}, \href{mailto:csmliu@outlook.com}{\color{black}csmliu}, \href{mailto:csylyi@outlook.com}{\color{black}csylyi\}@outlook.com}, \{\href{mailto:qinceli@hit.edu.cn}{\color{black}qinceli}, \href{mailto:wmzuo@hit.edu.cn}{\color{black}wmzuo\}@hit.edu.cn}, \href{mailto:rendongweihit@gmail.com}{\color{black}rendongweihit@gmail.com}.}
}

\maketitle

\begin{abstract}
Removing undesired reflection from an image captured through a glass surface is a very challenging problem with many practical application scenarios.
For improving reflection removal, cascaded deep models have been usually adopted to estimate the transmission in a progressive manner.
However, most existing methods are still limited in exploiting the result in prior stage for guiding transmission estimation.
In this paper, we present a novel two-stage network with reflection-aware guidance (RAGNet) for single image reflection removal (SIRR).
To be specific, the reflection layer is firstly estimated due to that it generally is much simpler and is relatively easier to estimate.
Reflection-aware guidance (RAG) module is then elaborated for better exploiting the estimated reflection in predicting transmission layer.
By incorporating feature maps from the estimated reflection and observation, RAG can be used (i) to mitigate the effect of reflection from the observation, and (ii) to generate mask in partial convolution for mitigating the effect of deviating from linear combination hypothesis.
A dedicated mask loss is further presented for reconciling the contributions of encoder and decoder features.
Experiments on five commonly used datasets demonstrate the quantitative and qualitative superiority of our RAGNet in comparison to the state-of-the-art SIRR methods.
The source code and pre-trained model are available at \emph{\url{https://github.com/liyucs/RAGNet}}.

\end{abstract}

\begin{figure*}[thbp]
    \small
    \centering
    \scalebox{1}{
        \begin{tabular}{cccc}
            \includegraphics[width=0.23\textwidth]{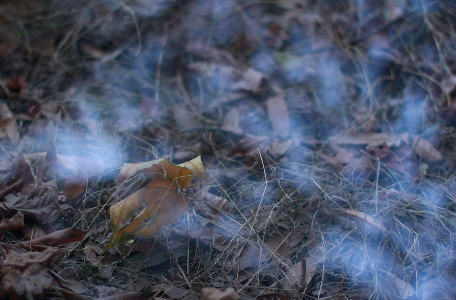}&\hspace{-4.2mm}
            \includegraphics[width=0.23\textwidth]{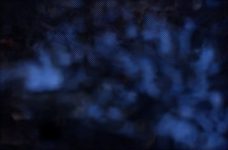}&\hspace{-4.2mm}
            \includegraphics[width=0.23\textwidth]{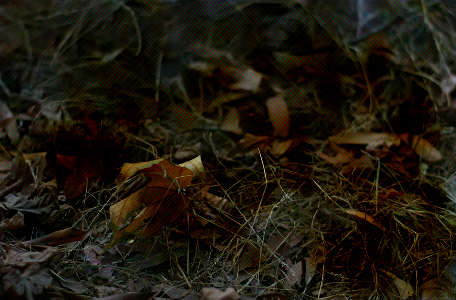}&\hspace{-4.2mm}
            \includegraphics[width=0.23\textwidth]{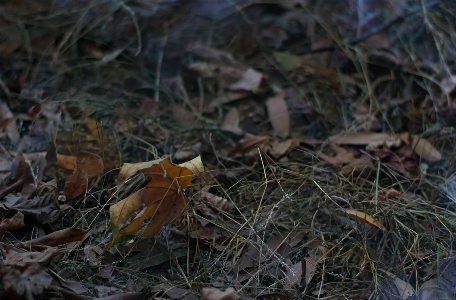}
            \\
            Observation $I$&\hspace{-4.2mm}
            Predicted Reflection $\hat{R}$&\hspace{-4.2mm}
            Visualization of $I-\hat{R}$& \hspace{-4.2mm}
            Ours
            \\
            \includegraphics[width=0.23\textwidth]{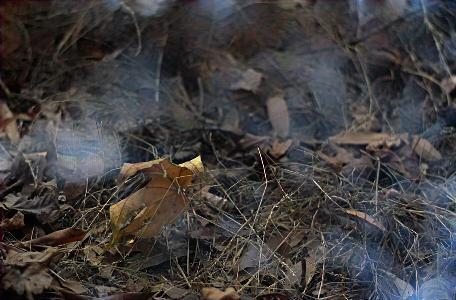}&\hspace{-4.2mm}
            \includegraphics[width=0.23\textwidth]{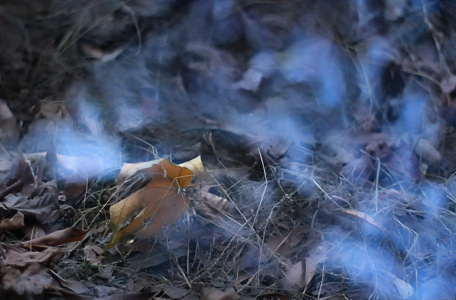}&\hspace{-4.2mm}
            \includegraphics[width=0.23\textwidth]{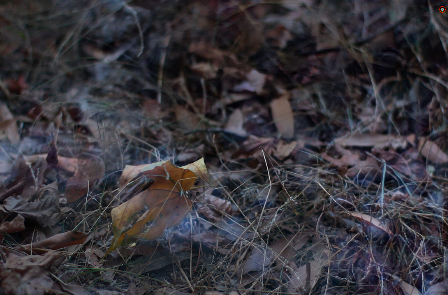}&\hspace{-4.2mm}
            \includegraphics[width=0.23\textwidth]{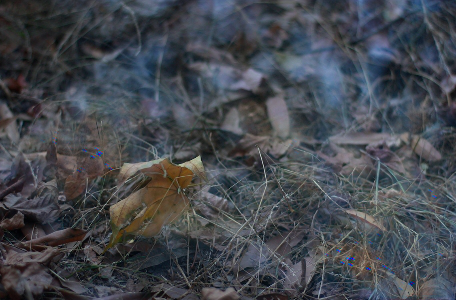}
            \\
            CEILNet~\cite{CEILNet}&\hspace{-4.2mm}
            BDN~\cite{BDN}&\hspace{-4.2mm}
            Lei \etal~\cite{Lei_2020_CVPR}&\hspace{-4.2mm}
            IBCLN~\cite{IBCLN}
            \\
        \end{tabular}}
    \vspace{-3mm}
    \captionsetup{font={small}}
    \caption{An example of real-world observation from \emph{Real 20} dataset and the reflection predicted by our model. It can be seen that the linear combination is violated in regions with heavy reflection. The results of competing methods are given in the second row.}
    \label{fig:introduction}
\end{figure*}

\section{Introduction}\label{sec:introduction}
Reflections are commonly observed in images taken through glass, where the light reflected by the glass (\ie, \emph{reflection}) and that from the objects behind the glass (\ie, \emph{transmission}) are both captured by the camera.
Undesired reflections inevitably degrade image visual quality and hinder many subsequent vision applications such as autonomous driving and entertainment, making reflection removal both very challenging and practically valuable in low level vision~\cite{Punnappurath_2019_CVPR, CEILNet, ERRNet}.
To mitigate the ill-posedness of reflection removal, early studies usually resort to multiple input images captured under different illumination conditions, focal lengths, polarizer angles, and camera viewpoints.
On the contrary, single image reflection removal (SIRR)~\cite{Zhang2018CVPR, BDN, IBCLN} is more common in task setting and practically more significant, and has received considerable interest in the recent few years.

Following the formulation in \cite{ERRNet, Zhang2018CVPR}, the input image, \ie, an observation $I$, can be viewed as a linear combination of a transmission layer $T$ and a reflection layer $R$, \ie, $I=T+R$.
The goal of SIRR is to recover the transmission layer $T$ for the observation $I$.
Unfortunately, the intrinsic ill-posedness of SIRR makes it very challenging to learn a direct mapping from the input image $I$ to the transmission $T$.
To ease the difficulty of training SIRR, cascaded deep models are usually adopted to mitigate the uncertainty of transmission estimation~\cite{CEILNet,BDN,IBCLN}.
For example, \cite{CEILNet} first estimates the edge map and then the transmission, while \cite{BDN} and \cite{IBCLN} progressively refine the estimation of reflection and transmission layers.
However, there remain several interesting issues to further investigate (i) what should be estimated in the prior stage and (ii) how to exploit the result in prior stage for guiding succeeding transmission estimation, thereby leaving some leeway in improving SIRR.

In this paper, we take a step forward to address the above issues, and present a novel  two-stage network with reflection-aware guidance (RAGNet).
To be specific, our RAGNet first estimates the reflection layer $\hat{R}$ in the first stage.
Then, it takes both the estimated reflection $\hat{R}$ and observation $I$ as the input to predict the transmission layer $\hat{T}$.
Fig.~\ref{fig:introduction} shows an example of an input image as well as the estimated reflection by our RAGNet.
The reasons to estimate the reflection layer in the first stage can be explained from three aspects.
First, the reflection layer generally is much simpler than the transmission layer and is relatively easier to estimate.
Second, the estimated reflection is beneficial to transmission estimation.
For example, the difference between the observation and estimated reflection, \ie, $I - \hat{R}$, can serve as a reasonable initialization of the transmission.
Third, the linear combination $I=T+R$ may not hold true for real-world observations, especially those images with heavy and bright reflection (see Fig.~\ref{fig:introduction}).
Also, the illumination of estimated reflection is helpful in finding the regions violating the linear combination hypothesis.

\begin{table}[t]
	\captionsetup{font={small}}
    \caption{\footnotesize Comparison of deep cascaded models for SIRR. $I$, $R$, $T$ and $E$ denote observation, reflection, transmission and edge maps, respectively.}
    \vspace{-3mm}
    \centering
    \label{tab:introduction}
    \scalebox{0.8}{
    \begin{tabular}{lcc}%
    \toprule
    Methods & Multi-stage Scheme & Cross-stage Fusion \\
    \midrule
    CEILNet~\cite{CEILNet} & $I \to E \to T$ & Concat  \\
    BDN~\cite{BDN} & $I \to T_0 \to R \to T$ & Concat  \\
    IBCLN~\cite{IBCLN} & $(\mathbf{0},I) \to (R_1, T_1) \to \ldots $  & Concat+LSTM  \\
    Ours & $I \to R \to T$ & RAG \\
    \bottomrule
    \end{tabular}}
\vspace{-2.5mm}
\end{table}

Furthermore, we elaborate a reflection-aware guidance (RAG) module for better exploiting the estimated reflection in the second stage.
As summarized in Table~\ref{tab:introduction}, the output of the prior stage is concatenated with the original observation as the input of the subsequent stage in~\cite{BDN,CEILNet}.
Li \etal~\cite{IBCLN} further applied a recurrent layer (\ie, LSTM~\cite{LSTM}) to exploit the deep features across stages.
In comparison, the second stage of RAGNet includes an encoder for the estimated reflection and an encoder-decoder for the original observation.
For each block of the decoder, an RAG module is elaborated for better leveraging the estimated reflection from two aspects.
On the one hand, the encoder feature map of the reflection is utilized to suppress that of the observation, and their difference is then concatenated with the decoder feature map of the observation.
On the other hand, the encoder feature maps of the reflection and observation are incorporated with the decoder feature map to generate a mask indicating the extent to which the transmission is corrupted by heavy reflection.
The mask is then collaborated with partial convolution to mitigate the effect of deviating from linear combination hypothesis.
And a dedicated mask loss is further presented for reconciling the contributions of encoder and decoder features.

Extensive experiments are conducted on five commonly used real-world datasets to evaluate our RAGNet.
With the RAG module, the estimated reflection is effective in guiding the transmission estimation and alleviating the effect of violating linear combination hypothesis.
In comparison with the state-of-the-art models~\cite{CEILNet, Zhang2018CVPR, BDN, ERRNet, IBCLN}, our RAGNet performs favorably in terms of both quantitative metrics and visual quality.
In general, the main contribution of this work is summarized as follows:
\begin{itemize}
    \item A SIRR network RAGNet is presented, which involves two stages, \ie, first estimating reflection and then predicting transmission. Its rationality is also explicated to differentiate from existing cascaded SIRR models.
    \item For better leveraging the estimated reflection in the second stage, the reflection-aware guidance (RAG) module is further introduced to guide the transmission estimation and alleviate the effect of violating linear combination hypothesis.
    \item Experiments on five popular datasets show that our RAGNet performs favorably against the state-of-the-art methods quantitatively and qualitatively.
\end{itemize}

\section{Related Work}\label{sec:related_work}
In this section, we briefly review two categories of relevant methods on reflection removal tasks, \ie, multiple image reflection removal and single image reflection removal.

\subsection{Multiple Image Reflection Removal}\label{subsec:MIRR}
Due to the ill-posed nature, multiple images have been exploited to solve the reflection removal problem by providing extra information.
Agrawal \etal~\cite{2005Removing} used two separate images as input, which are captured with and without flash light, respectively, and Fu \etal~\cite{Fu_2013_ICCV} further took into consideration the high frequency illumination.
Schechner \etal~\cite{2000Separation} treated the input image as a superimposition of two layers, and novel blur kernels were proposed to eliminate mutual penetration of the separated layers, using images taken with different focal lengths as a clue.
Polarization camera is also used to take advantage of the polarization properties of light~\cite{2004Separating, Wieschollek_2018_ECCV, Lei_2020_CVPR}.
{In particular, based on the linear hypothesis $I=T+R$ in raw RGB space, Lei \etal~\cite{Lei_2020_CVPR} adopted a simple two-stage model but took multiple polarized images as input.}
In these methods, the behavioral difference between the reflection layer and the transmission layer under different settings serves as an effective clue for reflection removal.

Unlike the aforementioned methods, whose input images are captured in a fixed position, taking images from different viewpoints is another popular way to exploit multiple image information in reflection removal tasks~\cite{2015A, 855826, Li_2013_ICCV, Guo_2014_CVPR, 2004Separating, 5765994, 2012Image, Yang_2016_CVPR, sun2016automatic, HAN_2017_CVPR, Simon_2015_CVPR}.
In particular, Punnappurath \etal~\cite{Punnappurath_2019_CVPR} leveraged the newly developed dual pixel (DP) cameras, and images captured from two sub-aperture views are readily accessible.
Although approaches in the multi-input paradigm can obtain impressive results, the necessity to prepare qualified input images with specially designed hardware or extra human efforts makes them inconvenient, and the prerequisites are unable to be satisfied in many practical scenarios, which boosts the recent prevalence of single image reflection removal.

\subsection{Single Image Reflection Removal}\label{subsec:SIRR}
As the name implies, SIRR methods merely require a single image as input.
Traditional methods employ various priors observed in real-world images to reduce the ill-posedness, such as gradient sparsity prior~\cite{4288165, 8300655, Arvanitopoulos_2017_CVPR, Sandhan_2017_CVPR}, smoothness prior~\cite{Li_2014_CVPR, 7532311} and ghosting cues~\cite{Shih_2015_CVPR}.
Unfortunately, the representation and generalization ability of such methods is limited, and they are prone to produce poor results in the absence of discriminative hints, which is often the case when using handcrafted priors.

As a remedy, learning based methods have been proposed to solve the SIRR problem by leveraging the capacity of deep convolutional neural networks (CNNs), which mainly focus on data synthesis, loss function and network design.
Due to the tremendous effort required to collect sufficient well-aligned real image pairs for model training, Fan \etal~\cite{CEILNet} generated reflection layers via smoothing and illumination intensity decay, which were then used to synthesize the observation by linear combination.
Zhang \etal~\cite{Zhang2018CVPR} further adapted the intensity decay to suit reflections with higher illuminance level, and applied random vignette to simulate different camera views.
Besides, several loss functions have been particularly designed to leverage the characteristic of reflections.
CEILNet~\cite{CEILNet} constrains the gradient maps of the predicted transmission layer to avoid blurry output.
Zhang \etal~\cite{Zhang2018CVPR} proposed an exclusion loss by minimizing the correlation between the gradient maps of estimated transmission and reflection layers, based on the observation that the edges in these two layers are unlikely to overlap.
Li \etal~\cite{IBCLN} took advantage of the synthesis model to reconstruct the observation, which is constrained to approximate the input.
%
As to network structure, most recent works follow two main schemes.
On the one hand, \cite{Zhang2018CVPR} and \cite{ERRNet} extracted input features via a pre-trained VGG-19~\cite{VGG} model and stack them together for subsequent usage, which exploits multi-scale features and is tolerant of mild misalignment between real-world training pairs.
On the other hand, the encoder-decoder framework is exploited in \cite{CEILNet, ERRNet, BDN, IBCLN}, where skip connections are usually integrated for better feature utilization.

Considering the difficulty of directly estimating the transmission layer, cascaded models have been widely adopted in the hope of finding a better reflection removal pathway.
Fan \etal~\cite{CEILNet} first predicted potential edge maps, and then the transmission layer is generated given the edge maps.
Yang \etal~\cite{BDN} alternated between reflection and transmission estimation, while Li \etal~\cite{IBCLN} further incorporated a long short-term memory (LSTM) module~\cite{LSTM} to generate the reflection and the transmission layer in a progressive manner.
Such multi-stage methods are closely related to our proposed method, however, as further discussed in Sec.~\ref{subsec:limitations}, they are still limited in exploiting the result from prior stage.

We note that the MIRR method~\cite{Lei_2020_CVPR} also uses a two-stage network but takes raw polarized images with different polarization angles as input.
Moreover, it adopts the linear hypothesis $I=T+R$ and {its} concatenation-based fusion is limited in exploiting the estimated reflection for the {subsequent} stage.
Empirically, the retraining of adjusted MIRR by feeding a single unpolarized image is still limited in removing heavy and bright reflections (See Fig.~\ref{fig:introduction}).
%

\section{Proposed Method}\label{sec:method}

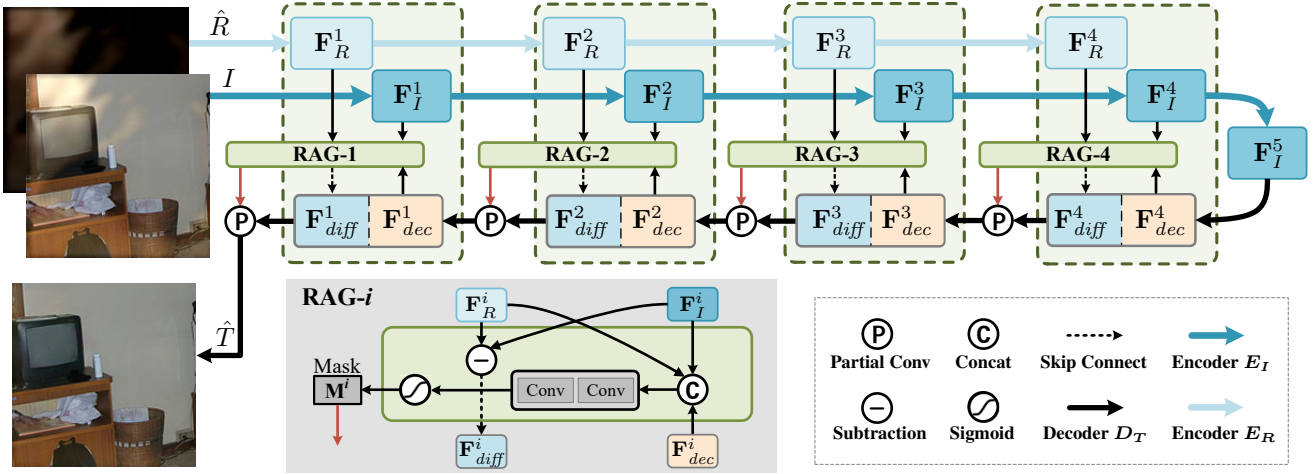
\begin{figure*}[htbp]
	\begin{overpic}[width=1\textwidth]{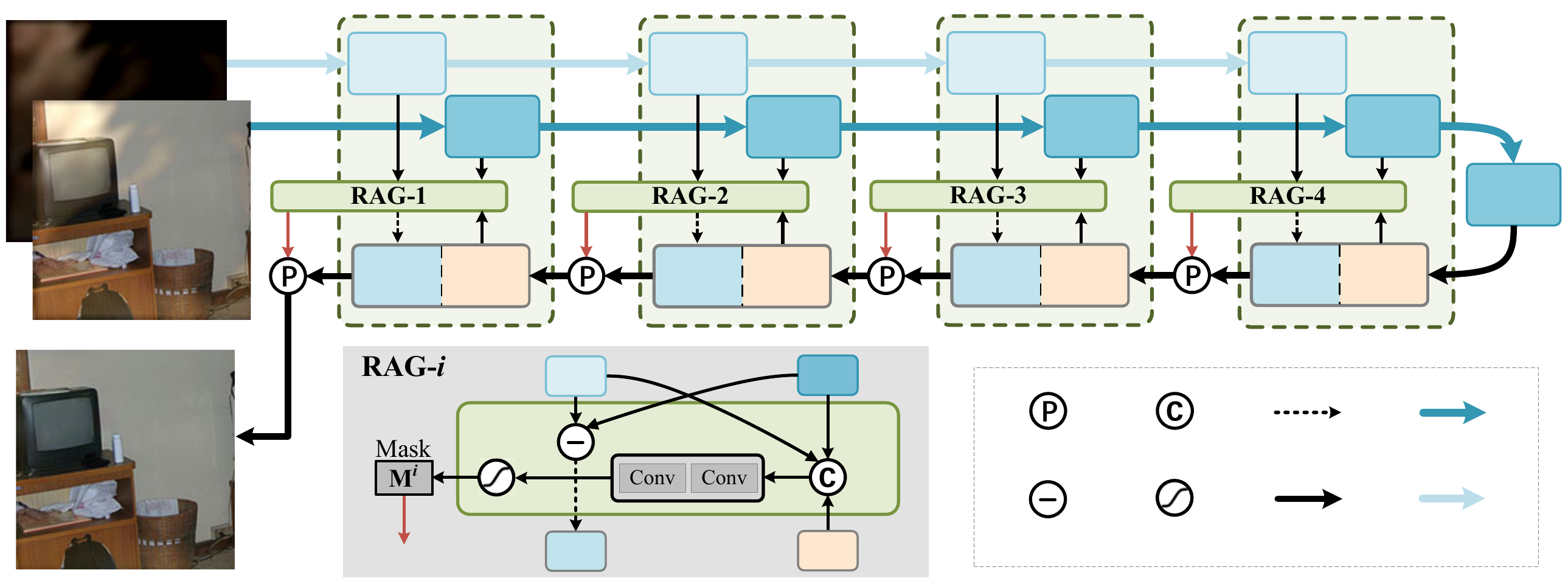}
		\put(16,33.8){\normalsize $\hat{R}$}
		\put(17,29.8){\normalsize $I$}
		\put(16.5,10.1){\normalsize $\hat{T}$}
		\put(24,32.6){\normalsize $\mathbf{F}_R^1$}
		\put(43,32.6){\normalsize $\mathbf{F}_R^2$}
		\put(62.2,32.6){\normalsize $\mathbf{F}_R^3$}
		\put(81.4,32.6){\normalsize $\mathbf{F}_R^4$}
		\put(30,28.6){\normalsize $\mathbf{F}_I^1$}
		\put(49,28.6){\normalsize $\mathbf{F}_I^2$}
		\put(68.2,28.6){\normalsize $\mathbf{F}_I^3$}
		\put(87.4,28.6){\normalsize $\mathbf{F}_I^4$}
		\put(95.5,24.2){\normalsize $\mathbf{F}_I^5$}
		\put(23.3,19.1){\normalsize $\mathbf{F}_\mathit{diff}^1$}
		\put(29.2,19.1){\normalsize $\mathbf{F}_\mathit{dec}^1$}
		\put(42.3,19.1){\normalsize $\mathbf{F}_\mathit{diff}^2$}
		\put(48.2,19.1){\normalsize $\mathbf{F}_\mathit{dec}^2$}
		\put(61.9,19.1){\normalsize $\mathbf{F}_\mathit{diff}^3$}
		\put(67.2,19.1){\normalsize $\mathbf{F}_\mathit{dec}^3$}
		\put(80.3,19.1){\normalsize $\mathbf{F}_\mathit{diff}^4$}
		\put(86.5,19.1){\normalsize $\mathbf{F}_\mathit{dec}^4$}

		\put(35.5,12.8){\footnotesize $\mathbf{F}_R^i$}
		\put(51.8,12.8){\footnotesize $\mathbf{F}_I^i$}
		\put(35,1.7){\footnotesize $\mathbf{F}_\mathit{diff}^i$}
		\put(51.2,1.7){\footnotesize $\mathbf{F}_\mathit{dec}^i$}
		\put(89.3,8.4){\textbf{\scriptsize Encoder $\bm{E_I}$}}
		\put(89.3,3){\textbf{\scriptsize Encoder $\bm{E_R}$}}
		\put(79.5,3){\textbf{\scriptsize Decoder $\bm{D_T}$}}
		\put(79.3,8.4){\textbf{\scriptsize Skip Connect}}
		\put(72.8,8.4){\textbf{\scriptsize Concat}}
		\put(63.3,8.4){\textbf{\scriptsize Partial Conv}}
		\put(63.5,3){\textbf{\scriptsize Subtraction}}
		\put(72.4,3){\textbf{\scriptsize Sigmoid}}
	\end{overpic}
    \vspace{-5mm}
    \captionsetup{font={small}}
    \caption{Structure of the second stage subnetwork ($G_T$) in RAGNet, which takes both the observation $I$ and the predicted reflection $\hat{R}$ as input, and generates the transmission layer $\hat{T}$. $G_T$ follows the U-Net~\cite{UNet} setting, while the concatenated feature through the skip-connection is $\mathbf{F}_\mathit{diff}=\mathbf{F}_I-\mathbf{F}_R$ rather than $\mathbf{F}_I$. More detailed structure settings please refer to the supplementary material.}
    \label{fig:architecture}
\end{figure*}

In this section, we first revisit the limitations of existing multi-stage SIRR methods.
Then, the overall architecture and the design to exploit reflection-aware guidance are illustrated in detail.
Finally, the dedicated mask loss and the learning objective are presented.


\subsection{Limitations of existing multi-stage methods}\label{subsec:limitations}
Given an observation $I$, CEILNet~\cite{CEILNet} predicts a potential edge map $E$ for the transmission layer, where a reflection smoothness prior is assumed to distinguish whether the image gradients belong to the transmission layer or the reflection layer.
Unfortunately, such smoothness prior is often violated in real scenarios, and the resulting inaccurate edge maps will lead to a degraded performance by providing wrong information to the second stage.

BDN~\cite{BDN} and IBCLN~\cite{IBCLN} resort to predicting reflection layer in the prior stage.
However, a simple concatenation is insufficient to make full use of the predicted reflection layer and is limited in suppressing the heavy and bright reflections, thus BDN~\cite{BDN} often fails to process the reflections in difficult regions (see Fig.~\ref{fig:introduction}).
IBCLN~\cite{IBCLN} takes the input image $I$ and black image $\mathbf{0}_{h\times w}$ as initial estimation of transmission and reflection, and iteratively refines them with a LSTM~\cite{LSTM} module to exploit cross-stage information, which significantly promotes the SIRR performance.
Nevertheless, such principle makes the method less efficient and greatly relies on the generation quality in the prior stages, and the errors will also be accumulated across stages, leading to undesired artifacts in some scenarios (\eg, zooming in Fig.~\ref{fig:introduction} for clear observation).

\subsection{Overall Architecture Design}\label{subsec:structure}
Considering the aforementioned limitations, we build a \emph{two-stage} model, where a \emph{reflection-aware guidance (RAG)} module is introduced in the second stage to better exploit the \emph{reflection layer} predicted in the first stage.
Specifically, we use a plain U-Net~\cite{UNet} model as the first stage subnetwork, namely $G_R$, which takes the observation $I$ as input and predicts the reflection layer, \ie, $\hat{R}=G_R(I)$.
The structure of the second stage subnetwork $G_T$ is shown in Fig.~\ref{fig:architecture}, $\hat{R}$ is fed into $G_T$ together with the observation $I$ to generate the transmission layer, \ie, $\hat{T}=G_T(I, \hat{R})$.
In particular, $G_T$ is composed of two encoders ($E_I$ and $E_R$) and one decoder ($D_T$), and an RAG module is elaborated in each block of $D_T$.
Kindly refer to the supplementary material for detailed structure configurations.

\begin{figure}[th]
    \small
    \centering
    \scalebox{.93}{
        \begin{tabular}{cc}
            \includegraphics[width=0.22\textwidth]{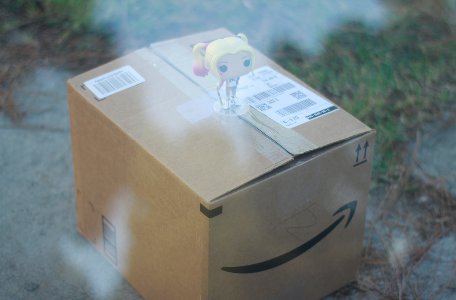}&\hspace{-4.2mm}
            \includegraphics[width=0.22\textwidth]{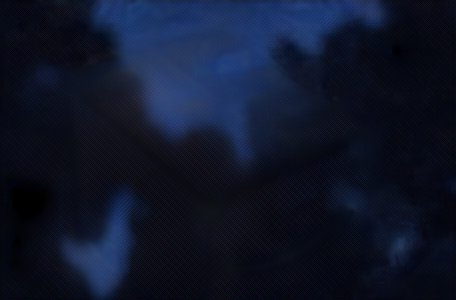}
            \\
            Input&\hspace{-4.2mm}
            Predicted $\hat{R}$
            \\
            \includegraphics[width=0.22\textwidth]{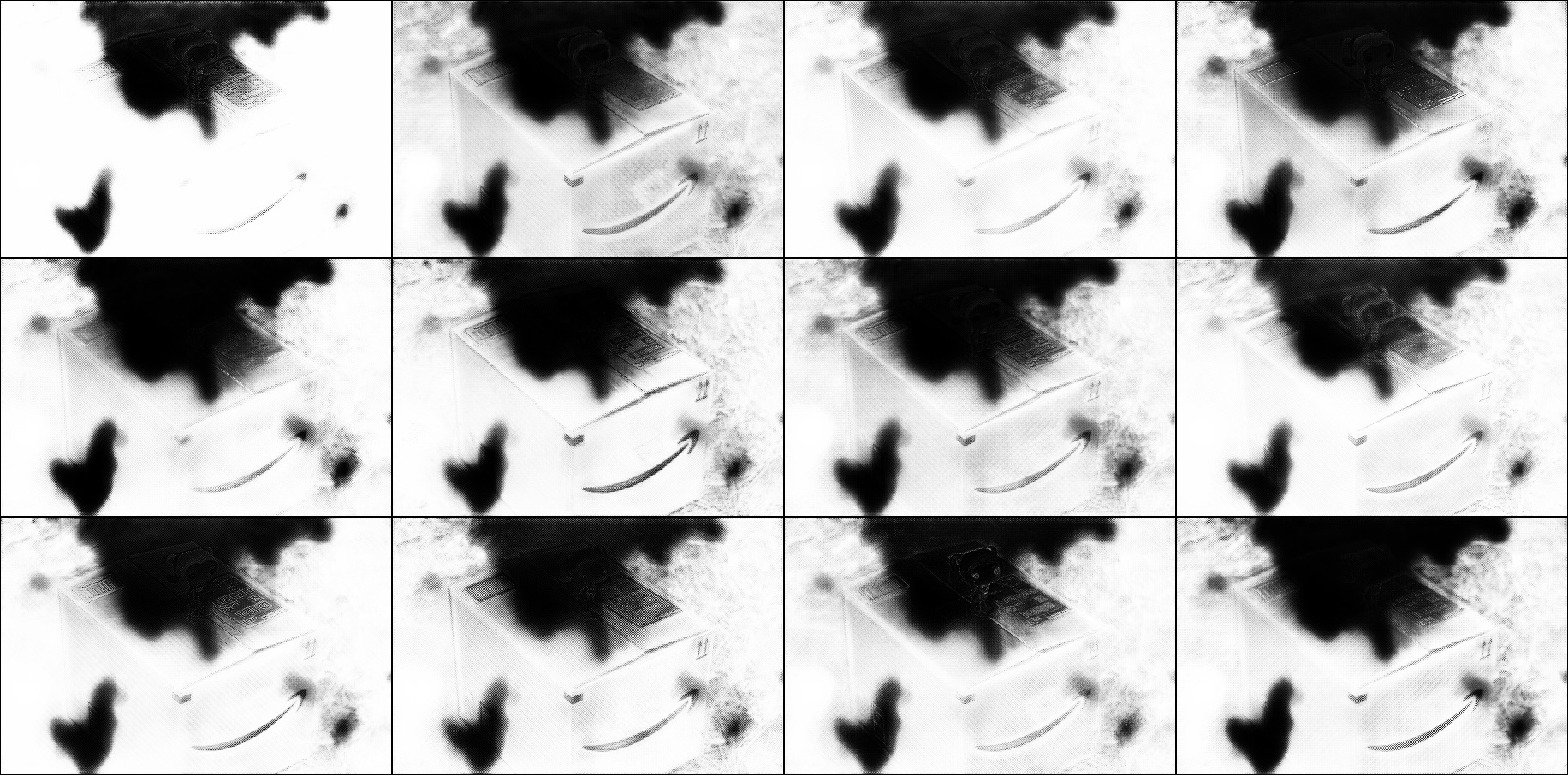}&\hspace{-4.2mm}
            \includegraphics[width=0.22\textwidth]{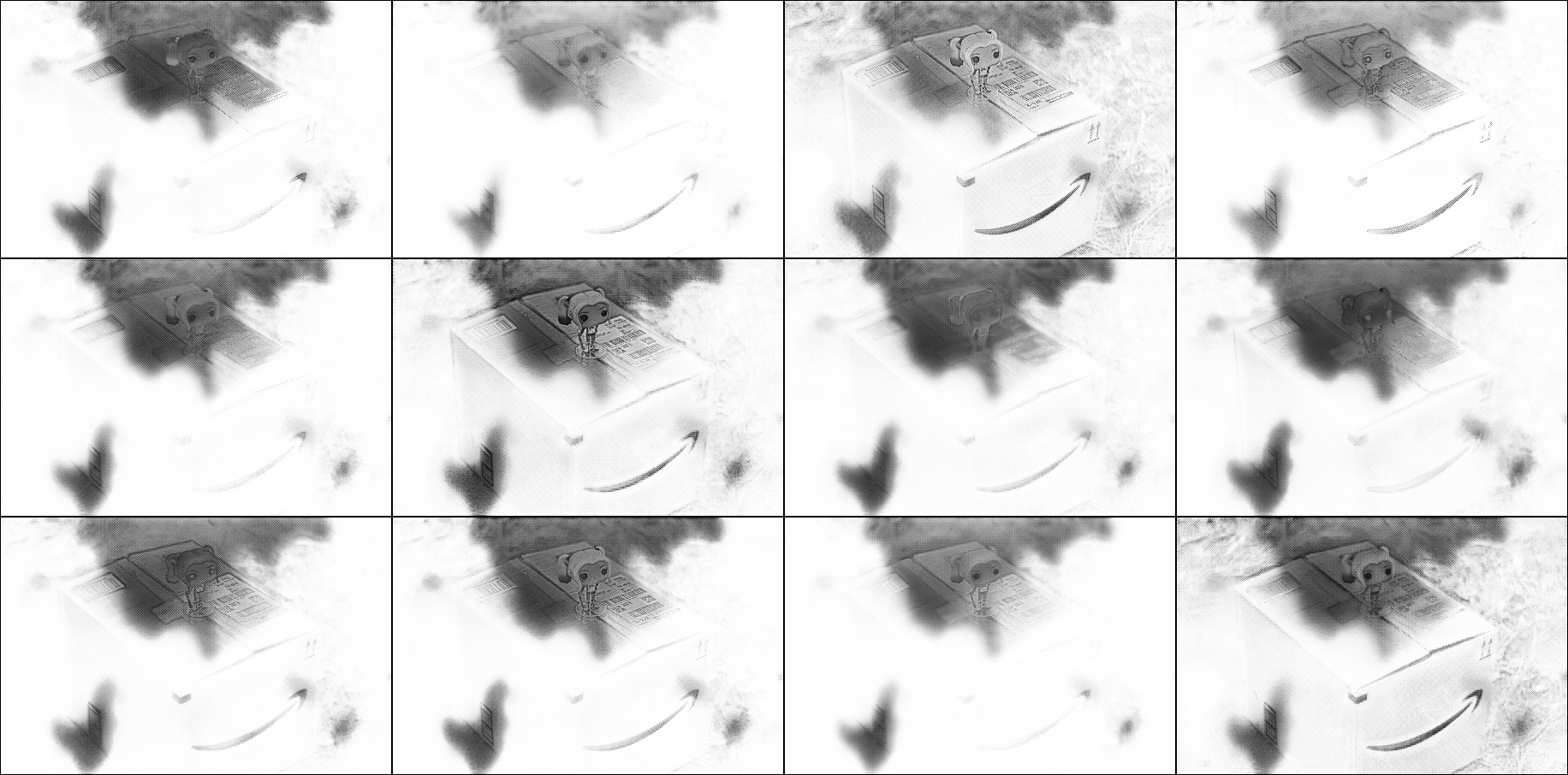}
            \\
            $\mathbf{M}_\mathit{diff}$&\hspace{-4.2mm}
            $\mathbf{M}_\mathit{dec}$
            \\
        \end{tabular}}
    \vspace{-2mm}
    \captionsetup{font={small}}
    \caption{Visualization of $\mathbf{M}_\mathit{diff}$ and $\mathbf{M}_\mathit{dec}$, and we randomly select $12$ channels for each. It can be seen that the values of $\mathbf{M}_\mathit{diff}$ tend to be $0$ (black in color) in heavy reflection areas, while the values in regions with minor reflections are close to $1$. The image is from the \emph{Real 20} Dataset.}
    \label{fig:mask}
    \vspace{-3mm}
\end{figure}

\begin{figure*}[thbp]
    \small
    \centering
    \scalebox{.95}{
        \begin{tabular}{cccccccc}
            \includegraphics[width=0.124\textwidth]{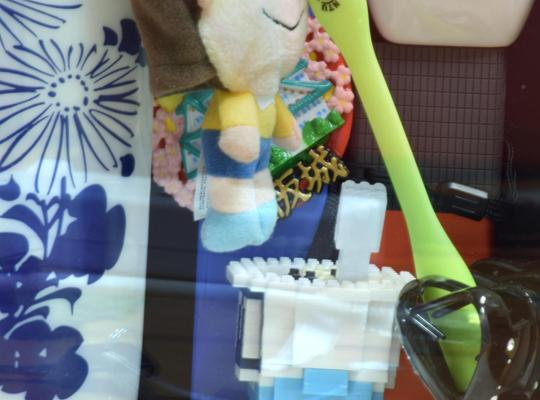}&\hspace{-4.2mm}
            \includegraphics[width=0.124\textwidth]{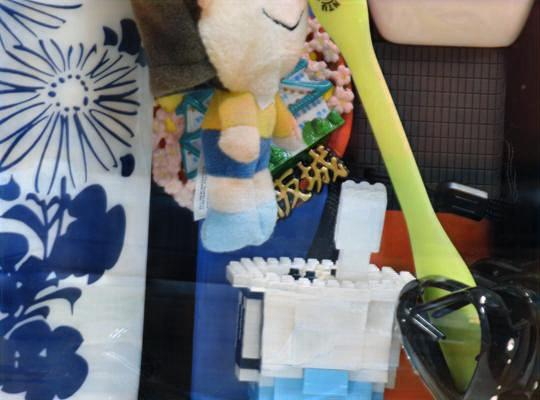}&\hspace{-4.2mm}
            \includegraphics[width=0.124\textwidth]{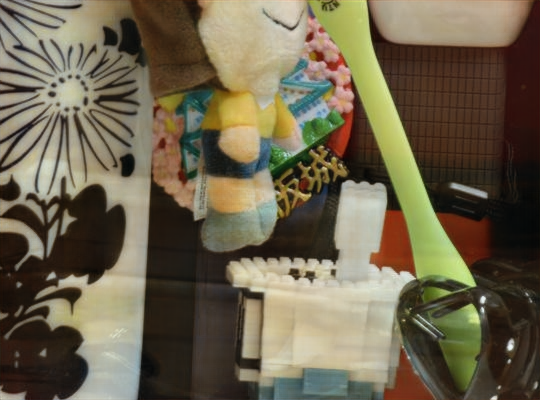}&\hspace{-4.2mm}
            \includegraphics[width=0.124\textwidth]{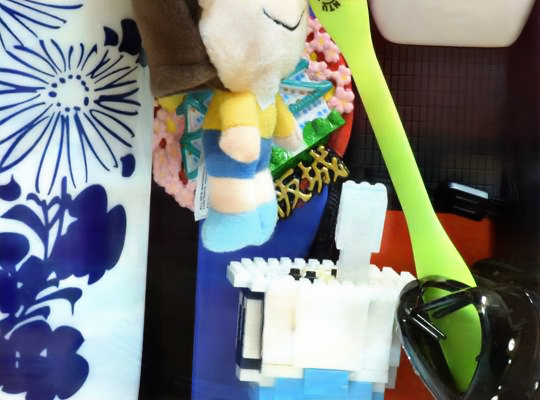}&\hspace{-4.2mm}
            \includegraphics[width=0.124\textwidth]{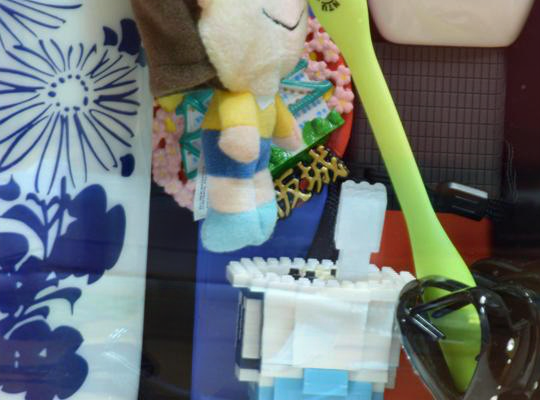}&\hspace{-4.2mm}
            \includegraphics[width=0.124\textwidth]{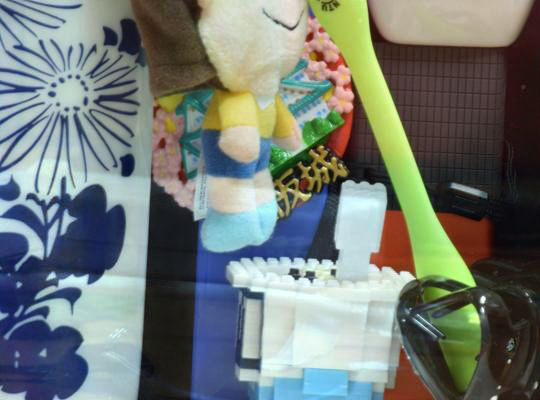}&\hspace{-4.2mm}
            \includegraphics[width=0.124\textwidth]{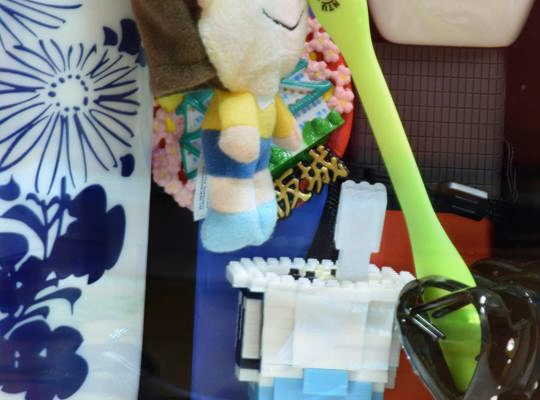}&\hspace{-4.2mm}
            \includegraphics[width=0.124\textwidth]{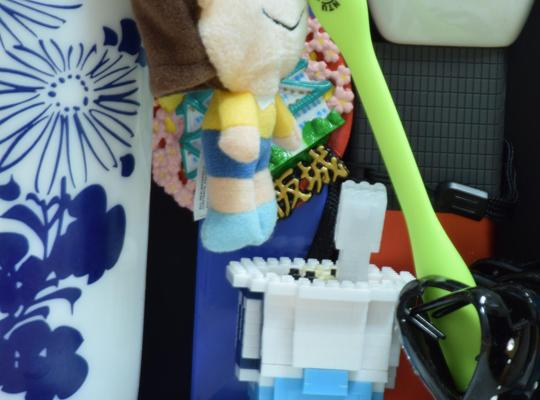}
            \\
            \includegraphics[width=0.124\textwidth]{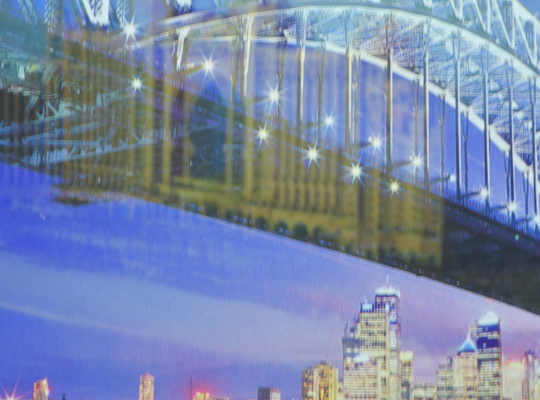}&\hspace{-4.2mm}
            \includegraphics[width=0.124\textwidth]{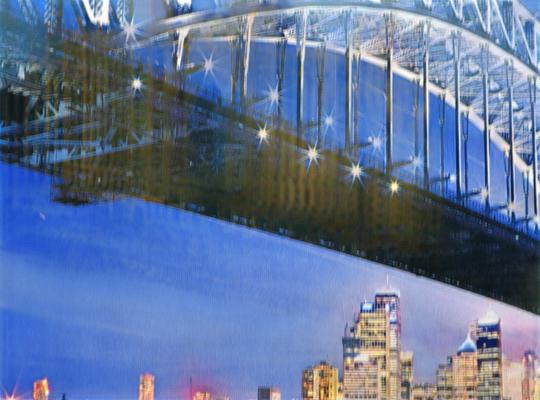}&\hspace{-4.2mm}
            \includegraphics[width=0.124\textwidth]{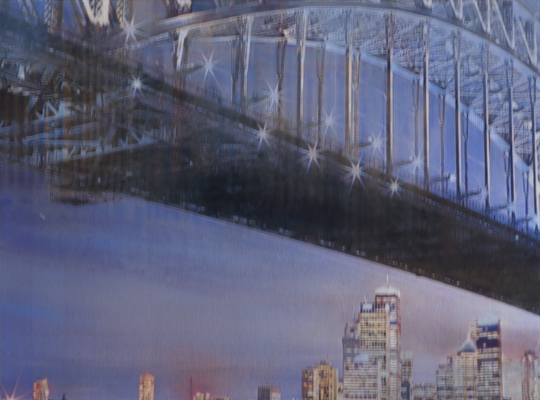}&\hspace{-4.2mm}
            \includegraphics[width=0.124\textwidth]{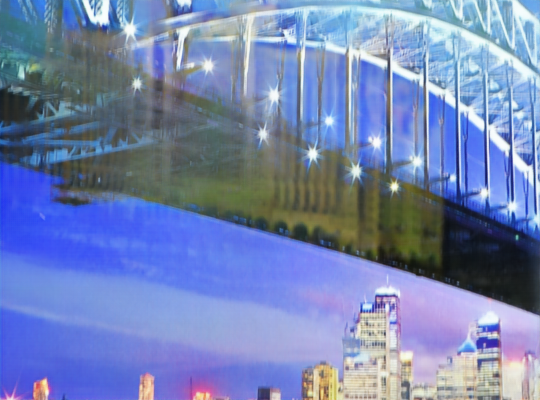}&\hspace{-4.2mm}
            \includegraphics[width=0.124\textwidth]{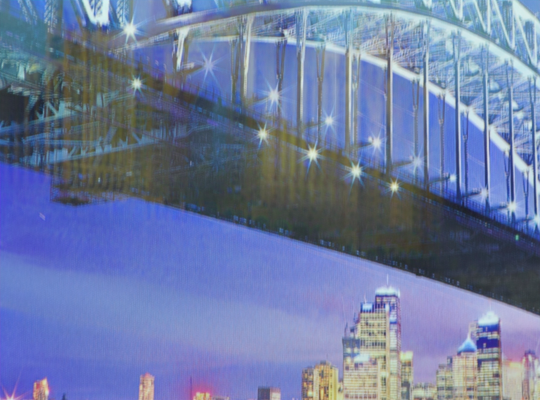}&\hspace{-4.2mm}
            \includegraphics[width=0.124\textwidth]{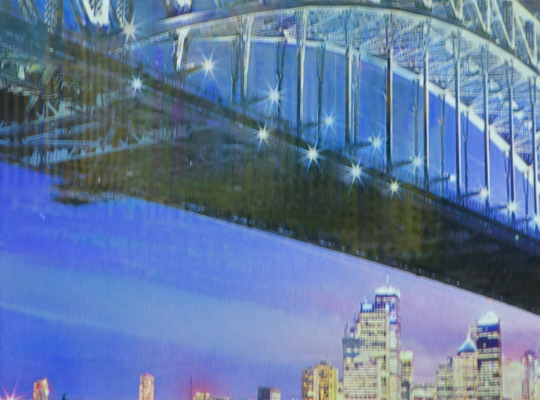}&\hspace{-4.2mm}
            \includegraphics[width=0.124\textwidth]{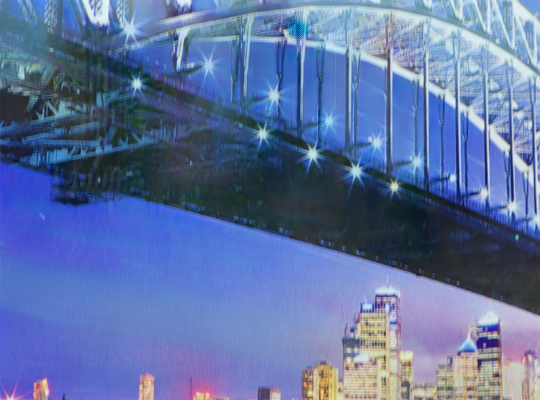}&\hspace{-4.2mm}
            \includegraphics[width=0.124\textwidth]{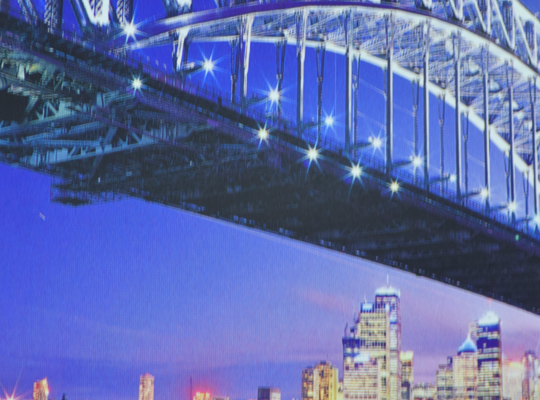}
            \\
            \includegraphics[width=0.124\textwidth]{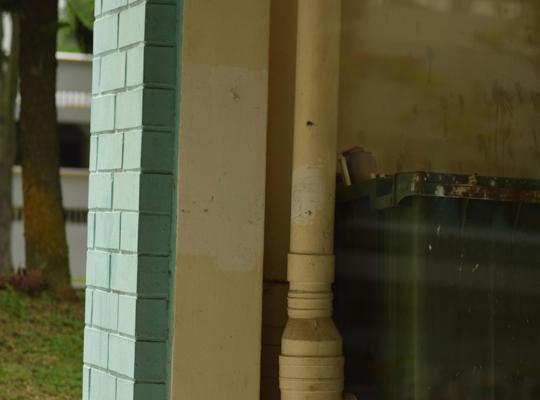}&\hspace{-4.2mm}
            \includegraphics[width=0.124\textwidth]{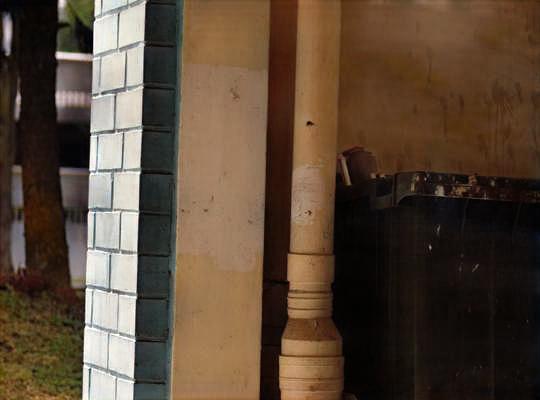}&\hspace{-4.2mm}
            \includegraphics[width=0.124\textwidth]{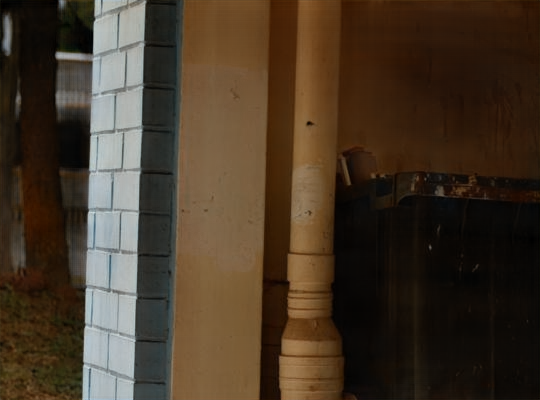}&\hspace{-4.2mm}
            \includegraphics[width=0.124\textwidth]{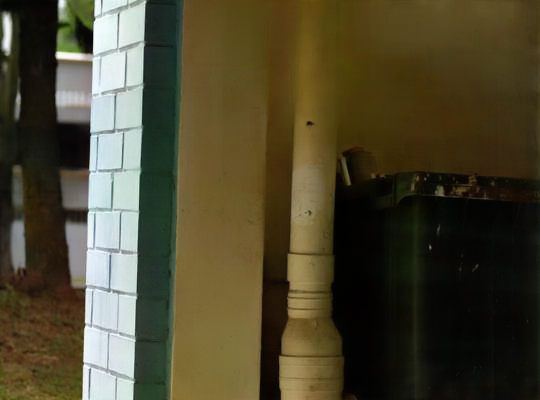}&\hspace{-4.2mm}
            \includegraphics[width=0.124\textwidth]{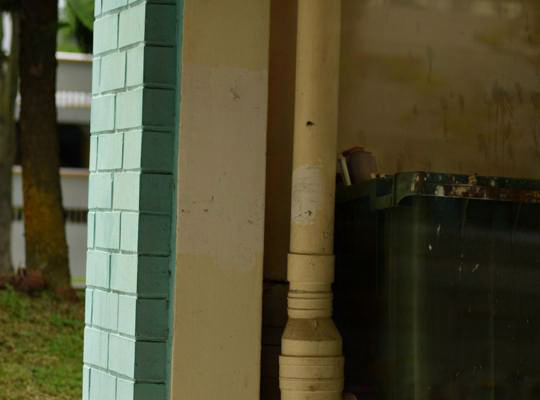}&\hspace{-4.2mm}
            \includegraphics[width=0.124\textwidth]{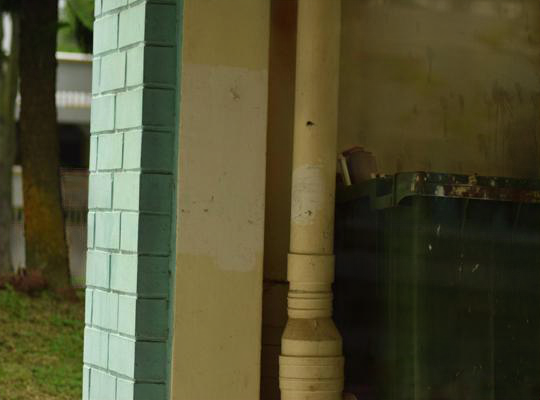}&\hspace{-4.2mm}
            \includegraphics[width=0.124\textwidth]{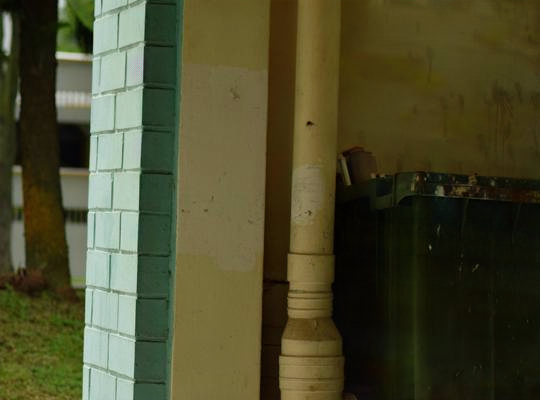}&\hspace{-4.2mm}
            \includegraphics[width=0.124\textwidth]{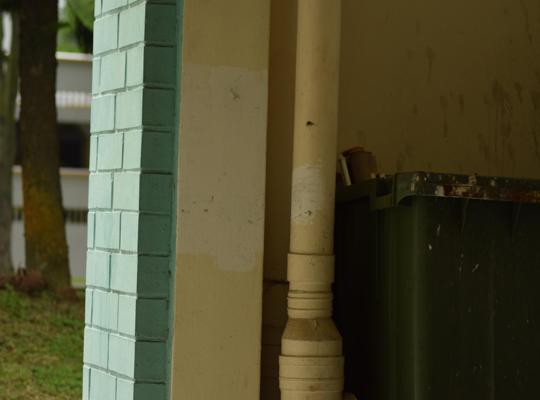}
            \\
            \includegraphics[width=0.124\textwidth]{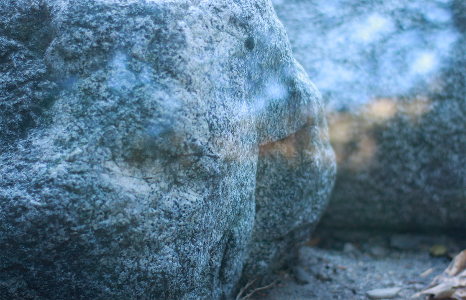}&\hspace{-4.2mm}
            \includegraphics[width=0.124\textwidth]{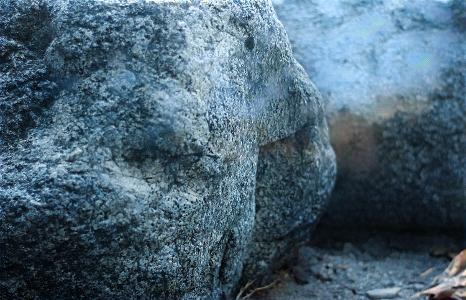}&\hspace{-4.2mm}
            \includegraphics[width=0.124\textwidth]{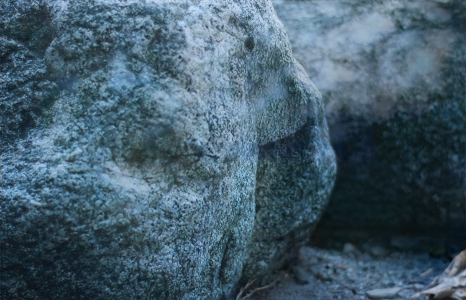}&\hspace{-4.2mm}
            \includegraphics[width=0.124\textwidth]{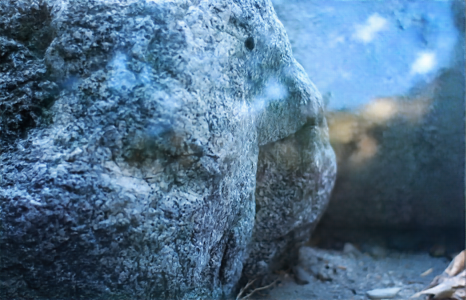}&\hspace{-4.2mm}
            \includegraphics[width=0.124\textwidth]{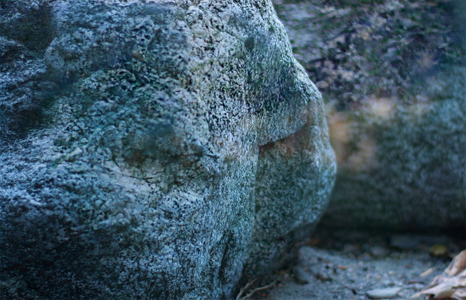}&\hspace{-4.2mm}
            \includegraphics[width=0.124\textwidth]{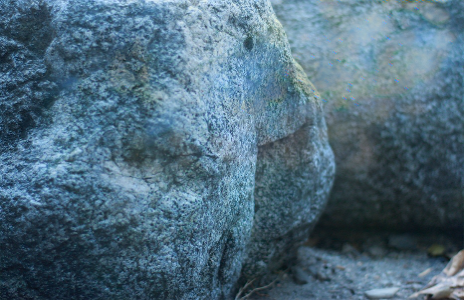}&\hspace{-4.2mm}
            \includegraphics[width=0.124\textwidth]{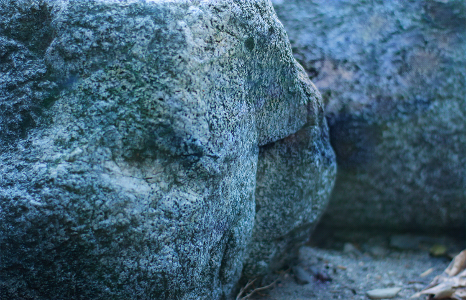}&\hspace{-4.2mm}
            \includegraphics[width=0.124\textwidth]{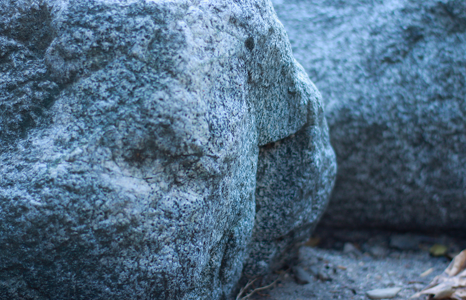}
            \\
            Input&\hspace{-4.2mm}
            CEILNet~\cite{CEILNet}&\hspace{-4.2mm}
            Zhang \etal~\cite{Zhang2018CVPR}&\hspace{-4.2mm}
            BDN~\cite{BDN}& \hspace{-4.2mm}
            ERRNet~\cite{ERRNet}& \hspace{-4.2mm}
            IBCLN~\cite{IBCLN}& \hspace{-4.2mm}
            Ours& \hspace{-4.2mm}
            Ground-truth
            \\
        \end{tabular}}
    \vspace{-3mm}
    \captionsetup{font={small}}
    \caption{Visual comparison on four real-world datasets. The first column and the last column are the inputs and  ground-truth transmission layers. From top to down, the four rows of images are from \emph{SIR$^2$ Solid}, \emph{SIR$^2$ Postcard}, \emph{SIR$^2$ Wild} and \emph{Real 20} datasets, respectively. More results are given in the supplementary material, and please zoom in for better observation.}
    \label{fig:comparison0}
\end{figure*}

\subsection{Reflection-Aware Guidance}\label{subsec:RAG}
As shown in Fig.~\ref{fig:introduction}, reflections are spatially nonuniform, whose location and intensity can be roughly predicted in the first stage.
The estimated reflection $\hat{R}$ can then be incorporated with the observation $I$ for succeeding transmission estimation.
However, the estimated reflection $\hat{R}$ may be inaccurate.
Furthermore, the linear combination hypothesis may not hold true in regions with high reflection intensities, where few informative features of the transmission are retained in $I-\hat{R}$ (see Fig.~\ref{fig:introduction}).
%
%
Therefore, estimating the transmission in such regions is more challenging, and is more likely to be an inpainting task.
Taking these factors into account, we elaborate an RAG module in each decoder block  for (i) generating feature complementary to the decoder and (ii) predicting a mask to indicate the heavy reflection regions for guiding image inpainting.


Given the observation feature $\mathbf{F}_I$, the reflection feature $\mathbf{F}_R$ and the decoder feature $\mathbf{F}_\mathit{dec}$, the RAG module is designed as shown in Fig.~\ref{fig:architecture}.
Using $\mathbf{F}_I$ and $\mathbf{F}_R$, the difference feature $\mathbf{F}_\mathit{diff}$ is generated via a subtraction operation to better suppress the reflections, \ie, $\mathbf{F}_\mathit{diff} = \mathbf{F}_I - \mathbf{F}_R$, which serves as a complement and enhancement to the decoder feature $\mathbf{F}_\mathit{dec}$.
Furthermore, taking $\mathbf{F}_I$, $\mathbf{F}_R$ and $\mathbf{F}_\mathit{dec}$ as input, the RAG module generates a mask $\mathbf{M}$ via two $1 \times 1$ convolution layers followed by a sigmoid operation.
Note that $\mathbf{M}=[\mathbf{M}_\mathit{diff}, \mathbf{M}_\mathit{dec}]$, where $[\cdot,\cdot]$ denotes the concatenation operation, and $\mathbf{M}_\mathit{diff}$ and $\mathbf{M}_\mathit{dec}$ are masks regarding to $\mathbf{F}_\mathit{diff}$ and $\mathbf{F}_\mathit{dec}$, respectively.
With such mask, the model is aware of the regions deviating from the linear combination hypothesis, and recovers the transmission layer in such areas mainly relying on the surrounding decoder feature using image inpainting.
The mask $\mathbf{M}$ is visualized in Fig.~\ref{fig:mask}.

Considering the similarity between inpainting and reflection removal in heavy reflection regions, we deploy a partial convolution~\cite{partial_conv} alongside each RAG module to better exploit the mask, which was first introduced in image inpainting methods~\cite{Liu_2018_ECCV,xie2019image}.
The mask generated by RAG module ranges from 0 to 1, thus the partial convolution in this paper can be formulated as,
\begin{equation}
\label{eqn:partial_conv}
\resizebox*{.9\linewidth}{!}{
	$\mathbf{F}'= \left \{
	\begin{array}{cl}
	\displaystyle (\mathbf{W}\ast(\mathbf{F}\circ\mathbf{M}))\circ\frac{1}{\mathbf{\bar{M}}_{3\times3}}+\mathbf{b}, & \mathbf{\bar{M}}_{3\times3} > 0 \\
	0, & \mathrm{otherwise}
	\end{array}\right.$
	\!,}
\end{equation}
where $\mathbf{W}$ and $\mathbf{b}$ represent the weight and bias of the partial convolution, $\mathbf{F}=[\mathbf{F}_\mathit{diff}, \mathbf{F}_\mathit{dec}]$, $\ast$ and $\circ$ are convolution and entry-wise product respectively. $\mathbf{\bar{M}}_{3\times3}$ contains the average values of $\mathbf{M}$ in $3\!\times\!3$ neighborhood regions, which can be efficiently calculated by a $3\!\times\!3$ average pooling operation.

\begin{figure*}[thbp]
	\small
	\centering
	\scalebox{.95}{
		\begin{tabular}{ccccccc}
			\includegraphics[width=0.14\textwidth]{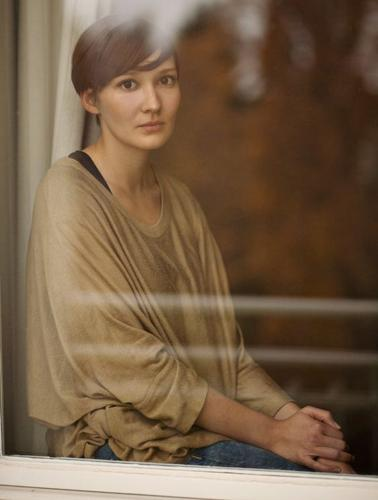}&\hspace{-4.2mm}
			\includegraphics[width=0.14\textwidth]{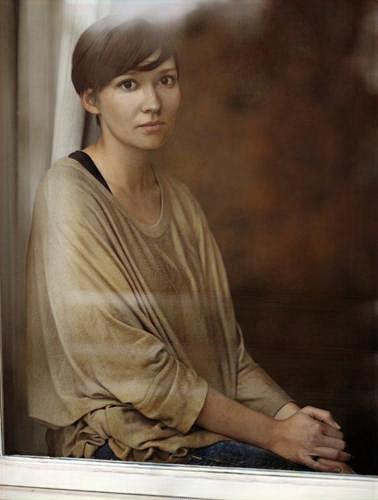}&\hspace{-4.2mm}
			\includegraphics[width=0.14\textwidth]{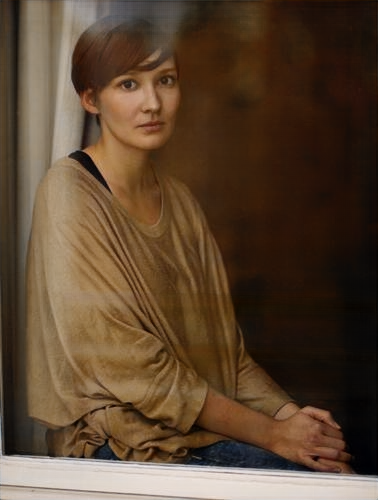}&\hspace{-4.2mm}
			\includegraphics[width=0.14\textwidth]{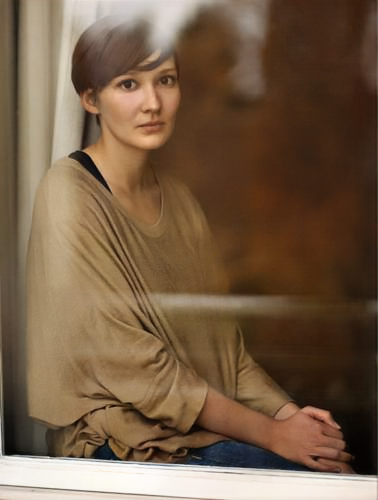}&\hspace{-4.2mm}
			\includegraphics[width=0.14\textwidth]{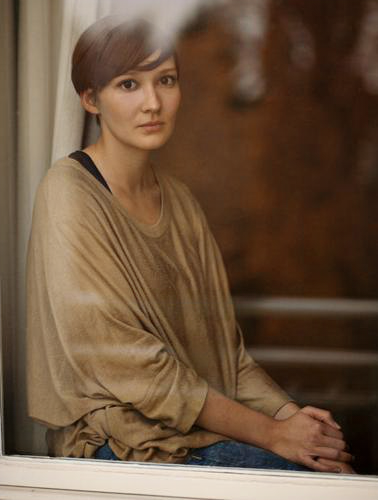}&\hspace{-4.2mm}
			\includegraphics[width=0.14\textwidth]{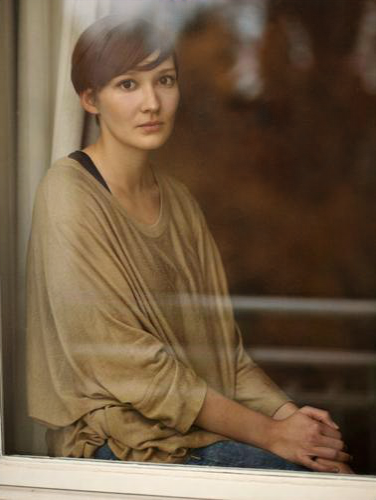}&\hspace{-4.2mm}
			\includegraphics[width=0.14\textwidth]{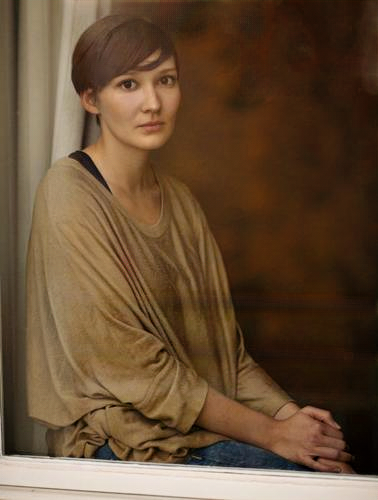}
			\\
			\includegraphics[width=0.14\textwidth]{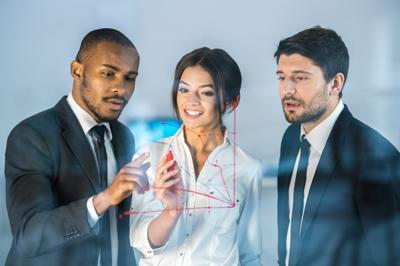}&\hspace{-4.2mm}
			\includegraphics[width=0.14\textwidth]{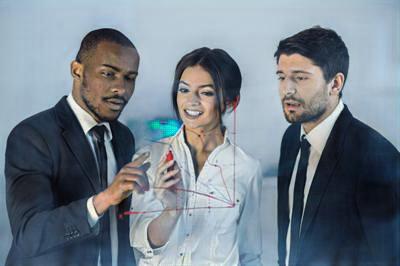}&\hspace{-4.2mm}
			\includegraphics[width=0.14\textwidth]{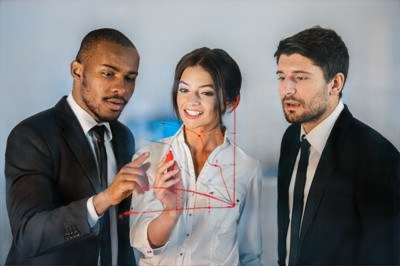}&\hspace{-4.2mm}
			\includegraphics[width=0.14\textwidth]{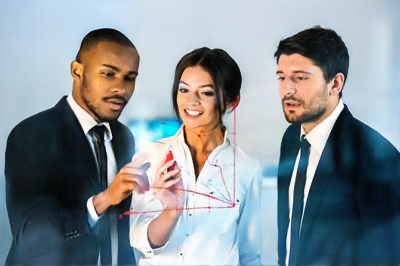}&\hspace{-4.2mm}
			\includegraphics[width=0.14\textwidth]{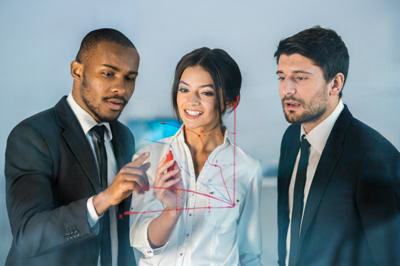}&\hspace{-4.2mm}
			\includegraphics[width=0.14\textwidth]{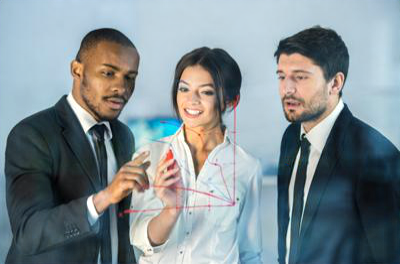}&\hspace{-4.2mm}
			\includegraphics[width=0.14\textwidth]{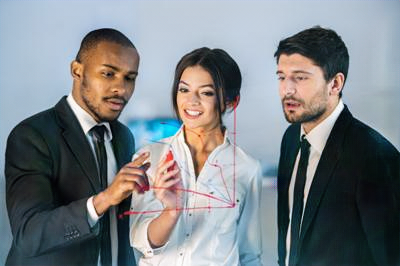}
			\\
			\includegraphics[width=0.14\textwidth]{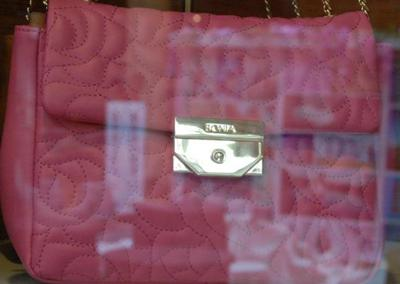}&\hspace{-4.2mm}
			\includegraphics[width=0.14\textwidth]{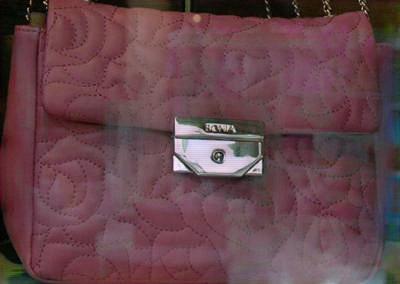}&\hspace{-4.2mm}
			\includegraphics[width=0.14\textwidth]{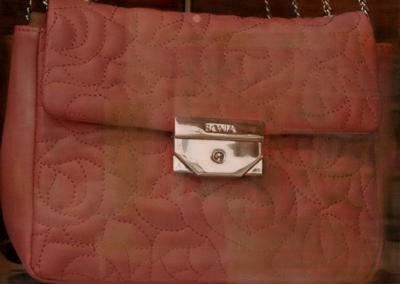}&\hspace{-4.2mm}
			\includegraphics[width=0.14\textwidth]{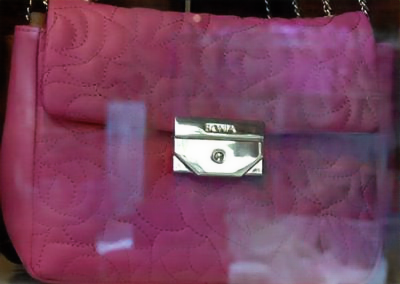}&\hspace{-4.2mm}
			\includegraphics[width=0.14\textwidth]{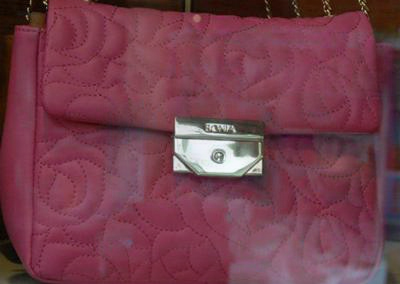}&\hspace{-4.2mm}
			\includegraphics[width=0.14\textwidth]{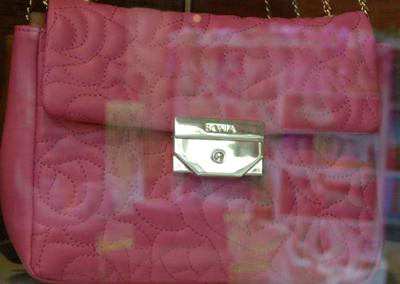}&\hspace{-4.2mm}
			\includegraphics[width=0.14\textwidth]{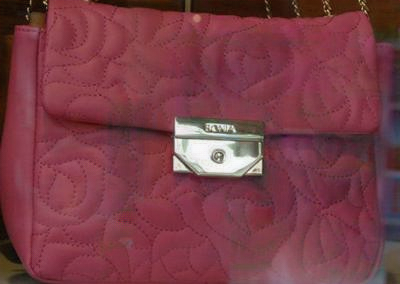}
			\\
			Input&\hspace{-4.2mm}
			CEILNet~\cite{CEILNet}&\hspace{-4.2mm}
			Zhang \etal~\cite{Zhang2018CVPR}&\hspace{-4.2mm}
			BDN~\cite{BDN}& \hspace{-4.2mm}
			ERRNet~\cite{ERRNet}& \hspace{-4.2mm}
			IBCLN~\cite{IBCLN}& \hspace{-4.2mm}
			Ours
			\\
	\end{tabular}}
	\vspace{-3mm}
	\captionsetup{font={small}}
	\caption{Visual comparison on \emph{Real 45} dataset, where the ground-truth transmission layer is unavailable. More results are given in the supplementary material. Please zoom in for better observation.}
	\label{fig:comparison1}
\end{figure*}

\begin{table*}[thbp]
	\centering
	\captionsetup{font={small}}
	\caption{PSNR/SSIM results by different methods for reflection removal on four real-world datasets with ground-truth. The best and the second-best results in each dataset are highlighted with {\color{red}red} and {\color{blue}blue}, respectively.}
	\vspace{-3mm}
	\label{tab:comparison}
	\scalebox{0.85}{
		\begin{tabular}{lccccccc}
			\toprule
			\multirow{2}{*}{Datasets}  &  CEILNet~\cite{CEILNet} & Zhang \etal~\cite{Zhang2018CVPR} & BDN~\cite{BDN} &  ERRNet~\cite{ERRNet} & Lei \etal~\cite{Lei_2020_CVPR} & IBCLN~\cite{IBCLN} & Ours   \\
			\cmidrule(l){2-8}
			{} & PSNR / SSIM & PSNR / SSIM  & PSNR / SSIM & PSNR / SSIM & PSNR / SSIM  & PSNR / SSIM & PSNR / SSIM  \\
			\midrule
			{\emph{SIR$^2$ Solid} (200)} &23.37 / 0.875 & 22.68 / 0.879 & 22.73 / 0.853 & 24.85 / {\color{blue}0.894} & 23.81 / 0.882 & {\color{blue}24.88} / 0.893 & {\color{red}26.15} / {\color{red}0.903}           \\
			\emph{SIR$^2$ Postcard} (199)& 20.09 / 0.786 & 16.81 / 0.797 & 20.71 / 0.857 & 21.99 / 0.874 & 21.48 / 0.873 & {\color{blue}23.39} / {\color{blue}0.875} & {\color{red}23.67} / {\color{red}0.879}          \\
			\emph{SIR$^2$ Wild} (55) & 21.07 / 0.805 & 21.52 / 0.832 & 22.34 / 0.821 & 24.16 / 0.847 & 23.84 / 0.866 & {\color{blue}24.71} / {\color{red}0.886} & {\color{red}25.52} / {\color{blue}0.880}          \\
			\emph{Real 20} (20) & 18.87 / 0.692 & 22.55 / 0.788 & 18.81 / 0.737 & {\color{red}23.19} / {\color{red}0.817} & 22.35 / {\color{blue}0.793} & 22.04 / 0.772 & {\color{blue}22.95} / {\color{blue}0.793}         \\
			Average (474) & 21.54 / 0.822 & 20.08 / 0.835& 21.67 / 0.846 & 23.50 / 0.877 & 22.77 / 0.873 & {\color{blue}24.11} / {\color{blue}0.880} & {\color{red}24.90} / {\color{red}0.886}         \\
			\bottomrule
	\end{tabular}}
\end{table*}

\subsection{Mask Loss}\label{subsec:mask_loss}
Furthermore, we design a novel mask loss to better exploit the mask $\mathbf{M}$ during training RAGNet.
As shown in Fig.~\ref{fig:introduction}, for areas with heavy reflections, $\mathbf{F}_\mathit{diff}$ can hardly provide informative features.
Therefore, we dispose the values of $\mathbf{M}_\mathit{diff}$ to approximate $0$ in such areas, \ie,
\begin{equation}
\label{eqn:mask_loss_diff}
\mathcal{L}_\mathrm{mask}^\mathit{diff}=\sum\limits_{i=1}^4\| \mathbf{M}_\mathit{diff}^i[R>\varphi]\|_1,
\end{equation}%
where $i$ means the $i$-th layer, $\|\cdot\|_{_1}$ denotes the $\ell_1$ norm, $A[condition]$ represents the part of $A$ that meets the $condition$ in the square brackets, $\varphi$ is the threshold that delimits heavy reflection areas.

However, with $\mathcal{L}_\mathrm{mask}^\mathit{diff}$ only, $\mathbf{M}$ may be optimized towards a trivial solution $\mathbf{0}$, which diminishes the effect of $\mathbf{F}_\mathit{diff}$.
Therefore, $\mathbf{M}$ is supposed to be $1$ for areas with few reflections, to avoid the trivial solution as a regularization term and force the partial convolution to exploit both $\mathbf{F}_\mathit{diff}$ and $\mathbf{F}_\mathit{dec}$, since both of them are reliable in such areas, \ie,
\begin{equation}
\label{eqn:mask_loss_reg}
\mathcal{L}_\mathrm{mask}^\mathit{reg}=\sum\limits_{i=1}^4\| \mathbf{M}^i[R<\xi] -1 \|_1,
\end{equation}%
where $\xi$ means threshold for regions with few reflections.

Note that we do not constrain the mask values in remaining regions, which will be automatically optimized for better transmission estimation.
Therefore, the mask loss is formulated as,
\begin{equation}
\label{eqn:mask_loss}
\mathcal{L}_\mathrm{mask}=\mathcal{L}_\mathrm{mask}^\mathit{diff}+\mathcal{L}_\mathrm{mask}^\mathit{reg},
\end{equation}%
and we empirically set $\xi=0.01$ and $\varphi=0.3$.
%
The effectiveness of mask loss will be further corroborated in ablation studies.

\subsection{Learning Objective}\label{subsec:learning_objectives}
%

\begin{figure*}[thbp]
	\small
	\centering
	\scalebox{.95}{
		\begin{tabular}{cccccccc}
			\includegraphics[width=0.14\textwidth]{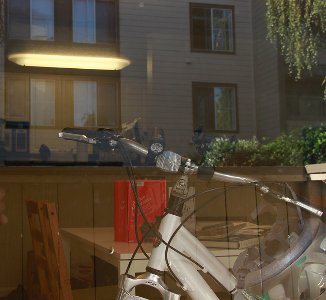}&\hspace{-4.2mm}
			\includegraphics[width=0.14\textwidth]{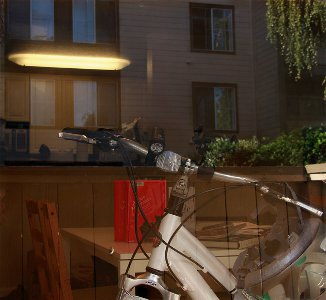}&\hspace{-4.2mm}
			\includegraphics[width=0.14\textwidth]{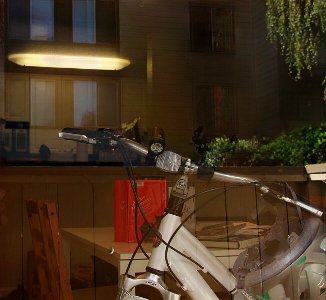}&\hspace{-4.2mm}
			\includegraphics[width=0.14\textwidth]{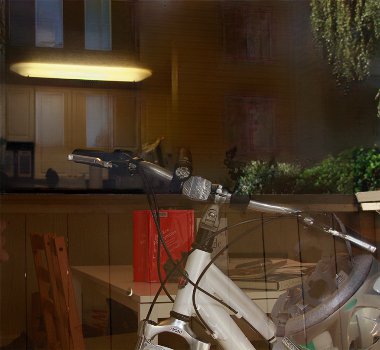}&\hspace{-4.2mm}
			\includegraphics[width=0.14\textwidth]{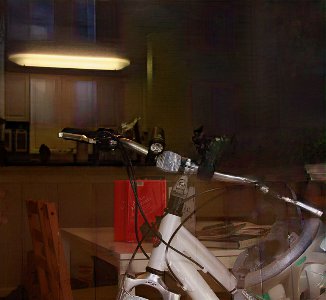}&\hspace{-4.2mm}
			\includegraphics[width=0.14\textwidth]{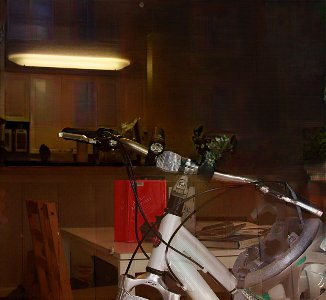}&\hspace{-4.2mm}
			\includegraphics[width=0.14\textwidth]{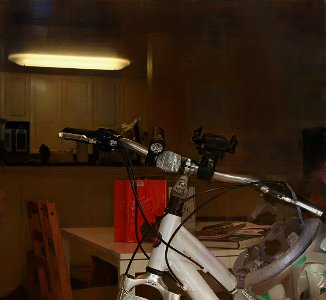}
			\\
			\includegraphics[width=0.14\textwidth]{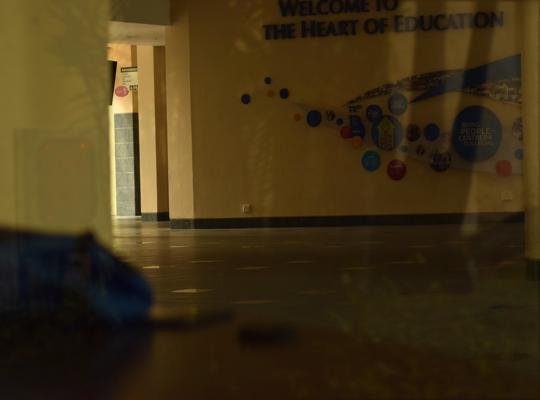}&\hspace{-4.2mm}
			\includegraphics[width=0.14\textwidth]{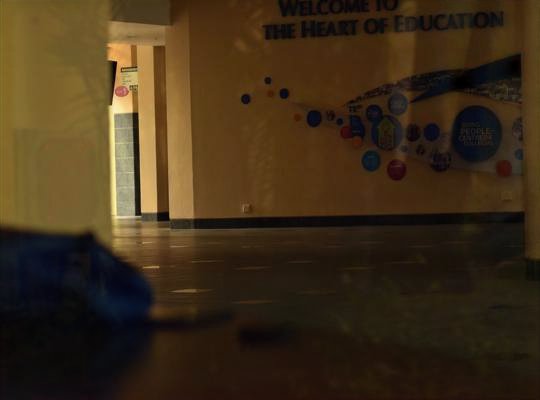}&\hspace{-4.2mm}
			\includegraphics[width=0.14\textwidth]{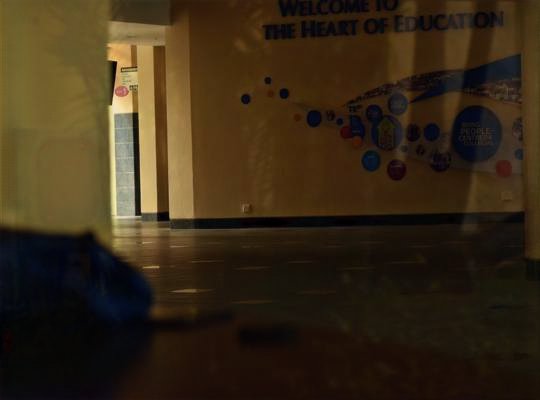}&\hspace{-4.2mm}
			\includegraphics[width=0.14\textwidth]{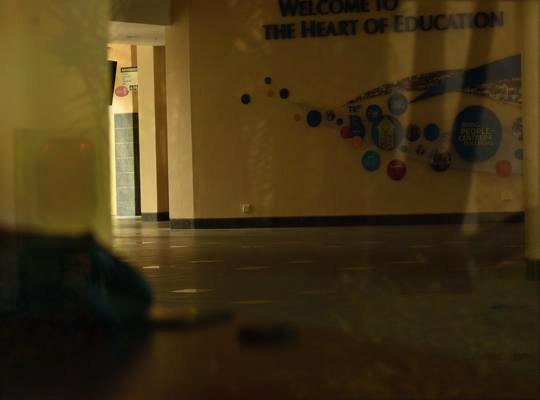}&\hspace{-4.2mm}
			\includegraphics[width=0.14\textwidth]{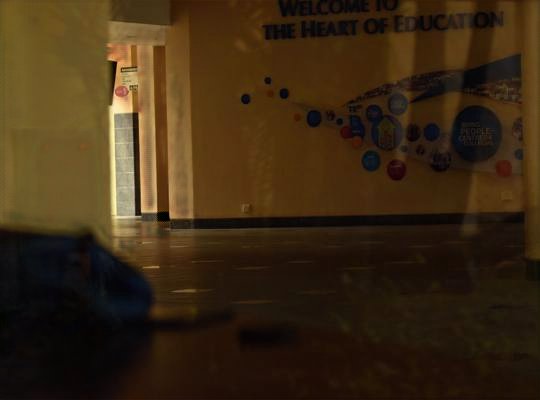}&\hspace{-4.2mm}
			\includegraphics[width=0.14\textwidth]{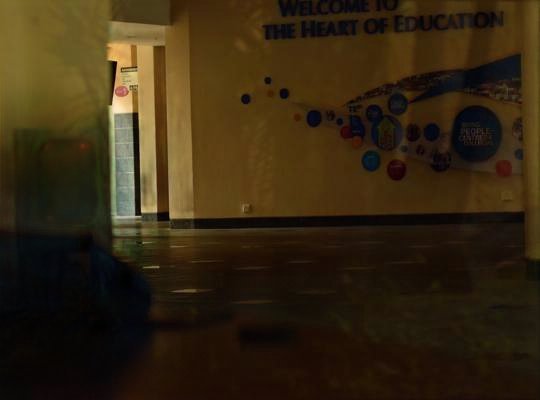}&\hspace{-4.2mm}
			\includegraphics[width=0.14\textwidth]{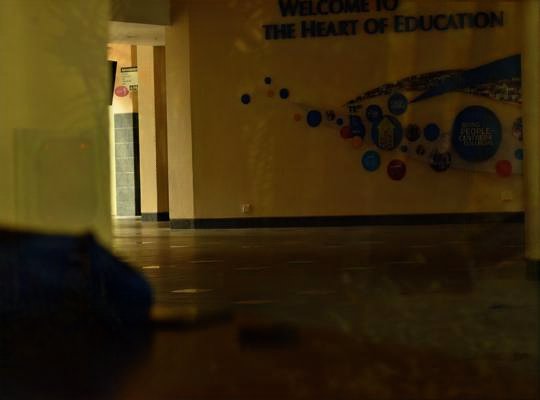}
			\\
			Input&\hspace{-4.2mm}
			RAGNet$_{I \to T}$&\hspace{-4.2mm}
			w/o $\mathbf{F}_\mathit{diff}$& \hspace{-4.2mm}
			w/o $\mathcal{L}_\mathrm{mask}$& \hspace{-4.2mm}
			w/o $\mathcal{L}_\mathrm{mask}^\mathit{diff}$& \hspace{-4.2mm}
			w/o $\mathcal{L}_\mathrm{mask}^\mathit{reg}$& \hspace{-4.2mm}
			RAGNet
			\\
	\end{tabular}}
	\vspace{-3mm}
	\captionsetup{font={small}}
	\caption{Visual results of ablation studies on structure and loss functions. The images are from \emph{Real 20} and \emph{SIR$^2$ Wild} datasets, respectively. Please zoom in for better observation.}
	\label{fig:ablation_structure}
	\vspace{-2mm}
\end{figure*}

To train our RAGNet, several commonly used loss functions are collaborated, including reconstruction loss, perceptual loss~\cite{LiFeiFei_styletransfer}, exclusion loss~\cite{Zhang2018CVPR} and adversarial loss~\cite{GAN}.
%

\noindent\textbf{Reconstruction loss.}
With synthetic image pairs, we are able to minimize the pixel-wise difference between network outputs $\hat{T}$, $\hat{R}$ and the corresponding ground-truths $T$, $R$,

\begin{equation}
\label{eqn:reconstruction_loss}
\mathcal{L}_\mathrm{rec}=\sum_{Y\in\{T,R\}}\| \hat{Y} -Y \|_1.
\end{equation}

\noindent\textbf{Perceptual loss.}
Given a pre-trained VGG-19~\cite{VGG} model $\phi$, we minimize the $\ell_1$ difference between $\phi(\hat{T})$, $\phi(\hat{R})$ and $\phi(T)$, $\phi(R)$ in the selected feature layers,

\begin{equation}
\begin{aligned}
\label{eqn:perceptual_loss}
\mathcal{L}_\mathrm{percep}=\sum_{Y\in\{T,R\}}\sum\limits_{l}\kappa_l\| \phi_l(\hat{Y}) -\phi_l(Y) \|_1,
\end{aligned}
\end{equation}
where $l$ indicates the index of $conv1\_2$, $conv2\_2$, $conv3\_2$, $conv4\_2$ and $conv5\_2$ layers.
The weights $\{\kappa_l\}$ are used to balance different layers.

\noindent\textbf{Exclusion loss.}
Following~\cite{Zhang2018CVPR}, the exclusion loss is formulated as,
\begin{equation}
\label{eqn:exclusion_loss}
\mathcal{L}_\mathrm{excl}=\frac{1}{N\!+\!1}\sum\limits_{n=0}^N\sqrt{\|\Psi(T^{\downarrow n},R^{\downarrow n}) \|_{_F}}.
\end{equation}
where $\Psi(T,R)={\tanh}(\lambda_{_T}\vert \nabla T \vert)\circ {\tanh}(\lambda_{_R}\vert \nabla R \vert)$, and $\lambda_{T}$ and $\lambda_{R}$ denote normalization factors.
$\nabla T$ and $\nabla R$ are gradients of $T$ and $R$.
$\|\cdot\|_{F}$ is the Frobenius norm.
$T^{\downarrow n}$ and $R^{\downarrow n}$ represent the $n\times$ down-sampling versions of $T$ and $R$, where $T^{\downarrow 0}$ and $R^{\downarrow 0}$ are the original inputs.
In practice, we set $N$ = 2, $\lambda_{T} = \frac{1}{2}$, and $\lambda_{R} = \frac{\|\nabla T \|_1}{\|\nabla{R}\|_1}$.

\noindent\textbf{Adversarial Loss.}
Adversarial loss is adopted to further enhance the visual quality of the output images.
We treat the whole RAGNet as the generator $G$ and additionally build a $4$-layer discriminator $D$, whose parameters are updated via,
\begin{equation}
\mathcal{L}_D=\mathbb{E}_{I,T}{\log}D(I,T)+\mathbb{E}_{I}(1-{\log}D(I,G(I))),
\end{equation}
while the parameters of $G$ are optimized by,
\begin{equation}
    \mathcal{L}_\mathrm{adv}=-\mathbb{E}_{I}{\log}D(I,G(I)).
\end{equation}

Taking the above loss functions into account, the learning objective to train our RAGNet can be formulated as,
\begin{equation}
\label{eqn:total_loss}
\mathcal{L}=\lambda_1\mathcal{L}_\mathrm{rec}+\lambda_2\mathcal{L}_\mathrm{percep}+\lambda_3\mathcal{L}_\mathrm{excl}+\lambda_4\mathcal{L}_\mathrm{adv}+\lambda_5\mathcal{L}_\mathrm{mask},
\end{equation}
where $\lambda_1=\lambda_2=\lambda_5=1, \lambda_3=0.2$ and $\lambda_4=0.01$.

\section{Experiments}\label{sec:experiments}

\subsection{Implementation Details}\label{subsec:implementation_details}

\noindent\textbf{Training Data.}
Following previous works, the proposed RAGNet is trained with both synthetic and real-world images, and we use the same data synthesis protocol as ERRNet~\cite{ERRNet}.
Specifically, $7,643$ image pairs are chosen from the PASCAL VOC dataset~\cite{VOC}, and each pair is used to generate the transmission $T$ and the reflection $R$, respectively.
In each iteration, we select a pair of two images, whose shorter sides are randomly scaled into $[224, 448]$, and two $224\times224$ patches are then cropped to synthesize the input $I$.
For real-world data, Zhang \etal~\cite{Zhang2018CVPR} collected 110 input-transmission pairs, and we follow the official split where 90 image pairs are used for training.
Since the ground-truth reflection $R$ is unavailable, the corresponding reconstruction and and perceptual losses on $\hat{R}$ are omitted when training with real-world images.

\noindent\textbf{Evaluation Datasets.}
Five commonly used real-world datasets are exploited for evaluation, including 20 image pairs (\emph{Real 20}) from~\cite{Zhang2018CVPR}, 45 images (\emph{Real 45}) from~\cite{CEILNet} and three subsets from SIR$^2$~\cite{Wan_2017_ICCV}, including
\romannumeral1)~\emph{SIR$^2$ Wild} with 55 wild scene images,
\romannumeral2)~\emph{SIR$^2$ Solid} with 200 controlled scene images of a set of daily-life objects, and
\romannumeral3)~\emph{SIR$^2$ Postcard} with 199 controlled scene images on postcards, which are obtained by using one postcard as transmission and another as reflection.
Note that we only give the quantitative results of \emph{Real 20} and three SIR$^2$ subsets, since the ground-truth transmission of \emph{Real 45} is unavailable.

\noindent\textbf{Implementation Details.}
The model is optimized by the Adam~\cite{Adam} optimizer with $\beta_1=0.9$, $\beta_2=0.999$ and a fixed learning rate of $1\times10^{-4}$.
For stable training, we first train $G_R$ for 50 epochs, and then the whole model is jointly trained for another 100 epochs.
All the experiments are conducted in the PyTorch~\cite{PyTorch} environment running on a PC with an Nvidia RTX 2080Ti GPU.
The source code and pre-trained model are available at \url{https://github.com/liyucs/RAGNet}.

\begin{figure*}[tbp]
	\small
	\centering
	\scalebox{1}{
		\begin{tabular}{ccccc}
			\includegraphics[width=0.16\textwidth]{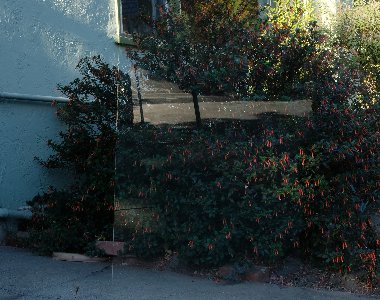}&\hspace{-4.2mm}
			\includegraphics[width=0.16\textwidth]{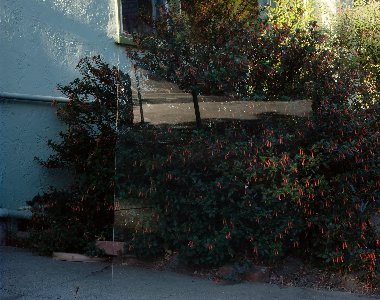}&\hspace{-4.2mm}
			\includegraphics[width=0.16\textwidth]{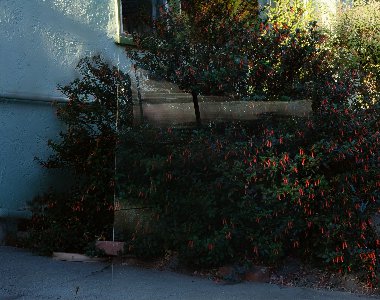}&\hspace{-4.2mm}
			\includegraphics[width=0.16\textwidth]{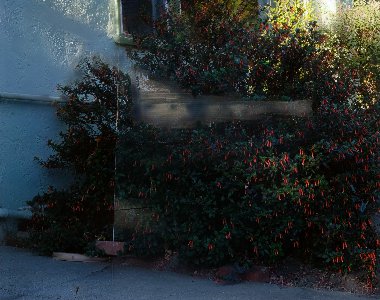}&\hspace{-4.2mm}
			\includegraphics[width=0.16\textwidth]{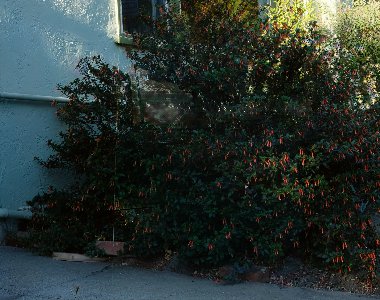}
			\\
			\includegraphics[width=0.16\textwidth]{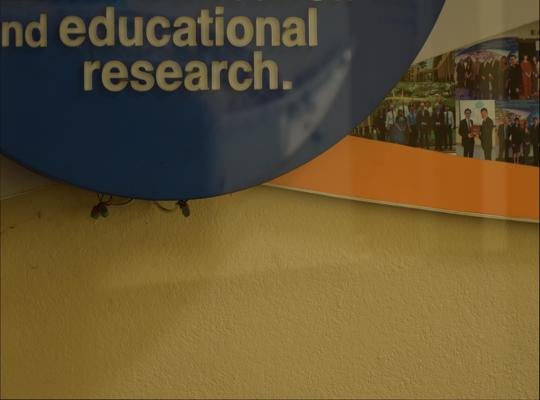}&\hspace{-4.2mm}
			\includegraphics[width=0.16\textwidth]{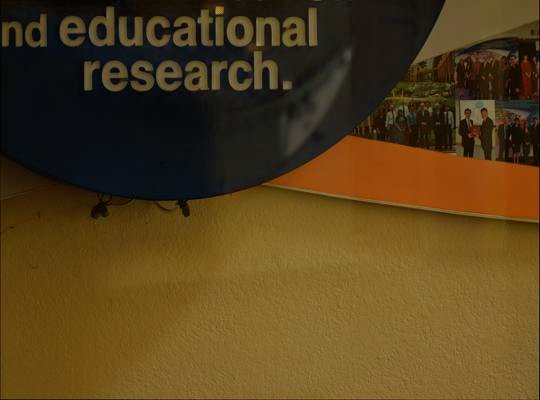}&\hspace{-4.2mm}
			\includegraphics[width=0.16\textwidth]{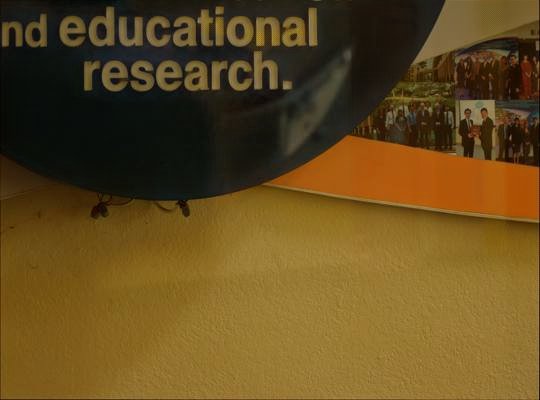}&\hspace{-4.2mm}
			\includegraphics[width=0.16\textwidth]{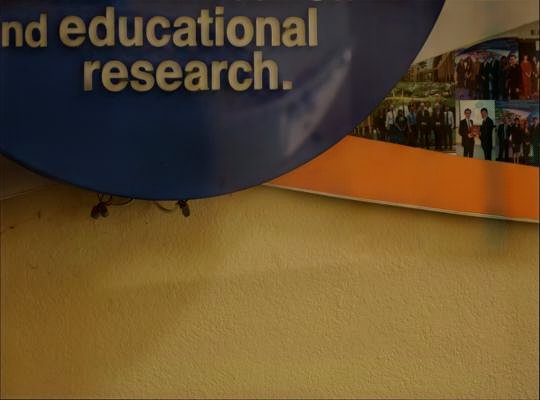}&\hspace{-4.2mm}
			\includegraphics[width=0.16\textwidth]{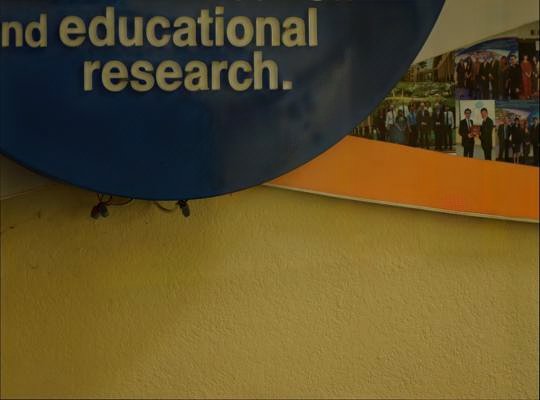}
			\\
			Input&\hspace{-4.2mm}
			RAGNet$_\mathbf{F}$&\hspace{-4.2mm}
			RAGNet$_{\mathbf{F}\circ\mathbf{M}}$& \hspace{-4.2mm}
			RAGNet$_{\mathbf{F}\circ\mathbf{M}^{2c}}$& \hspace{-4.2mm}
			RAGNet
			\\
	\end{tabular}}
	\vspace{-3mm}
	\captionsetup{font={small}}
	\caption{Visual results of ablation studies on mask utilization schemes. The images are from \emph{Real 20} and \emph{SIR$^2$ Wild} datasets, respectively. Please zoom in for better observation.}
	\label{fig:ablation_mask}
	\vspace{-2mm}
\end{figure*}

\begin{table}[t]
	\vspace{-3mm}
	\captionsetup{font={small}}
	\caption{Quantitative comparison of different settings on structure and loss functions.}
	\vspace{-2mm}
	\centering
	\label{tab:ablation_structure_loss}
	\scalebox{0.9}{
		\begin{tabular}{lcc}%
			\toprule
			\multirow{2}{*}{Model}   & \emph{Real 20}  & \emph{SIR$^2$ Wild} \\
			\cline{2-3}
			{}& PSNR / SSIM   &  PSNR / SSIM \\
			\midrule
			RAGNet$_{I \to T}$                &20.73 / 0.753  &   24.37 / 0.871\\
			w/o $\mathbf{F}_\mathit{diff}$                &20.99 / 0.758  &   24.57 / 0.865\\
			\midrule
			w/o ${\rm \mathcal{L}_{mask}}$ &21.32 / 0.764  &   25.09 / 0.878\\
			w/o $\mathcal{L}_\mathrm{mask}^\mathit{diff}$ &22.08 / 0.769  &   25.09 / 0.877\\
			w/o $\mathcal{L}_\mathrm{mask}^\mathit{reg}$ &22.11 / 0.773  &   25.12 / 0.876\\
			\midrule
			RAGNet                    &22.95 / 0.793   &   25.52 / 0.880\\
			\bottomrule
	\end{tabular}}
\vspace{-1mm}
\end{table}

\subsection{Comparison with State-of-the-arts}\label{subsec:comparison}
We compare the proposed RAGNet with five state-of-the-art SIRR methods, \ie, CEILNet~\cite{CEILNet}, Zhang \etal~\cite{Zhang2018CVPR}, BDN~\cite{BDN}, ERRNet~\cite{ERRNet} {,} IBCLN~\cite{IBCLN}, as well as {the adjusted MIRR method by only taking the observation $I$ as the network input, \ie, Lei \etal~\cite{Lei_2020_CVPR}}.
We finetune these models on our training dataset
and report the better result between the finetuned version
and the released one.
For Lei \etal~\cite{Lei_2020_CVPR}, we adjust the network input and retrain the model with our training set.
%

The PSNR and SSIM metrics of all competing methods are reported in Table~\ref{tab:comparison}.
Note that the results of \emph{Real 45} are omitted due to lack of ground-truth.
{Therefore, a user study is conducted on \emph{Real 45} to evaluate the reflection removal quality and the results also show the superiority of our RAGNet. The details are given in the supplementary material.}
It can be seen that the proposed RAGNet achieves the best performance by PSNR on three of the four datasets, and surpasses all other methods on average quantitative performance by a large margin ($\sim$0.8dB in PSNR).
{The quantitative gain against other deep cascaded models~\cite{CEILNet,BDN,IBCLN,Lei_2020_CVPR} also indicates the benefit of appropriate  reflection-aware guidance.}

Fig.~\ref{fig:comparison0} shows the qualitative results and the corresponding ground-truth on three SIR$^2$ subsets and \emph{Real 20} dataset, and the results on \emph{Real 45} dataset are given in Fig.~\ref{fig:comparison1}.
One can see that CEILNet may fail when the reflections have clear edges, where the smoothness assumption is violated (\eg, the second row of Fig.~\ref{fig:comparison0}).
Zhang \etal~\cite{Zhang2018CVPR} is prone to over-processing, which generates dim results with color aberration in most cases, while BDN~\cite{BDN} tends to generate brighter output images than input ones.
Moreover, the reflections seem insufficiently removed in the results of ERRNet~\cite{ERRNet} and IBCLN~\cite{IBCLN}.
In general, our method outperforms the competing methods, and removes the reflection areas more accurately and thoroughly.
Please refer to the supplementary material for more qualitative results.

\section{Ablation Study}\label{sec:ablation_study}

In this section, we perform ablation studies on two outdoor real-world datasets, \ie, \emph{Real 20} and \emph{SIR$^2$ Wild}.

\subsection{Two-stage Architecture}\label{subsec:ablation_two_stage}
Our RAGNet is designed as a two-stage network, which takes reflection as an intermediate result to ease the reflection removal procedure.
However, it may be concerned whether a well-designed one-stage network is sufficient for SIRR.
To show the necessity and superiority of the two-stage architecture, we additionally design a single-stage variant, namely RAGNet$_{I \to T}$, which directly estimates the transmission layer from the input.

According to Fig.~\ref{fig:architecture}, the generation of $T$ in RAGNet requires features from two sources, \ie, $\mathbf{M}_\mathit{diff} \circ \mathbf{F}_\mathit{diff}$ and $\mathbf{M}_{dec} \circ \mathbf{F}_\mathit{dec}$.
%
We expect that $\mathbf{M}_\mathit{diff}\to0$ when reflection is heavy as in Eqn.~(\ref{eqn:mask_loss_diff}), so that only $\mathbf{F}_\mathit{dec}$ provides information for recovering heavy reflection areas.
%
When designing RAGNet$_{I \to T}$, we retain the functionality of the skip connection and $\mathbf{F}_\mathit{dec}$:
one for providing auxiliary information via subtraction operations, the other focuses on gradually recovering $\hat{T}$.
The structure of the one-stage model is provided in the supplementary material.
%
The results in Table~\ref{tab:ablation_structure_loss} and Fig.~\ref{fig:ablation_structure} show that the one-stage variant RAGNet$_{I \to T}$ suffers a huge performance degradation, indicating the essentiality of the two-stage architecture for our RAGNet.

\begin{table}[t]
	\captionsetup{font={small}}
    \caption{Quantitative comparison of different settings on mask utilization schemes.}
    \vspace{-3mm}
    \centering
    \label{tab:ablation_mask}
    \scalebox{0.9}{
        \begin{tabular}{lcc}%
        \toprule
        \multirow{2}{*}{Model}   & \emph{Real 20}        & \emph{SIR$^2$ Wild} \\
        \cline{2-3}
        {}& PSNR / SSIM      &  PSNR / SSIM \\
        \midrule
        RAGNet$_\mathbf{F}$                  &20.93 / 0.750    &   24.47 / 0.875\\
        RAGNet$_{\mathbf{F}\circ\mathbf{M}}$                  &21.60 / 0.772    &   24.86 / 0.875\\
        RAGNet$_{\mathbf{F}\circ\mathbf{M}^{2c}}$                   &21.69 / 0.774    &   24.69 / 0.871\\
        \midrule
        RAGNet                    &22.95 / 0.793     &   25.52 / 0.880\\
        \bottomrule
    \end{tabular}}
\vspace{-1mm}
\end{table}

\subsection{RAG Module}\label{subsec:ablation_rag}

\noindent\textbf{Difference Features.}
To evaluate the effectiveness of difference features, we replace $\mathbf{F}_\mathit{diff}$ with $\mathbf{F}_{I}$ and keep the other parts of the model unchanged, \ie, the subtraction operation is discarded.
As shown in Fig.~\ref{fig:ablation_structure}, without $\mathbf{F}_\mathit{diff}$, the model performs poorly in suppressing the reflection, leading to an obvious performance drop in Table~\ref{tab:ablation_structure_loss}.

\noindent\textbf{Masks.}
In order to verify the settings about masks, we conduct ablation studies on three variants,
(i)~RAGNet$_\mathbf{F}$: the masks are totally discarded, where the partial convolution is accordingly replaced by vanilla convolution.
(ii)~RAGNet$_{\mathbf{F}\circ\mathbf{M}}$: the masks are multiplied with the feature $\mathbf{F}$, in other words, the renormalization operation in partial convolutions is removed.
(iii)~RAGNet$_{\mathbf{F}\circ\mathbf{M}^{2c}}$: a unified one-channel mask is predicted for $\mathbf{F}_\mathit{diff}$ and $\mathbf{F}_\mathit{dec}$ respectively, \ie, $\mathbf{M}_\mathit{diff}^{1c}$ and $\mathbf{M}_\mathit{dec}^{1c}$, where $\mathbf{M}^{2c}=[\mathbf{M}_\mathit{diff}^{1c}, \mathbf{M}_\mathit{dec}^{1c}]$.

Table~\ref{tab:ablation_mask} shows the quantitative results of the ablation studies on masks.
It can be seen that our RAGNet surpasses its variants RAGNet$_\mathbf{F}$ and RAGNet$_{\mathbf{F}\circ\mathbf{M}}$ in both PSNR and SSIM, indicating the effectiveness of the deployment of mask and partial convolution.
Furthermore, the ablation study on RAGNet$_{\mathbf{F}\circ\mathbf{M}^{2c}}$ shows that predicting a per-channel mask better suits the SIRR task in our framework.

Fig.~\ref{fig:ablation_mask} shows the qualitative results of the ablation studies.
It can be seen that, all of RAGNet$_\mathbf{F}$, RAGNet$_{\mathbf{F}\circ\mathbf{M}}$ and RAGNet$_{\mathbf{F}\circ\mathbf{M}^{2c}}$ fail to remove the heavy reflections, especially in complex environments (\eg, in the second row of Fig.~\ref{fig:ablation_mask}, the light on the ground is very similar to the reflections).
%
RAGNet outperforms the others by recovering the heavy reflection areas with contextual information, making the output image visually pleasant.
Moreover, by integrating partial convolution, the reflection removal performance is also enhanced for images with mild reflections.

\subsection{Mask Loss}\label{subsec:ablation_loss}
From the aspect of the proposed mask loss, we conduct three experiments, \ie , discarding $\mathcal{L}_{\mathrm{mask}_\mathit{diff}}$, $\mathcal{L}_{\mathrm{mask}_\mathit{util}}$ and both of them, respectively.
Table \ref{tab:ablation_structure_loss} and Fig.~\ref{fig:ablation_structure} show that with the mask loss, the performance of our RAGNet is improved by a large margin.
Furthermore, one can see that both items of the specifically designed mask loss are essential for RAGNet, and greatly boost the quantitative performance and visual quality.
The thresholds $\xi$ and $\varphi$ in Eqns. (\ref{eqn:mask_loss_diff}) and (\ref{eqn:mask_loss_reg}) are discussed in the supplementary material.

\section{Conclusion}\label{sec:conclusion}
In this paper, we investigated the two-stage framework for the single image reflection removal (SIRR) task, and presented an RAGNet to exploit the reflection-aware guidance via the RAG module.
The RAG module suppresses the reflection by using the difference between observation and reflection features, while a mask is generated to indicate the extent to which the transmission is corrupted by heavy reflection and to collaborate with partial convolution for mitigating the effect of deviating from the linear combination hypothesis.
A mask loss is accordingly designed for reconciling the contributions of encoder and decoder features.
Extensive experiments on five widely used real-world datasets indicate that the proposed RAGNet outperforms the state-of-the-art methods by a large margin.

\clearpage

\renewcommand\thesection{\Alph{section}}
\renewcommand\thefigure{\Alph{figure}}
\renewcommand\thetable{\Alph{table}}
\renewcommand\thesubsection{\thesection.\arabic{subsection}}

\newcommand*{\affaddr}[1]{#1} 
\newcommand*{\affmark}[1][*]{\textsuperscript{#1}}
\newcommand*{\email}[1]{\texttt{#1}}

\newcommand{\tabincell}[2]{\begin{tabular}{@{}#1@{}}#2\end{tabular}}

\twocolumn[
\begin{center}
	{\LARGE \textbf{Supplemental Materials\\~\\~\\}}
\end{center}]
\setcounter{section}{0}
\section{Network Structure}\label{sec:NetworkStructure}
Our RAGNet is comprised of two components, \ie, a $G_R$ to predict the reflection layer $\hat{R}$ and a $G_T$ to generate the transmission layer $\hat{T}$ on the basis of $\hat{R}$.

The subnetworks $G_R$ and $G_T$ both follow the structual configuration of U-Net. Especially, all the encoders of $G_R$ and $G_T$ share the same network structure, while the decoders differ slightly in the use of partial convolution. The detailed structures of $G_R$ and $G_T$ are shown in Table~\ref{tab:structure}.

\vspace{1mm}
\setcounter{table}{0}
\begin{strip}
	\centering\noindent
	\centering%
	\vspace{2mm}
	\begin{tabular}{ccc||ccc}%
		\specialrule{1pt}{0pt}{0pt}
		\multicolumn{3}{c||}{Encoder} & \multicolumn{3}{c}{Decoder} \\
		\hhline{---||---}
		Layer & Output size & Filter & Layer & Output size & Filter \\
		\hhline{===||===}
		Conv, ReLU & $224\times224$ & $3\rightarrow64$ & Transposed Conv &$28\times28$& $512\rightarrow512$\\
		Conv, ReLU & $224\times224$ & $64\rightarrow64$ & Concat, $Conv^{\ast}$, ReLU &$28\times28$& $(512+512)\rightarrow1024$ \\
		MaxPool & $112\times112$ & - & {[Conv, ReLU]} $\times$3 &$28\times28$& $1024\rightarrow1024$ \\
		{}&{}&{}&Conv, ReLU &$28\times28$& $1024\rightarrow256$ \\
		\hhline{---||---}
		Conv, ReLU  &$112\times112$& $64\rightarrow128$ & Transposed Conv &$56\times56$& $256\rightarrow256$ \\
		Conv, ReLU & $112\times112$ & $128\rightarrow128$ & Concat, $Conv^{\ast}$, ReLU &$56\times56$& $(256+256)\rightarrow512$  \\
		MaxPool & $56\times56$ & - & {[Conv, ReLU]} $\times$3 &$56\times56$& $512\rightarrow512$ \\
		{}&{}&{}&Conv, ReLU &$56\times56$& $512\rightarrow128$\\
		\hhline{---||---}
		Conv, ReLU & $56\times56$ & $128\rightarrow256$ & Transposed Conv &$112\times112$& $128\rightarrow128$ \\
		{[Conv, ReLU]} $\times$3 &$56\times56$& $256\rightarrow256$ & Concat, $Conv^{\ast}$, ReLU &$112\times112$& $(128+128)\rightarrow256$  \\
		MaxPool &$28\times28$& - & Conv, ReLU &$112\times112$& $256\rightarrow256$  \\
		{}&{}&{}&Conv, ReLU & $112\times112$ & $256\rightarrow64$  \\
		\hhline{---||---}
		Conv, ReLU &$28\times28$& $256\rightarrow512$ & Transposed Conv & $224\times224$ & $64\rightarrow64$ \\
		{[Conv, ReLU]} $\times$3 &$28\times28$&  $512\rightarrow512$ & Concat, $Conv^{\ast}$, ReLU& $224\times224$ & $(64+64)\rightarrow128$ \\
		MaxPool &$14\times14$&  - & Conv, ReLU & $224\times224$ & $128\rightarrow128$ \\
		\hhline{---||---}
		{[Conv, ReLU]} $\times$4 &$14\times14$& $512\rightarrow512$ & Conv, ReLU & $224\times224$ & $128\rightarrow64$ \\
		{}&{}& {} & Conv, Sigmoid & $224\times224$ & $64\rightarrow3$ \\
		\specialrule{1pt}{0pt}{0pt}
	\end{tabular}
	\vspace{1mm}
	\captionof{table}{Structure configuration of $G_R$ and $G_T$. The encoders of $G_R$ and $G_T$ have the same structure, which is shown in the left column. The structure of decoders is shown in the right column. $Conv^{\ast}$ denotes vanilla convolution in $G_R$ and partial convolution in $G_T$. For vanilla convolutions and partial convolutions, kernel size=3, stride=1, while for maxpooling layers and transposed convolutions, kernel size=2, stride=2.}%
	\label{tab:structure}%
	\vspace{2mm}
\end{strip}	
\vspace{1mm}
\section{One-stage Structure}\label{sec:onestagesec}
The structure of single-stage network RAGNet$_{I \to T}$ used in ablation study is shown in Fig.~\ref{fig:onestage}.
RAGNet$_{I \to T}$ predicts transmission layer $\hat{T}$ directly from observation $I$.
%

\section{Running Time and Number of Parameters}\label{sec:numofparam}
{RAGNet$_{I \to T}$ has 63.5M parameters, which is slightly less than Zhang \etal (77.6M) and BDN (75.2M) but achieves comparable performance according to Tables {2} and {3} in the main text. While RAGNet (including $G_R$ and $G_T$) has 130.9M parameters.
We also note that RAGNet has a running time of 0.15$s$ per image (540$\times$400) on an Nvidia RTX 2080Ti GPU, which is comparable to Zhang \etal and is faster than the other competing methods (see Table.~\ref{tab:time}).}

\section{Ablation Study on $\varphi$ and $\xi$}\label{sec:thresholds}

$\varphi$ and $\xi$ are thresholds that define the intensity levels of reflection, \ie, areas with reflection intensity higher than $\varphi$ are treated as heavy reflection areas, while areas with reflection intensity lower than $\xi$ are treated as weak reflection areas.
We conduct an ablation study on $\varphi$ and $\xi$ on \emph{Real 20} and SIR$^2$ test datasets with two groups of values, \ie, $\varphi=\{0.20, 0.25, 0.30, 0.35, 0.40\}$ and $\xi=\{0.005, 0.010, 0.015, 0.020\}$.
Considering the PSNR and SSIM performance exhibited in  Fig.~\ref{fig:thretholds}, the final thresholds are set as $\varphi = 0.3$ and $\xi = 0.01$.
%

\clearpage

\setcounter{figure}{0}
\begin{strip}
	\vspace{-3mm}
	\begin{overpic}[width=1\textwidth]{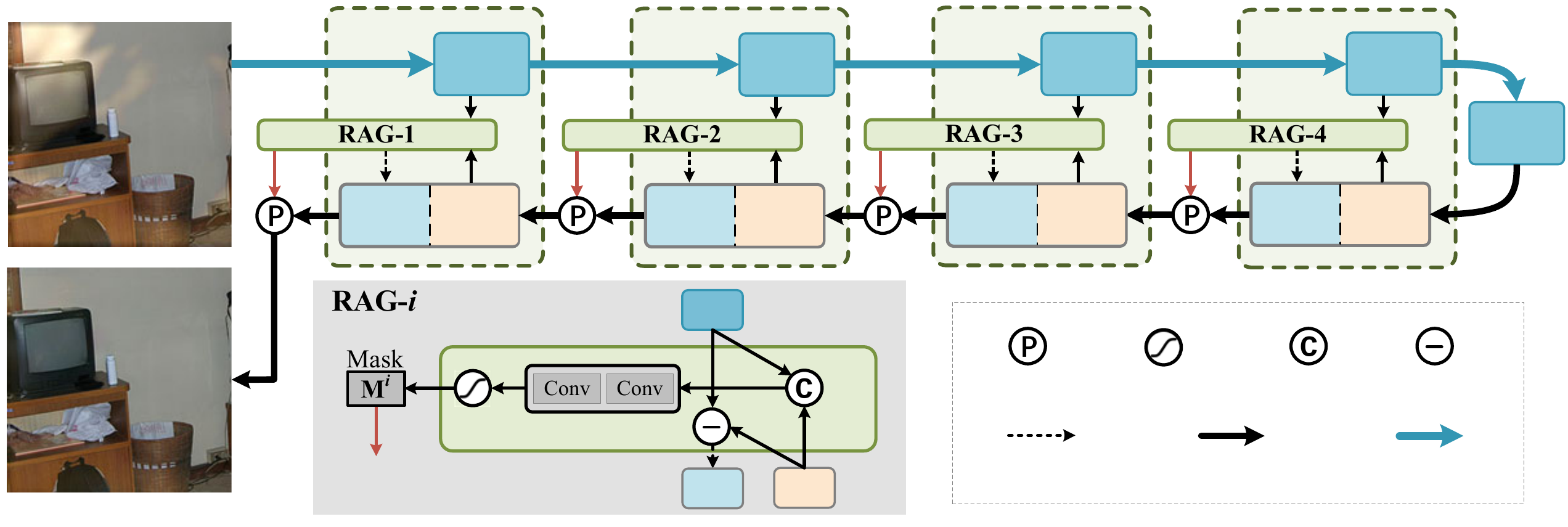}
		\put(15.5,30){\normalsize $I$}
		\put(15.5,10.1){\normalsize $\hat{T}$}
		\put(29.5,29){\normalsize $\mathbf{F}_I^1$}
		\put(48.8,29){\normalsize $\mathbf{F}_I^2$}
		\put(68,29){\normalsize $\mathbf{F}_I^3$}
		\put(87.2,29){\normalsize $\mathbf{F}_I^4$}
		\put(95.3,24.5){\normalsize $\mathbf{F}_I^5$}
		\put(22.6,19.3){\normalsize $\mathbf{F}_\mathit{diff}^1$}
		\put(28.5,19.3){\normalsize $\mathbf{F}_\mathit{dec}^1$}
		\put(41.9,19.3){\normalsize $\mathbf{F}_\mathit{diff}^2$}
		\put(47.8,19.3){\normalsize $\mathbf{F}_\mathit{dec}^2$}
		\put(61,19.3){\normalsize $\mathbf{F}_\mathit{diff}^3$}
		\put(66.8,19.3){\normalsize $\mathbf{F}_\mathit{dec}^3$}
		\put(80.1,19.3){\normalsize $\mathbf{F}_\mathit{diff}^4$}
		\put(86.2,19.3){\normalsize $\mathbf{F}_\mathit{dec}^4$}
		
		\put(44.2,13.4){\footnotesize $\mathbf{F}_I^i$}
		\put(43.6,2.1){\footnotesize $\mathbf{F}_\mathit{diff}^i$}
		\put(49.5,2.1){\footnotesize $\mathbf{F}_\mathit{dec}^i$}
		\put(88,8.6){\textbf{\scriptsize Subtraction}}
		\put(81.3,8.6){\textbf{\scriptsize Concat}}
		\put(71.8,8.6){\textbf{\scriptsize Sigmoid}}
		\put(75,3){\textbf{\scriptsize Decoder $\bm{D_T}$}}
		\put(62.1,3){\textbf{\scriptsize Skip Connect}}
		\put(62,8.6){\textbf{\scriptsize Partial Conv}}
		\put(87.8,3){\textbf{\scriptsize Encoder $\bm{E_I}$}}
	\end{overpic}
	\captionof{figure}{Structure of the single-stage network RAGNet$_{I \to T}$, which directly estimates the
		transmission layer from the observation $I$. RAGNet$_{I \to T}$ follows the U-Net setting, while the concatenated feature through the skip-connection is $\mathbf{F}_\mathit{diff}=\mathbf{F}_I-\mathbf{F}_\mathit{dec}$ rather than $\mathbf{F}_I$.}
	\label{fig:onestage}
	\vspace{-6mm}
\end{strip}

\begin{strip}
	\small
	\centering
	\scalebox{.92}{
		\begin{tabular}{c}
			\includegraphics[width=1\textwidth]{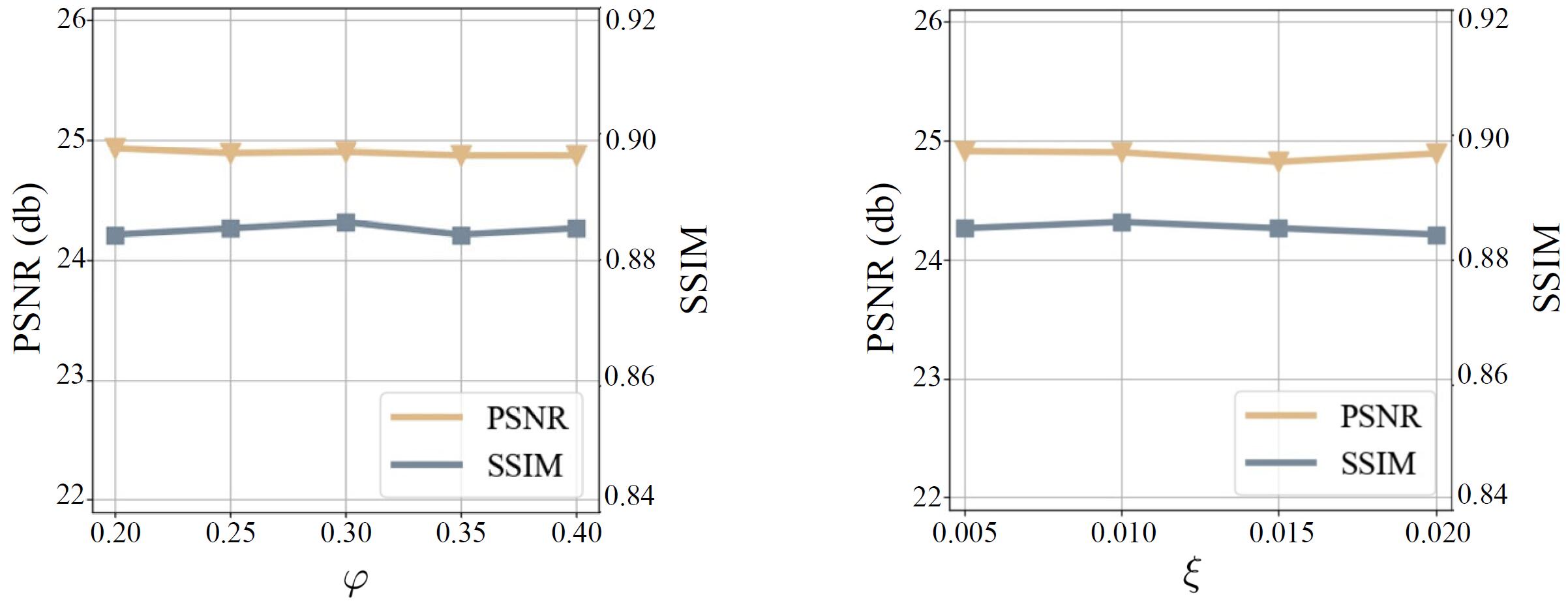}
			\\
	\end{tabular}}
	\vspace{-3mm}
	\captionof{figure}{PSNR and SSIM performance under different settings of $\varphi$ (left) and $\xi$ (right).}
	\label{fig:thretholds}
	\vspace{-5mm}
\end{strip}

\begin{strip}
	\small
	\centering
	\vspace{-3mm}
	\scalebox{0.95}{
		\begin{tabular}{cccc}
			\includegraphics[width=0.24\textwidth]{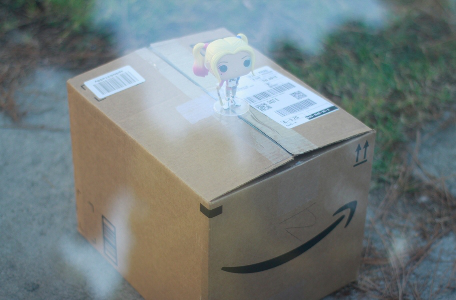}&\hspace{-4.2mm}
			\includegraphics[width=0.24\textwidth]{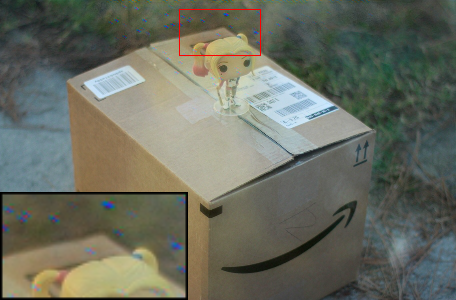}&\hspace{-4.2mm}
			\includegraphics[width=0.24\textwidth]{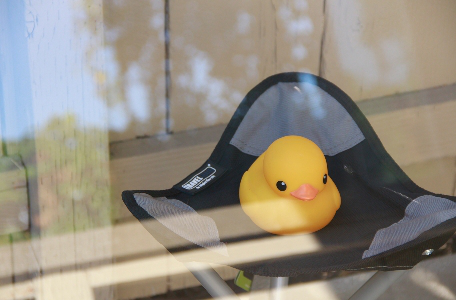}&\hspace{-4.2mm}
			\includegraphics[width=0.24\textwidth]{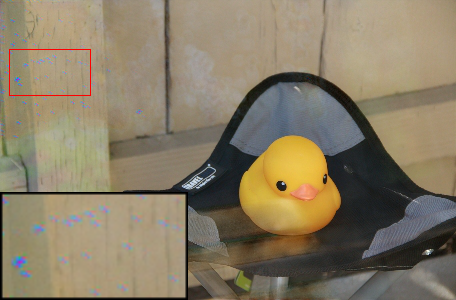}
			\\
			\includegraphics[width=0.24\textwidth]{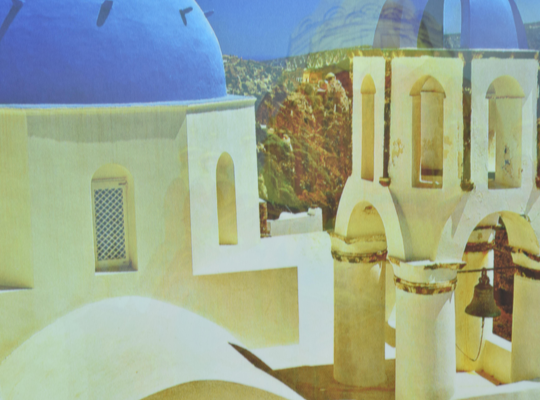}&\hspace{-4.2mm}
			\includegraphics[width=0.24\textwidth]{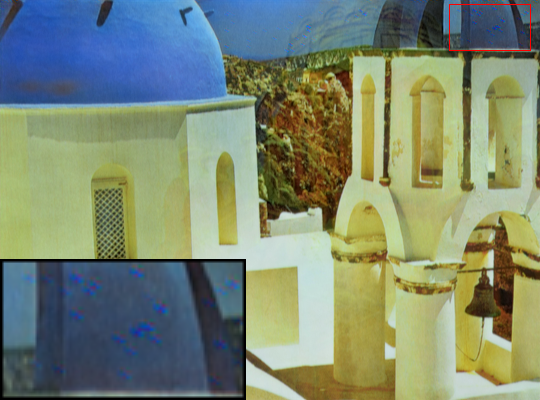}&\hspace{-4.2mm}
			\includegraphics[width=0.24\textwidth]{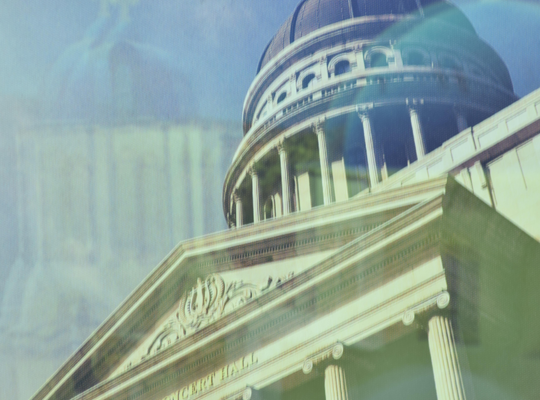}&\hspace{-4.2mm}
			\includegraphics[width=0.24\textwidth]{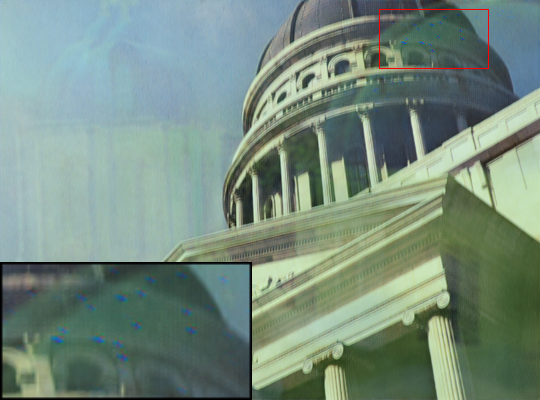}
			\\
			Input&\hspace{-4.2mm}
			IBCLN~\cite{IBCLN}&\hspace{-4.2mm}
			Input&\hspace{-4.2mm}
			IBCLN~\cite{IBCLN}			
			\\
	\end{tabular}}
	\captionsetup{font={small}}
	\vspace{1mm}
	\captionof{figure}{Undesired artifacts in some scenarios by IBCLN~\cite{IBCLN}. }
	\label{fig:ibcln}
\end{strip}


\begin{strip}
	\small
	\centering
	\scalebox{.95}{
		\begin{tabular}{ccccccc}
			\includegraphics[width=0.14\textwidth]{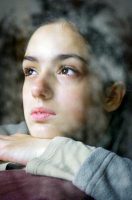}&\hspace{-4.2mm}
			\includegraphics[width=0.14\textwidth]{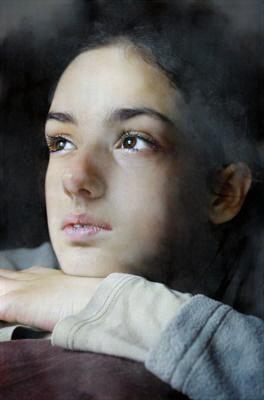}&\hspace{-4.2mm}
			\includegraphics[width=0.14\textwidth]{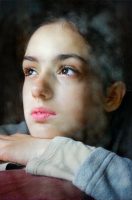}&\hspace{-4.2mm}
			\includegraphics[width=0.14\textwidth]{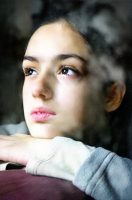}&\hspace{-4.2mm}
			\includegraphics[width=0.14\textwidth]{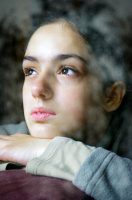}&\hspace{-4.2mm}
			\includegraphics[width=0.14\textwidth]{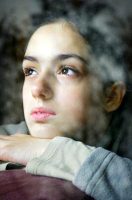}&\hspace{-4.2mm}
			\includegraphics[width=0.14\textwidth]{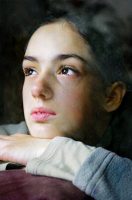}
			\\
			\includegraphics[width=0.14\textwidth]{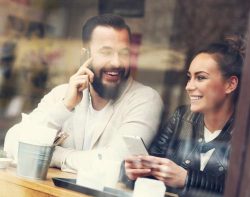}&\hspace{-4.2mm}
			\includegraphics[width=0.14\textwidth]{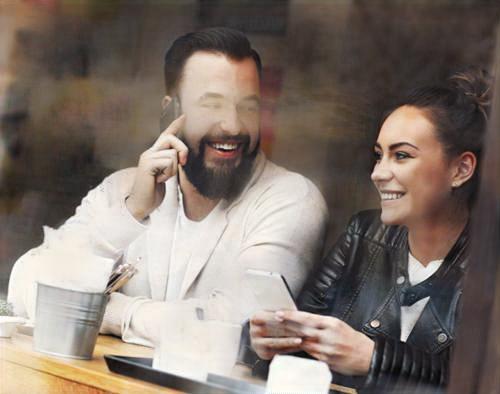}&\hspace{-4.2mm}
			\includegraphics[width=0.14\textwidth]{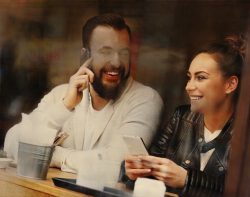}&\hspace{-4.2mm}
			\includegraphics[width=0.14\textwidth]{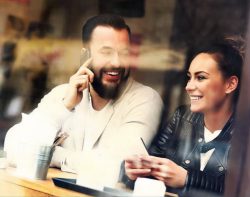}&\hspace{-4.2mm}
			\includegraphics[width=0.14\textwidth]{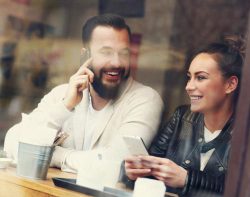}&\hspace{-4.2mm}
			\includegraphics[width=0.14\textwidth]{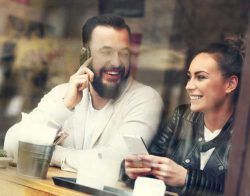}&\hspace{-4.2mm}
			\includegraphics[width=0.14\textwidth]{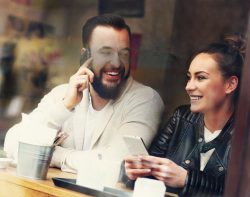}
			\\
			\includegraphics[width=0.14\textwidth]{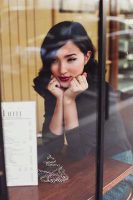}&\hspace{-4.2mm}
			\includegraphics[width=0.14\textwidth]{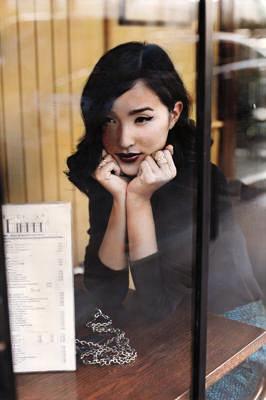}&\hspace{-4.2mm}
			\includegraphics[width=0.14\textwidth]{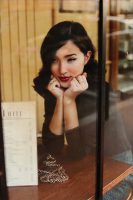}&\hspace{-4.2mm}
			\includegraphics[width=0.14\textwidth]{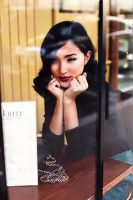}&\hspace{-4.2mm}
			\includegraphics[width=0.14\textwidth]{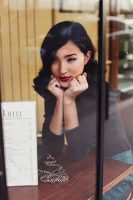}&\hspace{-4.2mm}
			\includegraphics[width=0.14\textwidth]{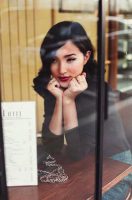}&\hspace{-4.2mm}
			\includegraphics[width=0.14\textwidth]{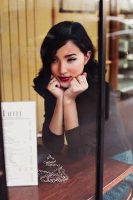}
			\\
			\includegraphics[width=0.14\textwidth]{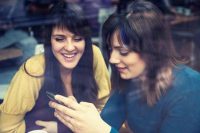}&\hspace{-4.2mm}
			\includegraphics[width=0.14\textwidth]{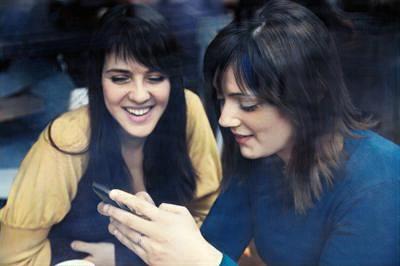}&\hspace{-4.2mm}
			\includegraphics[width=0.14\textwidth]{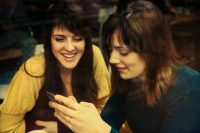}&\hspace{-4.2mm}
			\includegraphics[width=0.14\textwidth]{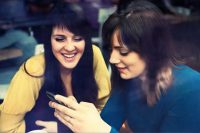}&\hspace{-4.2mm}
			\includegraphics[width=0.14\textwidth]{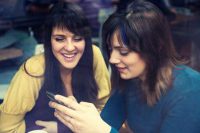}&\hspace{-4.2mm}
			\includegraphics[width=0.14\textwidth]{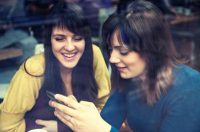}&\hspace{-4.2mm}
			\includegraphics[width=0.14\textwidth]{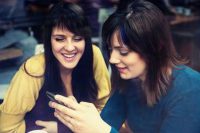}
			\\
			\includegraphics[width=0.14\textwidth]{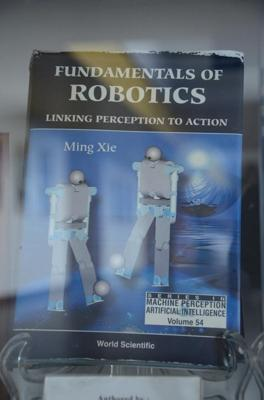}&\hspace{-4.2mm}
			\includegraphics[width=0.14\textwidth]{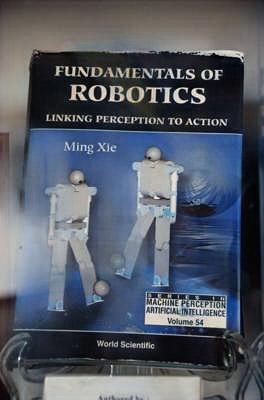}&\hspace{-4.2mm}
			\includegraphics[width=0.14\textwidth]{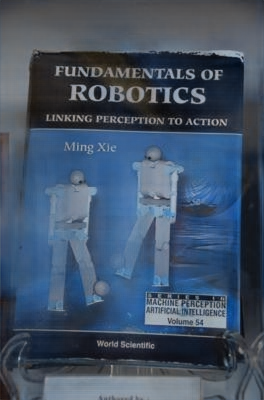}&\hspace{-4.2mm}
			\includegraphics[width=0.14\textwidth]{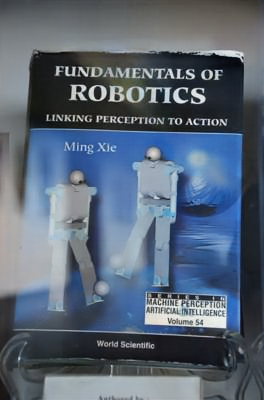}&\hspace{-4.2mm}
			\includegraphics[width=0.14\textwidth]{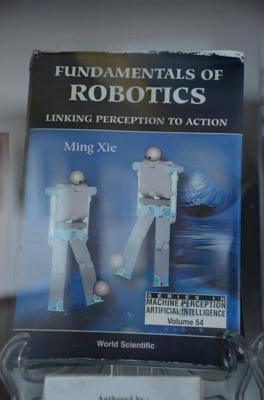}&\hspace{-4.2mm}
			\includegraphics[width=0.14\textwidth]{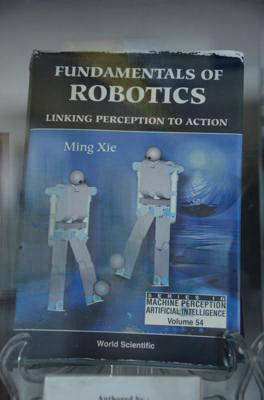}&\hspace{-4.2mm}
			\includegraphics[width=0.14\textwidth]{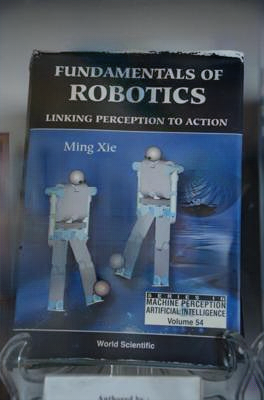}
			\\
			\includegraphics[width=0.14\textwidth]{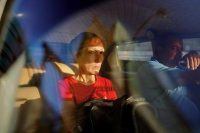}&\hspace{-4.2mm}
			\includegraphics[width=0.14\textwidth]{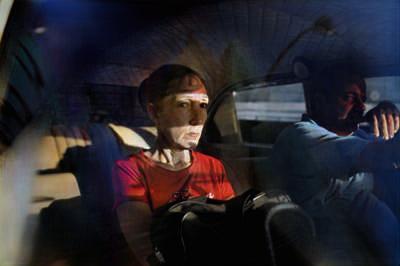}&\hspace{-4.2mm}
			\includegraphics[width=0.14\textwidth]{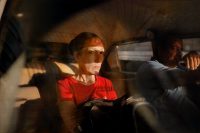}&\hspace{-4.2mm}
			\includegraphics[width=0.14\textwidth]{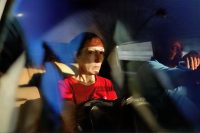}&\hspace{-4.2mm}
			\includegraphics[width=0.14\textwidth]{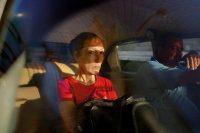}&\hspace{-4.2mm}
			\includegraphics[width=0.14\textwidth]{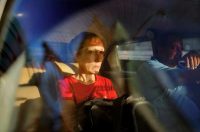}&\hspace{-4.2mm}
			\includegraphics[width=0.14\textwidth]{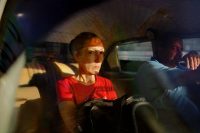}
			\\
			Input&\hspace{-4.2mm}
			CEILNet~\cite{CEILNet}&\hspace{-4.2mm}
			Zhang \etal~\cite{Zhang2018CVPR}&\hspace{-4.2mm}
			BDN~\cite{BDN}& \hspace{-4.2mm}
			ERRNet~\cite{ERRNet}& \hspace{-4.2mm}
			IBCLN~\cite{IBCLN}& \hspace{-4.2mm}
			Ours
			\\
	\end{tabular}}
	\captionof{figure}{Visual comparison on \emph{Real 45} dataset, where the ground-truth is unavailable. Please zoom in for better observation.}
	\label{fig:comparison01}
\end{strip}

\begin{strip}
	\centering\noindent
	\centering%
	\vspace{-4mm}
	\captionof{table}{The comparison of running time and number of parameters.}
	\begin{tabular}{cccccccc}
		\toprule
		Method & CEILNet~\cite{CEILNet} & Zhang~\cite{Zhang2018CVPR} & BDN~\cite{BDN} & ERRNet~\cite{ERRNet} & IBCLN~\cite{IBCLN}& RAGNet$_{I \to T}$ &RAGNet\\
		\midrule
		Time ($s$) & 0.26 & 0.16 & 0.22 & 0.27 & 0.22 & 0.08 & 0.15\\
		Parameter (M) &2.2  & 77.6 & 75.2 & 86.7 & 21.2 & 63.5 &130.9\\
		\bottomrule
	\end{tabular}%
	\label{tab:time}
\end{strip}

\clearpage

\begin{table*}[t]
	\centering
	\captionsetup{font={small}}
	\caption{The performance of RAGNet trained with an additional dataset.  We expand our training and test datasets with the Nature dataset~\cite{IBCLN}, which contains 200 image pairs for training and 20 image pairs for testing.}
	\vspace{-3mm}
	\label{tab:retrain}
	\scalebox{1}{
		\begin{tabular}{lcccccc}
			\toprule
			\multirow{2}{*}{Datasets}  &  {\emph{SIR$^2$ Solid} (200)} & \emph{SIR$^2$ Postcard} (199) & \emph{SIR$^2$ Wild} (55) &  \emph{Real 20} (20) & \emph{Nature} (20)& Average (494) \\
			\cmidrule(l){2-7}
			{} & PSNR / SSIM & PSNR / SSIM  & PSNR / SSIM & PSNR / SSIM & PSNR / SSIM & PSNR / SSIM  \\
			\midrule
			RAGNet &26.16 / 0.907 & 24.42 / 0.883 & 25.99 / 0.890 & 22.48 / 0.786 & 22.95 / 0.792& 25.16 / 0.886 \\
			\bottomrule
	\end{tabular}}
\end{table*}

\begin{table*}[htbp]
	\centering
	\captionsetup{font={small}}
	\caption{Comparison with SoTA methods on regions with weak and strong reflection on the 4 test sets. The PSNR for weak and strong reflection regions are respectively listed in the first and second row.}
	\label{tabC}
	\begin{tabular}{ccccccc}
		\toprule
		Reflection Intensity & CEILNet~\cite{CEILNet} & Zhang~\cite{Zhang2018CVPR} & BDN~\cite{BDN} & ERRNet~\cite{ERRNet} & IBCLN~\cite{IBCLN} & Ours\\
		\midrule
		Weak & 21.51 & 19.38 & 21.86 & 23.79 & 24.28 & 25.18\\
		Strong & 21.01 & 20.29 & 21.14 & 21.32 & 21.43 & 23.40\\
		\bottomrule
	\end{tabular}%
\end{table*}

\begin{strip}
	\small
	\centering
	\scalebox{.95}{
		\begin{tabular}{cccccccc}
			\includegraphics[width=0.124\textwidth]{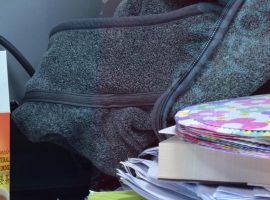}&\hspace{-4.2mm}
\includegraphics[width=0.124\textwidth]{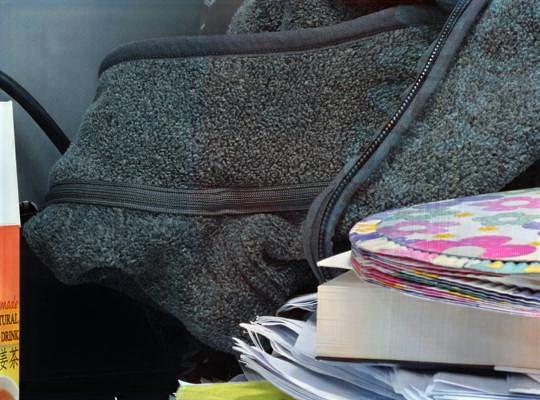}&\hspace{-4.2mm}
\includegraphics[width=0.124\textwidth]{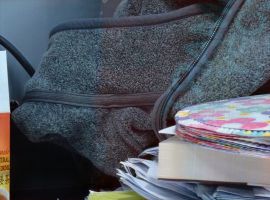}&\hspace{-4.2mm}
\includegraphics[width=0.124\textwidth]{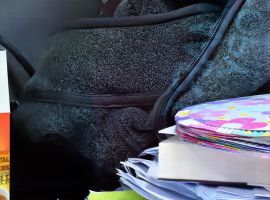}&\hspace{-4.2mm}
\includegraphics[width=0.124\textwidth]{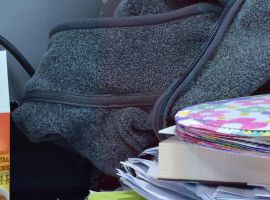}&\hspace{-4.2mm}
\includegraphics[width=0.124\textwidth]{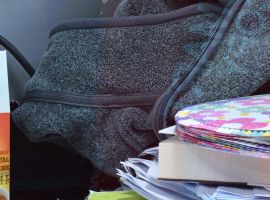}&\hspace{-4.2mm}
\includegraphics[width=0.124\textwidth]{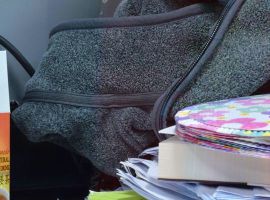}&\hspace{-4.2mm}
\includegraphics[width=0.124\textwidth]{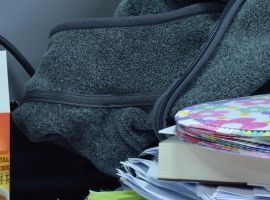}
\\
\includegraphics[width=0.124\textwidth]{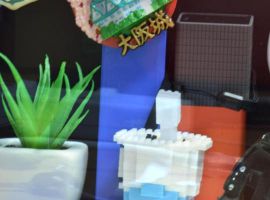}&\hspace{-4.2mm}
\includegraphics[width=0.124\textwidth]{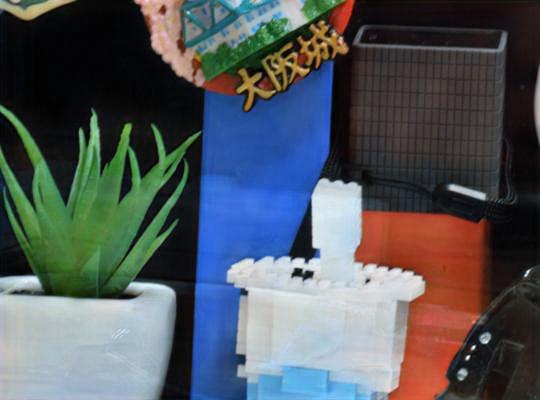}&\hspace{-4.2mm}
\includegraphics[width=0.124\textwidth]{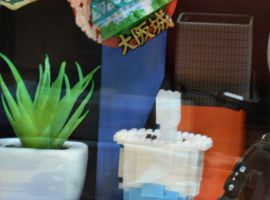}&\hspace{-4.2mm}
\includegraphics[width=0.124\textwidth]{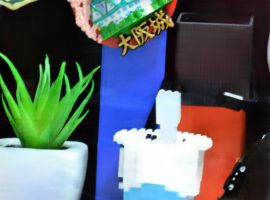}&\hspace{-4.2mm}
\includegraphics[width=0.124\textwidth]{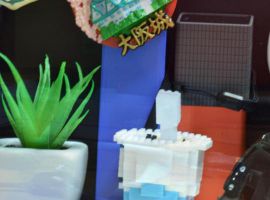}&\hspace{-4.2mm}
\includegraphics[width=0.124\textwidth]{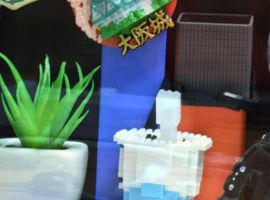}&\hspace{-4.2mm}
\includegraphics[width=0.124\textwidth]{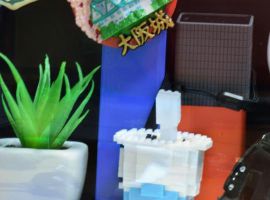}&\hspace{-4.2mm}
\includegraphics[width=0.124\textwidth]{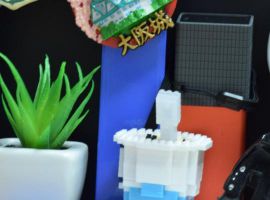}
\\
\includegraphics[width=0.124\textwidth]{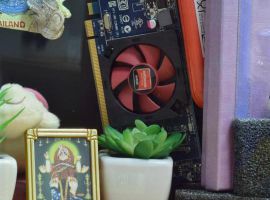}&\hspace{-4.2mm}
\includegraphics[width=0.124\textwidth]{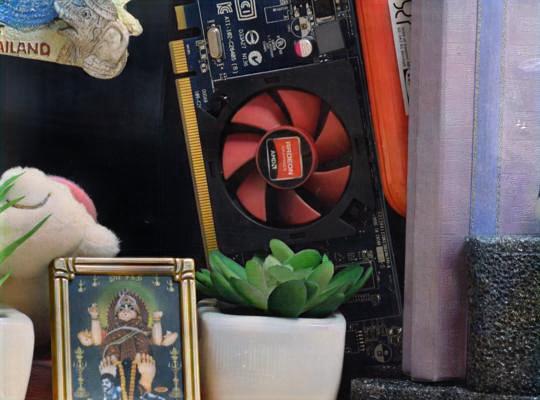}&\hspace{-4.2mm}
\includegraphics[width=0.124\textwidth]{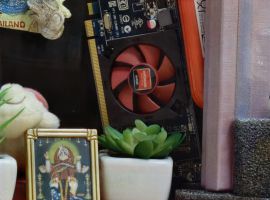}&\hspace{-4.2mm}
\includegraphics[width=0.124\textwidth]{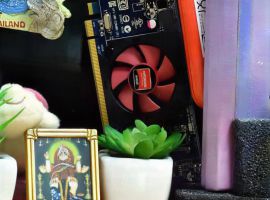}&\hspace{-4.2mm}
\includegraphics[width=0.124\textwidth]{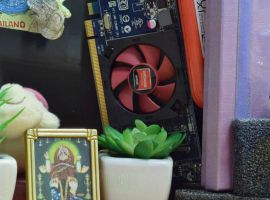}&\hspace{-4.2mm}
\includegraphics[width=0.124\textwidth]{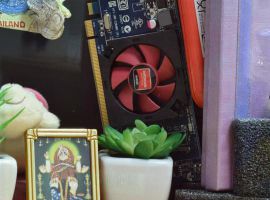}&\hspace{-4.2mm}
\includegraphics[width=0.124\textwidth]{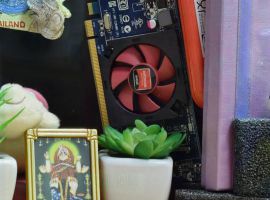}&\hspace{-4.2mm}
\includegraphics[width=0.124\textwidth]{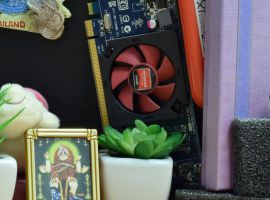}
\\
\includegraphics[width=0.124\textwidth]{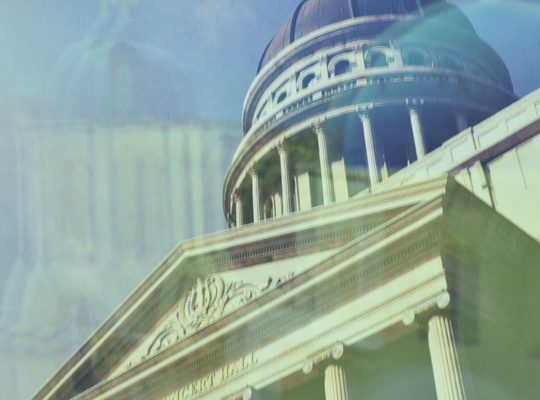}&\hspace{-4.2mm}
\includegraphics[width=0.124\textwidth]{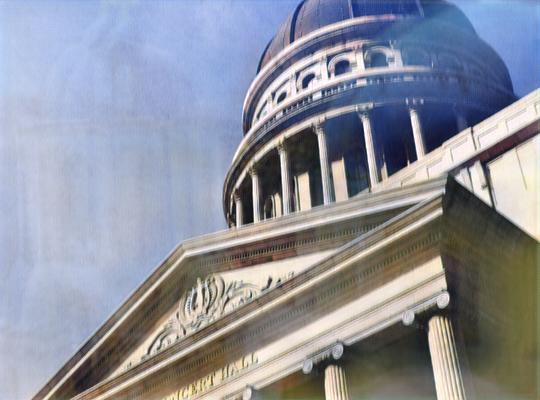}&\hspace{-4.2mm}
\includegraphics[width=0.124\textwidth]{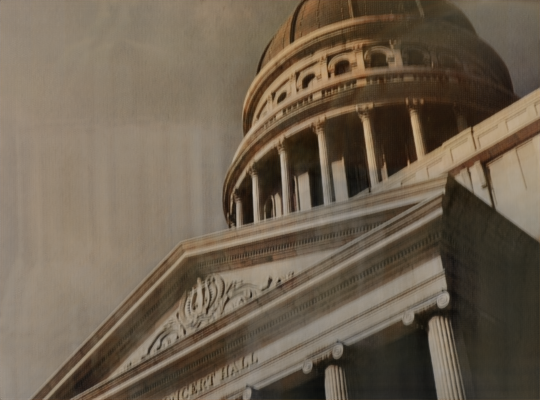}&\hspace{-4.2mm}
\includegraphics[width=0.124\textwidth]{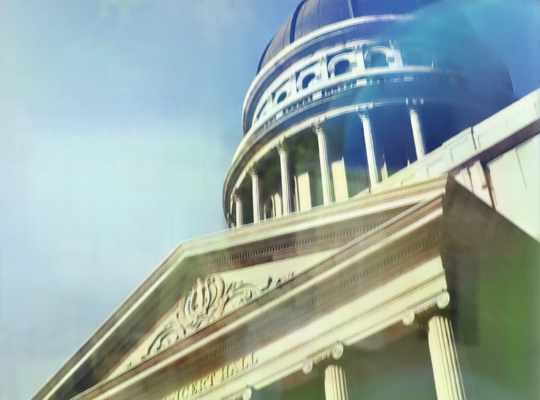}&\hspace{-4.2mm}
\includegraphics[width=0.124\textwidth]{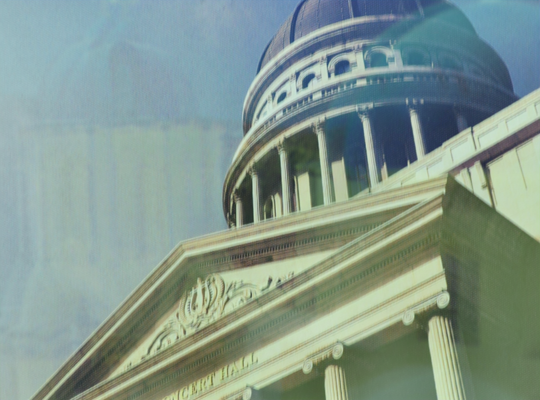}&\hspace{-4.2mm}
\includegraphics[width=0.124\textwidth]{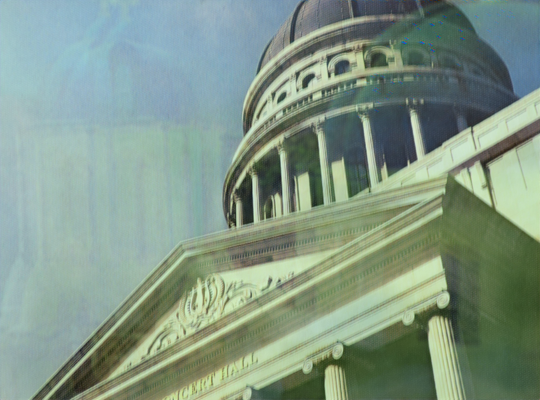}&\hspace{-4.2mm}
\includegraphics[width=0.124\textwidth]{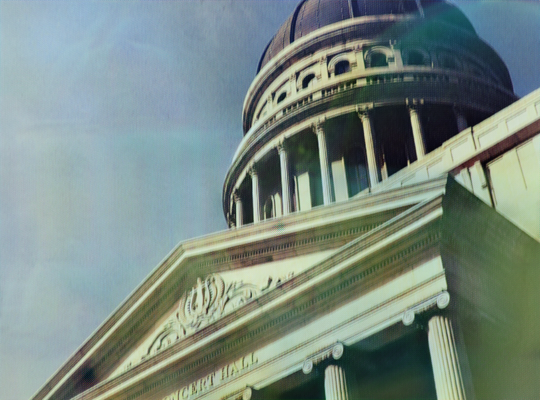}&\hspace{-4.2mm}
\includegraphics[width=0.124\textwidth]{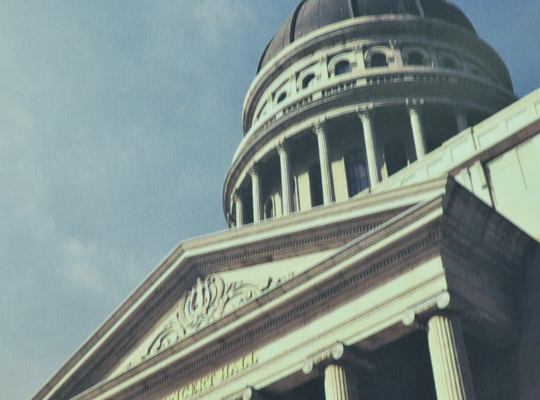}
\\
\includegraphics[width=0.124\textwidth]{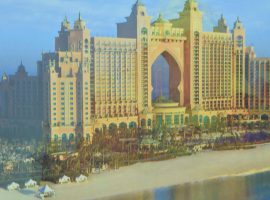}&\hspace{-4.2mm}
\includegraphics[width=0.124\textwidth]{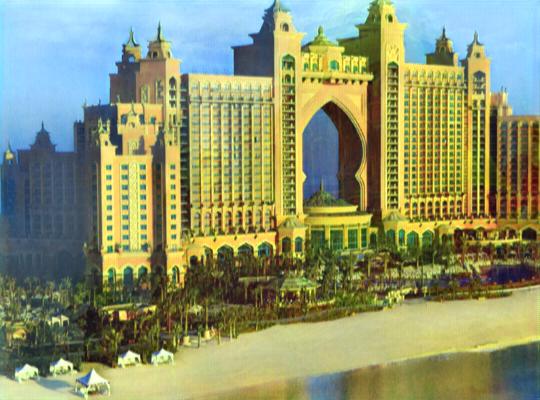}&\hspace{-4.2mm}
\includegraphics[width=0.124\textwidth]{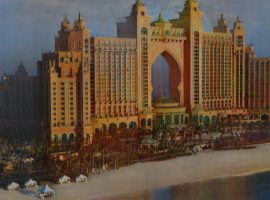}&\hspace{-4.2mm}
\includegraphics[width=0.124\textwidth]{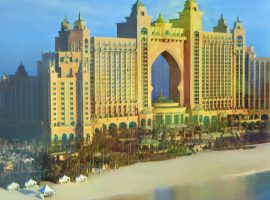}&\hspace{-4.2mm}
\includegraphics[width=0.124\textwidth]{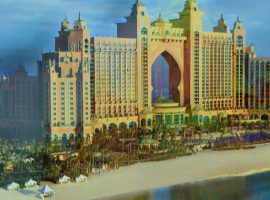}&\hspace{-4.2mm}
\includegraphics[width=0.124\textwidth]{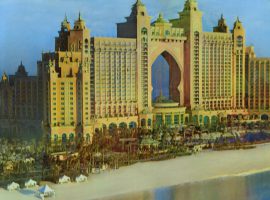}&\hspace{-4.2mm}
\includegraphics[width=0.124\textwidth]{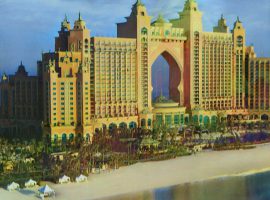}&\hspace{-4.2mm}
\includegraphics[width=0.124\textwidth]{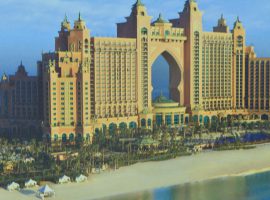}
\\
\includegraphics[width=0.124\textwidth]{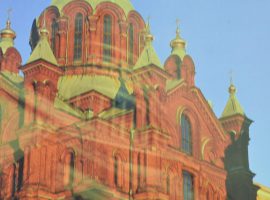}&\hspace{-4.2mm}
\includegraphics[width=0.124\textwidth]{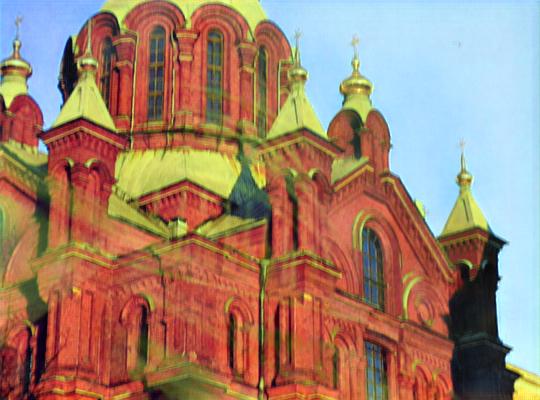}&\hspace{-4.2mm}
\includegraphics[width=0.124\textwidth]{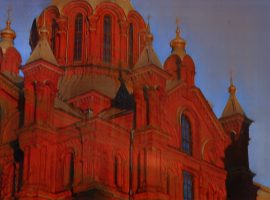}&\hspace{-4.2mm}
\includegraphics[width=0.124\textwidth]{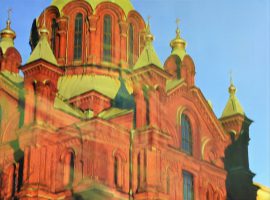}&\hspace{-4.2mm}
\includegraphics[width=0.124\textwidth]{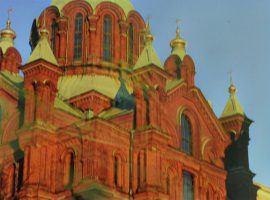}&\hspace{-4.2mm}
\includegraphics[width=0.124\textwidth]{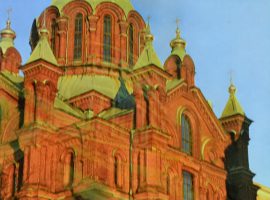}&\hspace{-4.2mm}
\includegraphics[width=0.124\textwidth]{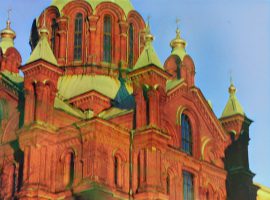}&\hspace{-4.2mm}
\includegraphics[width=0.124\textwidth]{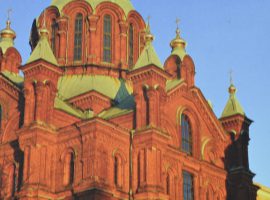}
\\
\includegraphics[width=0.124\textwidth]{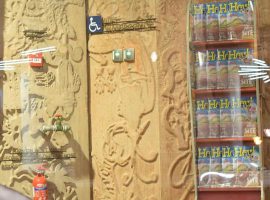}&\hspace{-4.2mm}
\includegraphics[width=0.124\textwidth]{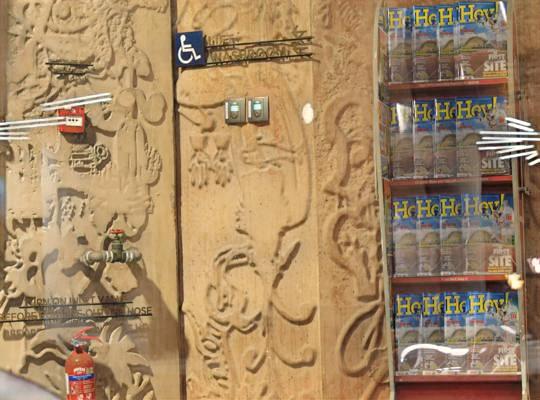}&\hspace{-4.2mm}
\includegraphics[width=0.124\textwidth]{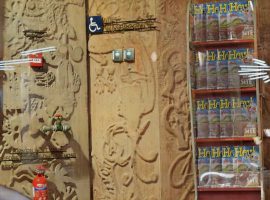}&\hspace{-4.2mm}
\includegraphics[width=0.124\textwidth]{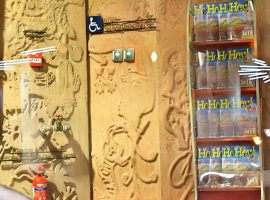}&\hspace{-4.2mm}
\includegraphics[width=0.124\textwidth]{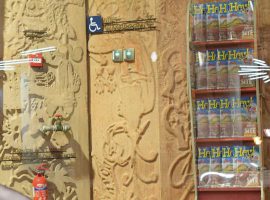}&\hspace{-4.2mm}
\includegraphics[width=0.124\textwidth]{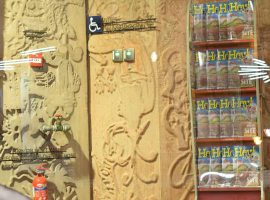}&\hspace{-4.2mm}
\includegraphics[width=0.124\textwidth]{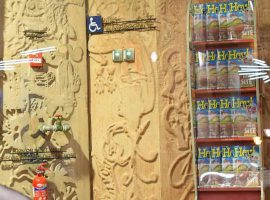}&\hspace{-4.2mm}
\includegraphics[width=0.124\textwidth]{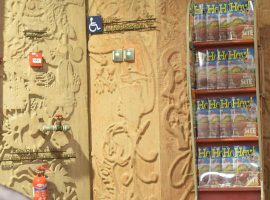}
\\
\includegraphics[width=0.124\textwidth]{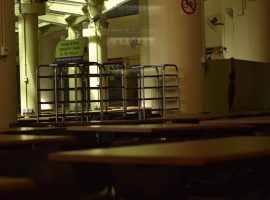}&\hspace{-4.2mm}
\includegraphics[width=0.124\textwidth]{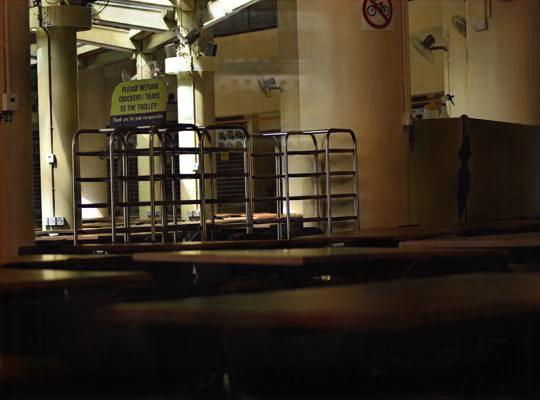}&\hspace{-4.2mm}
\includegraphics[width=0.124\textwidth]{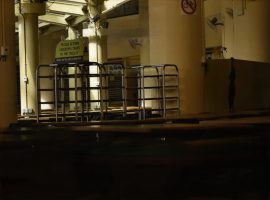}&\hspace{-4.2mm}
\includegraphics[width=0.124\textwidth]{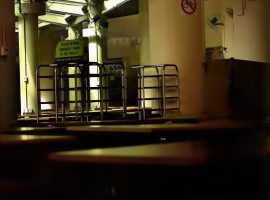}&\hspace{-4.2mm}
\includegraphics[width=0.124\textwidth]{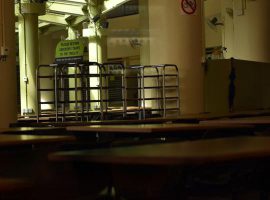}&\hspace{-4.2mm}
\includegraphics[width=0.124\textwidth]{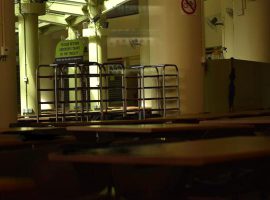}&\hspace{-4.2mm}
\includegraphics[width=0.124\textwidth]{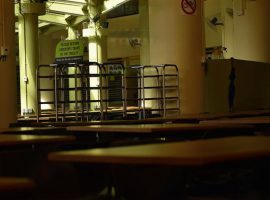}&\hspace{-4.2mm}
\includegraphics[width=0.124\textwidth]{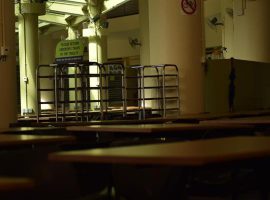}
\\
\includegraphics[width=0.124\textwidth]{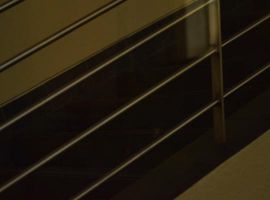}&\hspace{-4.2mm}
\includegraphics[width=0.124\textwidth]{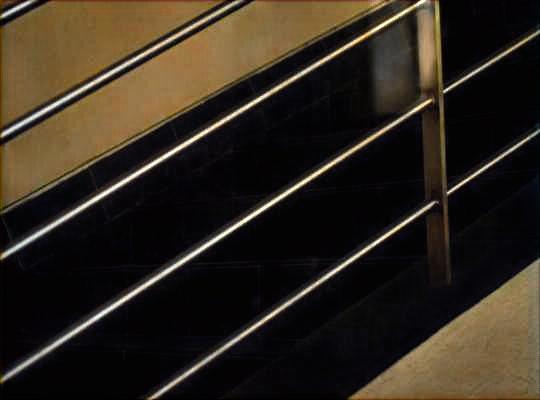}&\hspace{-4.2mm}
\includegraphics[width=0.124\textwidth]{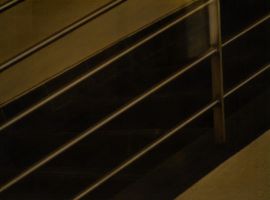}&\hspace{-4.2mm}
\includegraphics[width=0.124\textwidth]{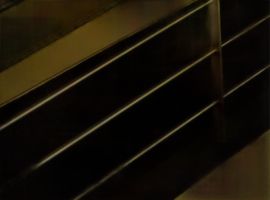}&\hspace{-4.2mm}
\includegraphics[width=0.124\textwidth]{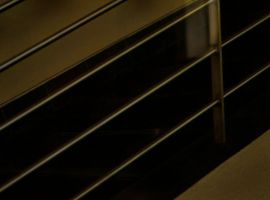}&\hspace{-4.2mm}
\includegraphics[width=0.124\textwidth]{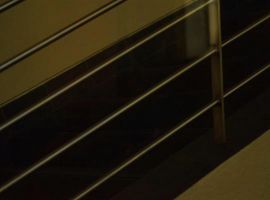}&\hspace{-4.2mm}
\includegraphics[width=0.124\textwidth]{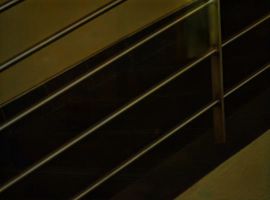}&\hspace{-4.2mm}
\includegraphics[width=0.124\textwidth]{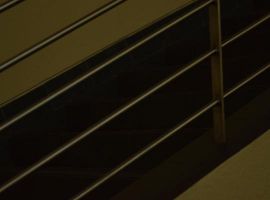}
\\
\includegraphics[width=0.124\textwidth]{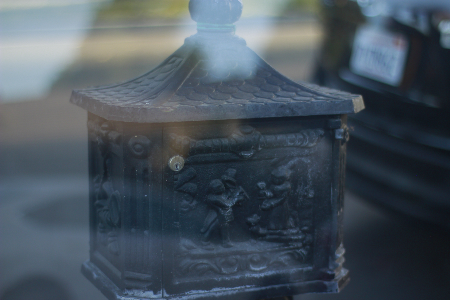}&\hspace{-4.2mm}
\includegraphics[width=0.124\textwidth]{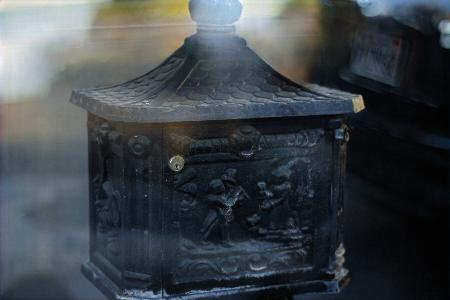}&\hspace{-4.2mm}
\includegraphics[width=0.124\textwidth]{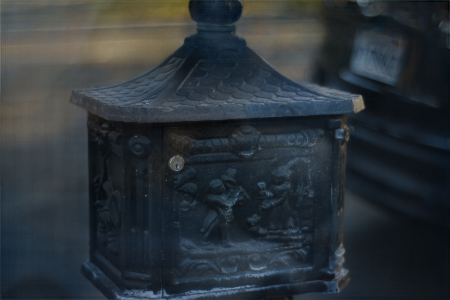}&\hspace{-4.2mm}
\includegraphics[width=0.124\textwidth]{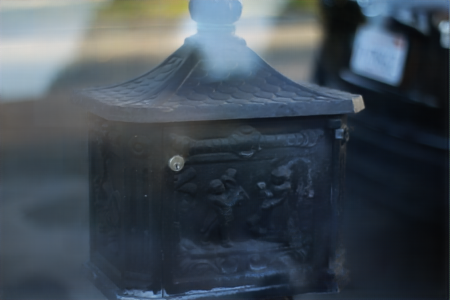}&\hspace{-4.2mm}
\includegraphics[width=0.124\textwidth]{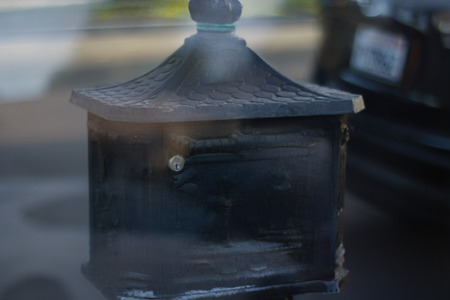}&\hspace{-4.2mm}
\includegraphics[width=0.124\textwidth]{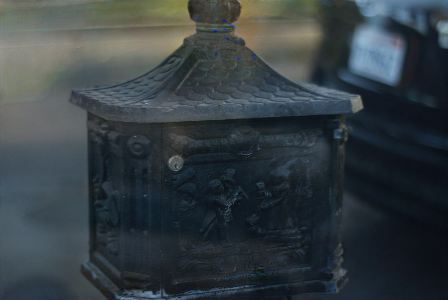}&\hspace{-4.2mm}
\includegraphics[width=0.124\textwidth]{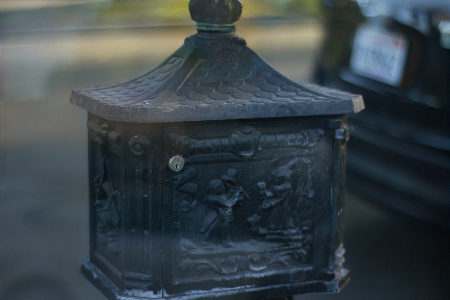}&\hspace{-4.2mm}
\includegraphics[width=0.124\textwidth]{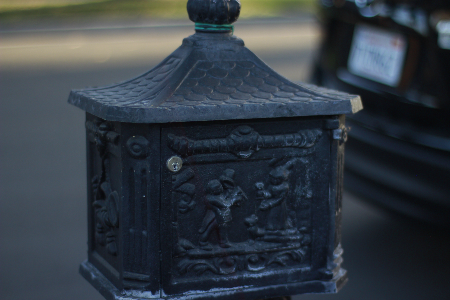}
\\
\includegraphics[width=0.124\textwidth]{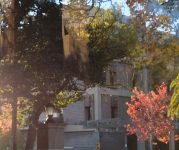}&\hspace{-4.2mm}
\includegraphics[width=0.124\textwidth]{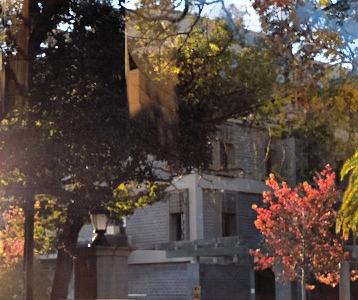}&\hspace{-4.2mm}
\includegraphics[width=0.124\textwidth]{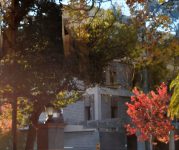}&\hspace{-4.2mm}
\includegraphics[width=0.124\textwidth]{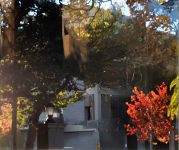}&\hspace{-4.2mm}
\includegraphics[width=0.124\textwidth]{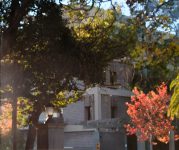}&\hspace{-4.2mm}
\includegraphics[width=0.124\textwidth]{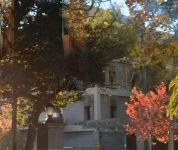}&\hspace{-4.2mm}
\includegraphics[width=0.124\textwidth]{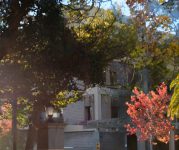}&\hspace{-4.2mm}
\includegraphics[width=0.124\textwidth]{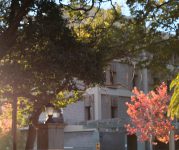}
\\
\includegraphics[width=0.124\textwidth]{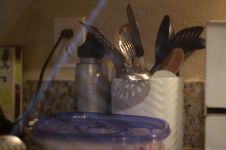}&\hspace{-4.2mm}
\includegraphics[width=0.124\textwidth]{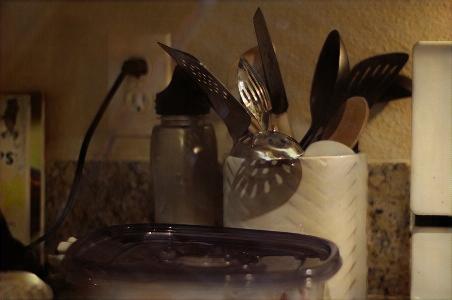}&\hspace{-4.2mm}
\includegraphics[width=0.124\textwidth]{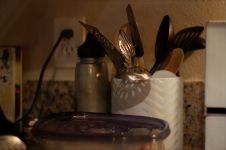}&\hspace{-4.2mm}
\includegraphics[width=0.124\textwidth]{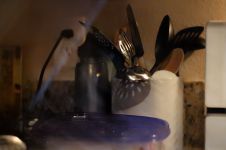}&\hspace{-4.2mm}
\includegraphics[width=0.124\textwidth]{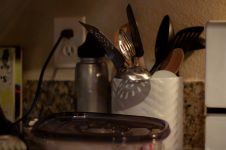}&\hspace{-4.2mm}
\includegraphics[width=0.124\textwidth]{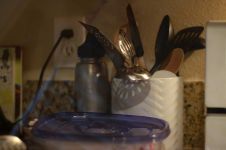}&\hspace{-4.2mm}
\includegraphics[width=0.124\textwidth]{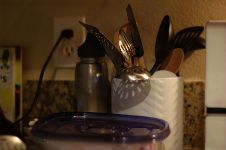}&\hspace{-4.2mm}
\includegraphics[width=0.124\textwidth]{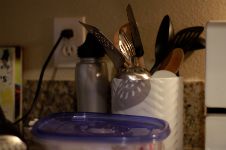}
\\
Input&\hspace{-4.2mm}
CEILNet~\cite{CEILNet}&\hspace{-4.2mm}
Zhang \etal~\cite{Zhang2018CVPR}&\hspace{-4.2mm}
BDN~\cite{BDN}& \hspace{-4.2mm}
ERRNet~\cite{ERRNet}& \hspace{-4.2mm}
IBCLN~\cite{IBCLN}& \hspace{-4.2mm}
Ours& \hspace{-4.2mm}
Ground-truth
\\
	\end{tabular}}
	\captionof{figure}{Visual comparison on four real-world datasets. The first column and the last column are the inputs and  ground-truth transmission layers. From top to down, the images are from \emph{SIR$^2$ Solid}, \emph{SIR$^2$ Postcard}, \emph{SIR$^2$ Wild} and \emph{Real 20} datasets, respectively, and we show three images for each dataset. Please zoom in for better observation.}
	\label{fig:comparison00}
\end{strip}

\begin{figure}[t]
	\centering
	\tiny
	\scriptsize
	\scriptsize
	\small
	\begin{tabular}{cccc}
		\includegraphics[width=0.11\textwidth]{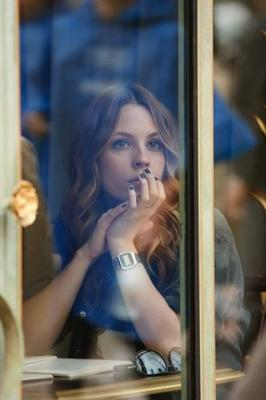}&\hspace{-4mm}
		\includegraphics[width=0.11\textwidth]{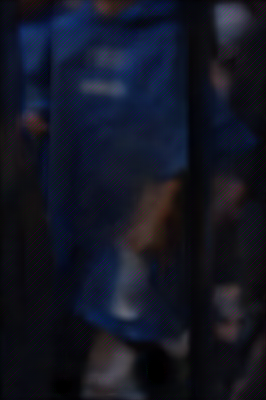}&\hspace{-4mm}
		\includegraphics[width=0.11\textwidth]{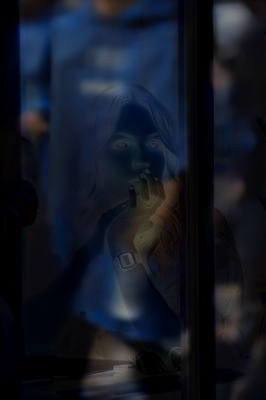}&\hspace{-4mm}
		\includegraphics[width=0.11\textwidth]{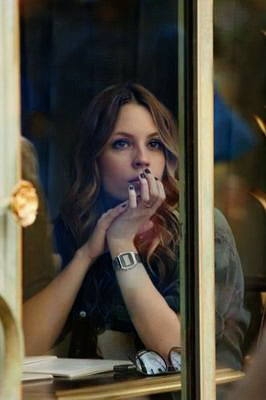}
		\\
		\includegraphics[width=0.11\textwidth]{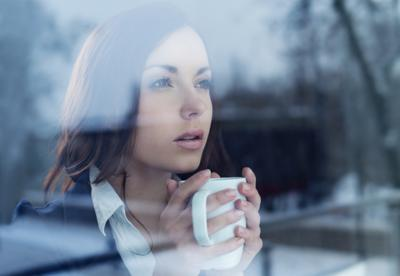}&\hspace{-4mm}
		\includegraphics[width=0.11\textwidth]{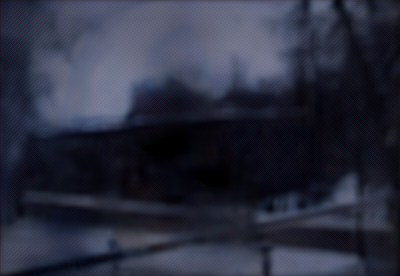}&\hspace{-4mm}
		\includegraphics[width=0.11\textwidth]{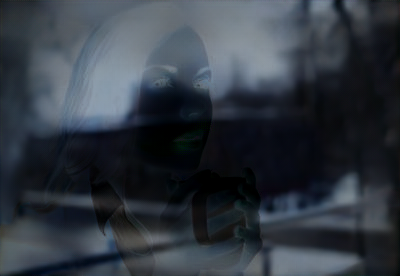}&\hspace{-4mm}
		\includegraphics[width=0.11\textwidth]{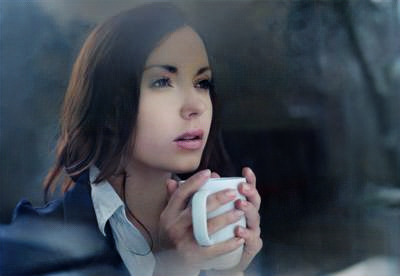}
		\\
		\includegraphics[width=0.11\textwidth]{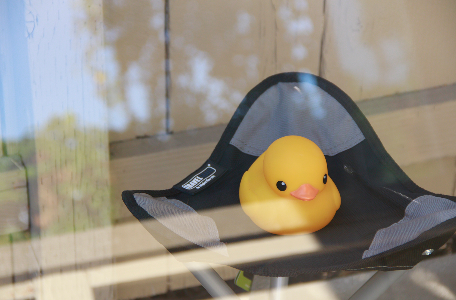}&\hspace{-4mm}
		\includegraphics[width=0.11\textwidth]{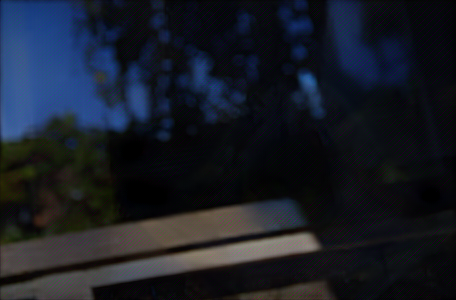}&\hspace{-4mm}
		\includegraphics[width=0.11\textwidth]{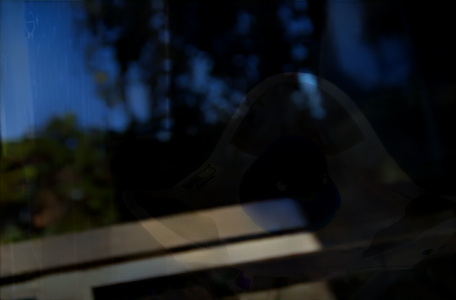}&\hspace{-4mm}
		\includegraphics[width=0.11\textwidth]{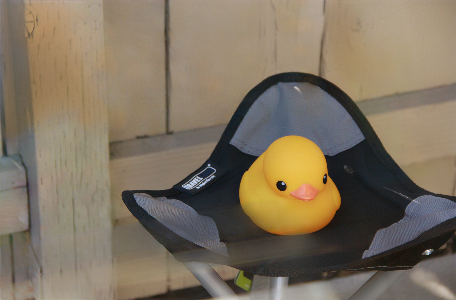}
		\\
		\includegraphics[width=0.11\textwidth]{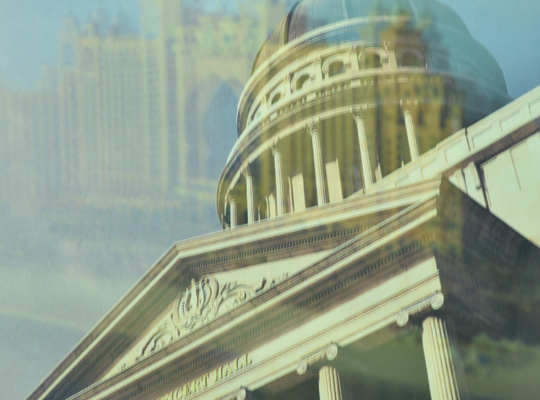}&\hspace{-4mm}
		\includegraphics[width=0.11\textwidth]{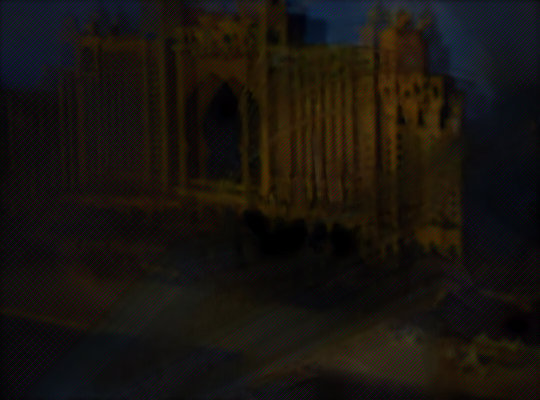}&\hspace{-4mm}
		\includegraphics[width=0.11\textwidth]{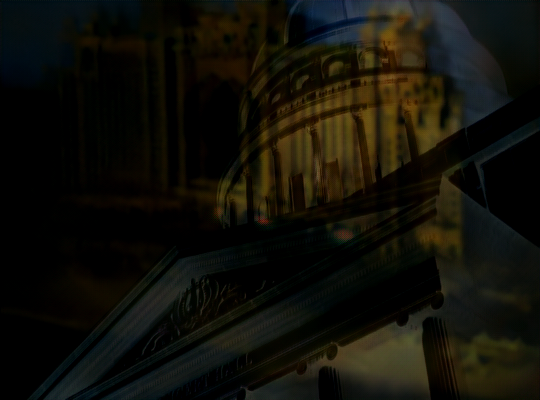}&\hspace{-4mm}
		\includegraphics[width=0.11\textwidth]{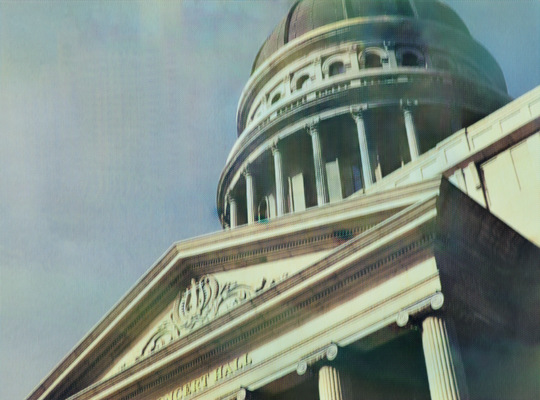}
		\\
		{$I$}&\hspace{-4mm}
		{$\hat{R}$}&\hspace{-4mm}
		{$I-\hat{T}$}&\hspace{-4mm}
		{$\hat{T}$}\
		\\
	\end{tabular}
	\caption{Performance when images are fully covered by reflection.}
	\label{fig:whole}
	\vspace{-3mm}
\end{figure}

\section{Artifacts in IBCLN~\cite{IBCLN}}\label{sec:Results}
IBCLN~\cite{IBCLN} iteratively refines the transmission layer and reflection layer, which promotes the SIRR performance but leads to undesired artifacts in some scenarios. Fig.~\ref{fig:ibcln} shows the phenomena.

\section{Reflection Removal Results}\label{sec:Results}

Fig.~\ref{fig:comparison01} and Fig.~\ref{fig:comparison00} show more reflection removal results of all competing methods (\ie, CEILNet~\cite{CEILNet}, Zhang \etal~\cite{Zhang2018CVPR}, BDN~\cite{BDN}, ERRNet~\cite{ERRNet} and IBCLN~\cite{IBCLN} and our RAGNet).
Fig.~\ref{fig:comparison01} shows results from \emph{Real 45}, where the ground-truth is unavailable.
Fig.~\ref{fig:comparison00} are results from \emph{SIR$^2$ Solid}, \emph{SIR$^2$ Postcard}, \emph{SIR$^2$ Wild} and \emph{Real 20} test datasets, and we show three images from each dataset.

\section{Results on Images with Fully Covering Reflection}\label{sec:wholeref}
Fig.~\ref{fig:whole} further shows the results on four images with fully covering reflection.
It can be seen that RAGNet performs equally well in this scenario.

\section{User Study on \emph{Real45}}\label{sec:userstudy}
{A user study on \emph{Real45} dataset is conducted to evaluate the quality of reflection removal.
Our method is compared with three other methods, including ERRNet~\cite{ERRNet} and IBCLN~\cite{IBCLN} which perform the second and third best in terms of PSNR / SSIM metrics, and Zhang \etal\cite{Zhang2018CVPR} which performs well on \emph{Real20} dataset.
We randomly select 20 image pairs from \emph{Real45} dataset.
30 participants are invited for the user study and each of them is given 20 questions.
Each question contains the original input image and four choices, which are the results of the four candidate methods.
The choices are arranged in a random order for a fair comparison.
The users are instructed to choose the best result in terms of reflection removal ability and image quality.
The results are reported in Table~\ref{userstudy}.
It can be seen that our RAGNet has a much higher probability to be chosen as the the best results among the four methods.
}

\begin{table}[t]
	\centering
	\captionsetup{font={small}}
	\caption{User study on \emph{Real 45} dataset. The percentages represent the voting results by the participants.}
	\label{userstudy}
	\begin{tabular}{ccccc}
		\toprule
		Method & Zhang \etal & ERRNet & IBCLN & Ours\\
		\midrule
		\emph{Real 45} & 30.3$\%$ & 10.5$\%$ & 9.0$\%$ & 50.17$\%$\\
		\bottomrule
	\end{tabular}%
\end{table}

\section{{Performance with Additional Training Data}}\label{sec:Nature}

{We retrain our model by including additional 200 training image pairs from~\cite{IBCLN} and report the results in Table~\ref{tab:retrain}.
Except \emph{Real20}, the inclusion of additional images in training is beneficial to network performance on all the other datasets.


\begin{figure}[t]
	\centering
	\tiny
	\scriptsize
	\scriptsize
	\small
	\begin{tabular}{cccc}
		\includegraphics[width=0.11\textwidth]{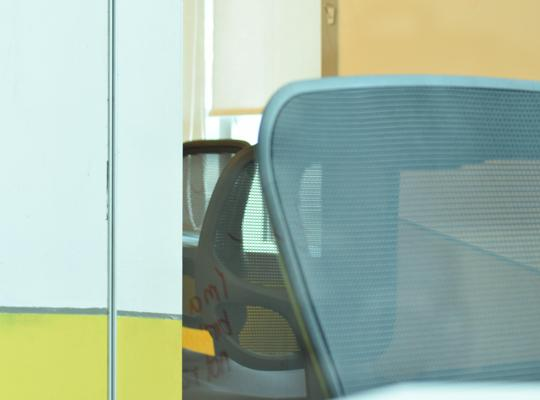}&\hspace{-4mm}
		\includegraphics[width=0.11\textwidth]{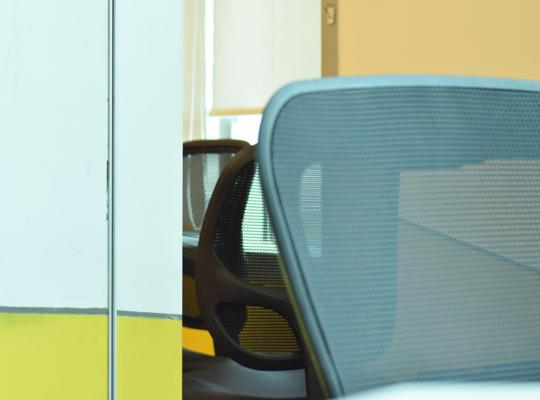}&\hspace{-4mm}
		\includegraphics[width=0.11\textwidth]{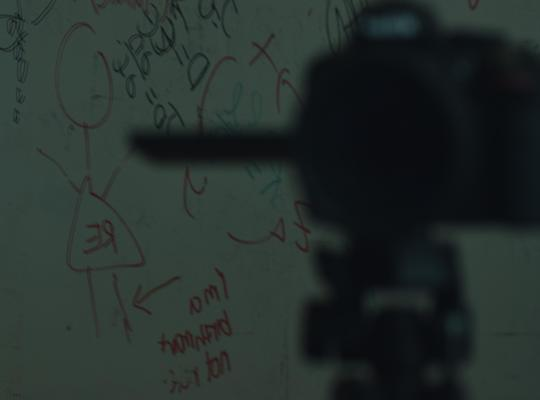}&\hspace{-4mm}
		\includegraphics[width=0.11\textwidth]{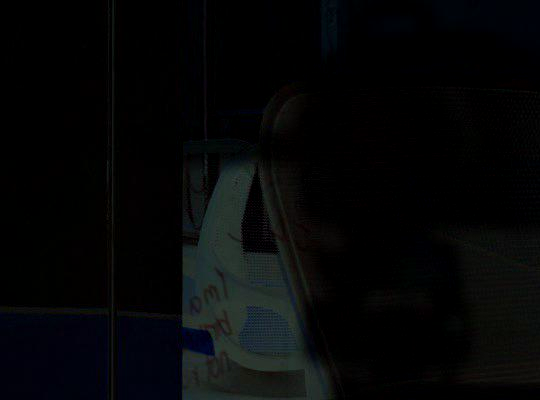}
		\\
		\includegraphics[width=0.11\textwidth]{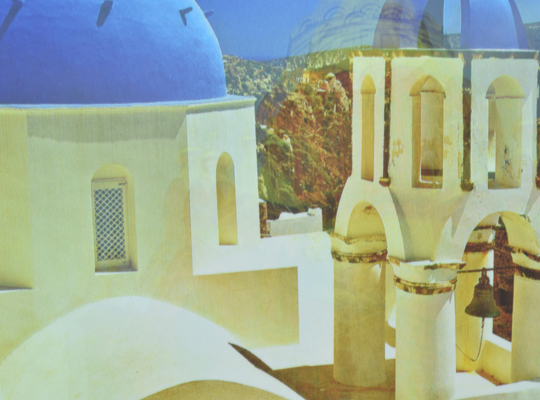}&\hspace{-4mm}
		\includegraphics[width=0.11\textwidth]{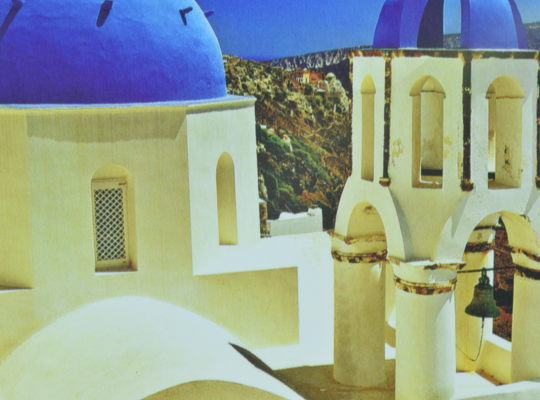}&\hspace{-4mm}
		\includegraphics[width=0.11\textwidth]{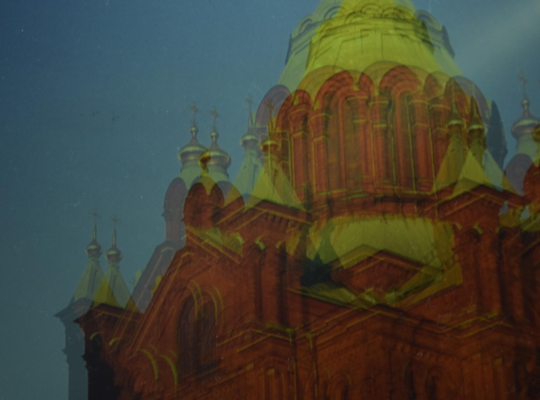}&\hspace{-4mm}
		\includegraphics[width=0.11\textwidth]{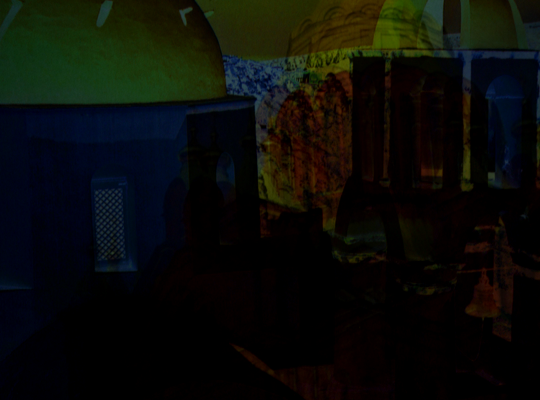}
		\\
		{input}&\hspace{-4mm}
		{$T_{gt}$}&\hspace{-4mm}
		{$R_{gt}$}&\hspace{-4mm}
		{$I-T_{gt}$}\
		\\
	\end{tabular}
	\caption{Some regions of the $R_{gt}$s provided by \textit{SIR$^2$} are unobservable in the corresponding areas of blended images.}
	\label{fig:SIR}
	\vspace{-2mm}
\end{figure}

\section{Detection of Reflection} \label{sec:evalR}
To verify whether the reflection region can be successfully detected by our network, we compare the similarity between the estimated reflection $\hat{R}$ and the residual $I-T_{gt}$. As a result, the average PSNR is 24.01 dB on the 4 test sets of \emph{SIR$^2$} and \emph{Real 20}.
It should be noted that the given ground truth $R_{gt}$ in \textit{SIR$^2$} generally covers the whole image but is partially invisible in input $I$ (see Fig.~\ref{fig:SIR}). Therefore, we use $I-T_{gt}$ instead of $R_{gt}$ for evaluating the detection of reflection.

\section{{Ability in Removing Strong and Weak Reflection}}\label{sec:Reflection_aware}
While our RAGNet is suggested for handling strong reflection regions, it also works well on the regions with weak reflection. 
To verify this point, we calculate the PSNR respectively in the weak and strong reflection regions on the 4 test sets (\emph{SIR$^2$} and \emph{Real 20} datasets).
As shown in Fig.~\textcolor{red}{3}, the elements of $\mathbf{M}_\mathit{diff}$ tend to be 0 in strong reflection regions and 1 in weak reflection regions.
{Therefore, we can separately evaluate the network performance by calculating PSNR in regions with weak reflection intensity based on $\mathbf{M}_w = \mathbf{M}_\mathit{diff} > \tau (=0.40)$.}
In specific, given the $H\times W$ estimated transmission $\hat{T}$ and the ground truth transmission $T_{gt}$, the PSNR for weak reflection regions is calculated as:
\begin{equation}
\label{eqn:weighted_PSNR}
{\rm PSNR}=10\,{\rm log}_{10}\frac{I_{max}}{{\rm MSE}_{mask}},
\end{equation}
where
\begin{equation}
\label{eqn:weighted_MSE}
{\rm MSE}_{mask}\!=\!\frac{\sum_{i=0}^{H-1}\!\sum_{j=0}^{W-1}\!{\rm \mathbf{M}}_w(i,j)[\hat{T}(i,j)-T_{gt}(i,j)]^2}{\sum_{i=0}^{H-1}\!\sum_{j=0}^{W-1}\!{\rm \mathbf{M}}_w(i,j)},
\end{equation}
where $I_{max}$ is the maximum possible pixel value.
Accordingly, the PSNR for strong reflection regions can be calculated by replacing ${\rm \mathbf{M}}_\mathit{w}(i,j)$ with ${\rm \mathbf{M}}_\mathit{s}(i,j) = 1-{\rm \mathbf{M}}_\mathit{w}(i,j)$.
%
%
The results are shown in Table \ref{tabC}.
{It can be seen that: 1) For weak / no reflection regions, our method can achieve 0.9 dB PSNR gain. 2) For heavy reflection regions, the PSNR by our method exceeds the others by 1.9 dB.}

\begin{table}
	\small
	\centering
	\scalebox{.92}{
		\begin{tabular}{c}
			\includegraphics[width=0.5\textwidth]{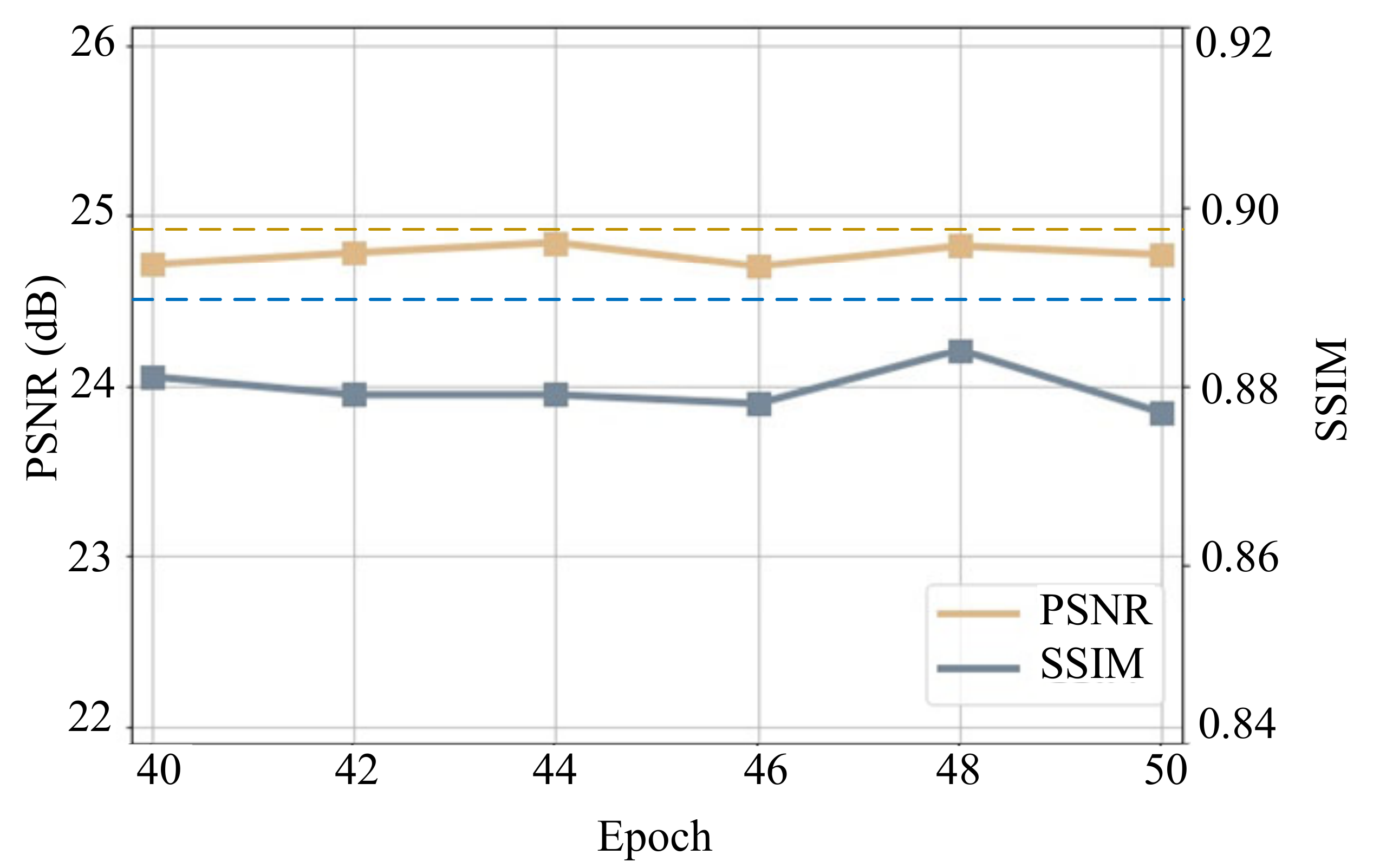}
			\\
	\end{tabular}}
	\vspace{-3mm}
	\captionof{figure}{The fluctuation of PSNR / SSIM with $G_R$ being individually  trained for different epochs. The dotted lines refer to the PSNR / SSIM under the original setting.}
	\label{fig:R}
\end{table}

\section{Robustness Against $\hat{R}$} \label{sec:evalR}
Empirically, $\hat{R}$ with higher quality yields better network performance.
Nonetheless, RAGNet also exhibits considerable robustness against $\hat{R}$ within a certain range.
To show this, we have individually trained a $G_R$ for different epochs and feed its prediction to $G_T$.
As shown in Fig.~\ref{fig:R}, only slight fluctuation of network performance is observed given $\hat{R}$ generated in different epochs, indicating the robustness of RAGNet against $\hat{R}$.

\end{document}